\newif\iffullversion
\fullversiontrue

\documentclass[conference]{IEEEtran}
\ifCLASSINFOpdf
\else
\fi

\usepackage{booktabs}
\usepackage{color}
\usepackage{graphics}
\usepackage{amsmath,amsfonts, amssymb}
\usepackage[caption=false,font=normalsize,labelfont=sf,textfont=sf]{subfig}
\usepackage{mathtools}
\usepackage{epstopdf}
\usepackage{amsthm}
\usepackage{bbm}
\usepackage{algorithm}
\usepackage[noend]{algpseudocode}
\usepackage{dsfont}
\usepackage{cite}

\usepackage[breaklinks]{hyperref}
\usepackage{multirow}
\newcommand{\revised}[1]{{\color{black}{#1}}}

\begin{document}
%
\title{Byzantines can also Learn from History:\\ Fall of Centered Clipping in Federated Learning}



%


\author{
\IEEEauthorblockN{Kerem Özfatura}
\IEEEauthorblockA{ KUIS AI Center\\
Koç University\\
aozfatura22@ku.edu.tr}
\and
\IEEEauthorblockN{Emre Özfatura}
\IEEEauthorblockA{ IPC Lab\\
Imperial College London\\
m.ozfatura@imperial.ac.uk}
\and
\IEEEauthorblockN{Alptekin Küpçü}
Dept. of Computer Engineering \\
\IEEEauthorblockA{Koç University\\
akupcu@ku.edu.tr}
\and
\IEEEauthorblockN{Deniz Gunduz}
\IEEEauthorblockA{ IPC Lab \\ Imperial College London\\
d.gunduz@imperial.ac.uk}
}

\maketitle

\begin{abstract}
The increasing popularity of the federated learning (FL) framework due to its success in a wide range of collaborative learning tasks also induces certain security concerns. Among many vulnerabilities, the risk of Byzantine attacks is of particular concern, which refers to the possibility of malicious clients participating in the learning process. Hence, a crucial objective in FL is to neutralize the potential impact of Byzantine attacks and to ensure that the final model is trustable. It has been observed that the higher the variance among the clients' models/updates, the more space there is for Byzantine attacks to be hidden. As a consequence, by utilizing momentum, and thus, reducing the variance, it is possible to weaken the strength of known Byzantine attacks. The \textit{centered clipping} (CC) framework has further shown that the momentum term from the previous iteration, besides reducing the variance, can be used as a reference point to neutralize Byzantine attacks better. In this work, we first expose vulnerabilities of the CC framework, and introduce a novel attack strategy that can circumvent the defences of CC and other robust aggregators and reduce their test accuracy up to \%33 on best-case scenarios in image classification tasks. Then, we propose a new robust and fast defence mechanism that is effective against the proposed and other existing Byzantine attacks. 
\end{abstract}

\section{Introduction}
Federated learning (FL) is a novel learning paradigm whose is to enable large-scale {\em collaborative learning} in a distributed manner, possibly among edge devices and without sharing local datasets, addressing, to some extent, the privacy concerns of the end-users  \cite{FedAVG1}. Due the these privacy concerns, it allows many IoT and edge devices to participate in collaborative learning \cite{semi1,semi3}. In FL, the learning process is often orchestrated by a central entity called the \textit{parameter server} (PS). Participating clients first update their local models using their local private datasets, and then communicate these local models with the PS to seek a consensus with other participating clients. The PS utilizes an aggregation rule to obtain the consensus/global model, which is then sent back to the clients to repeat the process again until the global model achieves a certain generalization capability. 

Since FL allows on-device training and can be scaled to a large number of clients, it has become the \textit{de facto} solution for several practical and commercial implementations, e.g., learning keyboard prediction mechanisms on edge devices \cite{FL.keyboard1,FL.keyboard2}, or digital healthcare /remote diagnosis  \cite{FL.health1,FL.health2,FL.health3}. However, its widespread popularity also accompanies certain concerns regarding privacy, security, and robustness, particularly for those applications involving highly sensitive financial, \cite{FL.finance} or medical \cite{FL.health4, FL.health6} datasets . Hence, ultimately, the target is to make FL secure, privacy-preserving \cite{privacy1, privacy2}, and robust against data heterogeneity \cite{robustu,semi2}.

With the scaling of FL, the PS has less control over the participating clients: that is, malicious clients can also participate with the goal of impairing the final model. Thus, the key challenge for FL is to ensure that the consensus model is trustable despite the potential presence of {\em adversaries}. In the machine learning literature, adversaries have been studied in many different contexts and scenarios. They can be analyzed from the perspective of {\em robustness} and {\em security}. Adverserial robustness in FL refers to mitigating adversaries that attack a trained model at \textit{test time} by hand-crafted test samples to make the learned model fail the task \cite{adv1, adv2, adv3, adv4}, whereas the security is considered against adversaries that can attack the \textit{training process} with the aim of manipulating the model so that it fails the task at test time. In this work, we focus on the latter. 

\textbf{A brief taxonomy of security threat models.}
The threat models that target the training/learning process are often described as {\em poisoning attacks}. We classify them based on three aspects: {\em activation}, {\em location}, and {\em target}. 
We use the term {\em activation} to describe attacks that are embedded into the model during training as a {\em trojan} and activated at test time using a specific triggering signal, which is often referred to as a {\em backdoor attack} \cite{DBA,backdoor_FL,Backdoor_tail,TBA,HTBA}, or the attack does not require a triggering mechanism, which is often called a {\em Byzantine attack} \cite{ALIE,IPM,LMPA,CC,Byz.momentum,Bucketing,Bulyan,Trimmed_mean,Krum}. The second aspect is the \textit{location} of the interaction of the attack with the learning process. When a poisoning attack targets the training data, we describe it as a {\em data poisoning attack}, whereas if the attack directly targets the model, e.g., through the local updates in FL, we describe it as a {\em model poisoning attack}. The last aspect we consider is whether the attack is {\em targeted}; that is, failure of the learning task is desired for a particular type of test data (for instance, in classification tasks, certain classes might be the target), or {\em untargeted}; that is, the failure of the task is desired for any possible test data. Attacks can also be classified based on their knowledge of the learning process (white box / black box), or on their capability and control over the system. For further discussion on the taxonomy of poisoning attacks, we refer the reader to \cite{backdoor_FL, B2DB, AFLAL}.

\textbf{Scope and Contributions.}
In this work, we focus on untargeted model poisoning attacks as they are the most effective against FL to disrupt the collaborative learning process \cite{CC, ndss2022_Attack}. Various Byzantine attack strategies against FL have been introduced in the literature \cite{ALIE, IPM, LMPA, ndss2022_Attack, backdoor_FL}, as well as a number of robust aggregation frameworks that replace federated averaging (FedAVG) \cite{FedAVG1} for defence against these \cite{CC, Byz.momentum, Bucketing, Bulyan, Trimmed_mean, Krum}. One common observation regarding untargeted model poisoning attacks, often valid for other attacks as well, is that when the variance among the honest updates is low, that is, the client updates are more correlated, and an attack can be easily detected as an outlier. On the other hand, it is harder to detect an attack as an outlier when the variance of the honest updates is large. 

As a direct consequence of the observation above, {\em variance reduction} techniques for stochastic gradient descent (SGD) can provide protection against Byzantine attacks, in addition to their primary task of accelerating the convergence. Indeed, in \cite{Byz.momentum}, the authors have shown that it is possible to improve the robustness of the existing defence mechanisms by using local momentum. They also argue that the use of global momentum does not serve the same purpose. In \cite{CC}, the authors further extend the use of local momentum, and show that its advantage against Byzantine attacks is two-fold: First, it helps reduce the variance, and thus, leaves less space for an attack to be embedded without being detected, and second, the aggregated local momentum from the previous iteration can be utilized as a reference point for the {\em centered clipping (CC)} mechanism to achieve a robust aggregation rule. Indeed, it has been shown that such a combined mechanism of local momentum and CC demonstrates impressive results against many state-of-the-art (SotA) Byzantine attacks, such as {\em a little is enough (ALIE)} \cite{ALIE} and {\em inner product multiplication (IPM)} \cite{IPM} attacks. This design has been further extended to the {\em decentralized} framework and combined with other variance reduction techniques \cite{SCC,Bucketing}.

In this work, we show that the CC mechanism gives a false sense of security, and its performance against ALIE and IMP might be misleading. We show that the existing ALIE and IMP attacks can be redesigned to circumvent the CC defence. We first identify the main vulnerabilities of the CC method, and then, by taking into account certain design aspects of ALIE and IPM, we propose a new Byzantine attack that targets the CC defence. We show numerically that our proposed strategy successfully circumvents the CC defence, as well as other well-known defense mechanisms (i.e., Robust FedAVG (RFA) \cite{RFA} and Trimmed-Mean (TM) \cite{Trimmed_mean}). We then propose an improved defence against the proposed attack and show its effectiveness. The contributions of this work can be summarized as follows:

\begin{itemize}
\revised{\item We first analyze the CC framework and identify its key vulnerabilities. To the best of our knowledge, we are the first to investigate these vulnerabilities, while other succeeding works \cite{decentcc} also mention the weaknesses in Decentralized FL settings.}

\item We revisit the known time-coupled attacks ALIE and IPM, and by taking into account the vulnerabilities of CC that we identified, we introduce a new Byzantine attack called relocated orthogonal perturbation (ROP) that utilizes the broadcasted momentum term to circumvent the CC defence. 

\item By conducting extensive simulations on various datasets, both with independent and identically distributed (IID) and non-IID data distributions across the clients, and with different neural network architectures, we show that the proposed {\em ROP} attack is effective against the CC defence, as well as other robust aggregators such as RFA and TM. 

\item Finally, we introduce a new defence mechanism against the proposed Byzantine attack as well as others, with the same asymptotic complexity as the CC aggregator. We show the robustness of the proposed defence mechanism for both IID and non-IID data distributions.

\end{itemize}

\section{Background and Related Work}

\textbf{Robust aggregators.} Defence mechanisms against the presence of Byzantine agents in collaborative learning and distributed computations has been studied in the literature for nearly 40 years \cite{byzantine_orig}. With the increasing deployment of large-scale systems that employ collaborative learning, such as FL, the risks and potential consequences of such attacks are also growing \cite{FL_problems}. Many robust aggregation methods have been studied to counter possible adversarial clients. Most solutions replace FedAVG \cite{FedAVG1} with more robust aggregation methods built upon various statistical and geometrical assumptions, such as coordinate-wise median (CM) \cite{Trimmed_mean}, geometric median \cite{geo_def1,statistical_related,RFA}, and consensus reaching methods like majority voting \cite{SignSGD}. However, since these aggregators are based on purely statistical assumptions, their robustness fails against the  SotA attacks, in which the adversaries can statistically conceal themselves as benign clients. Furthermore, these assumptions may not hold in real FL implementations, in which the data distributions tend to be heterogeneous (non-IID) across clients. As a result of which, benign clients may be labeled as adversaries, and discarded by the aggregator.

TM \cite{Trimmed_mean} has been proposed as an improvement over CM by calculating the mean after discarding the outliers. RFA \cite{RFA}  addressed the issue of heterogeneous data distributions by employing a geometric median in their aggregator. Krum/Multi-Krum \cite{Krum} calculate a geometric distance from each client to every other participating client to score them based on their distances, then discard the clients with the lowest scores. One particular downside of the Krum and Multi-Krum methods is that due to the scoring function, they are slower aggregators ($O(k^{2})$) compared to other aggregators ($O(k)$), where $k$ denotes the number of clients. Bulyan \cite{Bulyan} is proposed to prevent Byzantine clients that target very specific locations in the gradients, while being close to the mean in the rest of the gradient values. Bulyan uses the same selection method in Krum, and then applies TM to the selected subset of clients. Nevertheless, these traditional aggregators discard outlier benign clients one way or another \cite{CC}; and therefore, their robustness tends to fail in the case of heterogeneous data distributions. Recently, CC \cite{CC} is proposed to aggregate all the participating clients, where outlier gradients are scaled and clipped based on the center that is selected by the aggregator using the history of the previous aggregations. This provides better convergence by not fully discarding the outliers and a natural defense mechanism against Byzantines that can act as an outlier.

\textbf{Incorporating acceleration frameworks.}
The momentum SGD \cite{SGD.nesterov,SGD.polyak} has been introduced to accelerate the convergence of the SGD framework and to escape saddle points to achieve a better minima \cite{SGD.opt1,SGD.opt2}. These advantages of the acceleration methods, particularly momentum SGD, promote their use also in the FL framework, with the possibilities of incorporating momentum locally at the clients and globally at the server \cite{FL.acc1}. In the context of Byzantine resilience, only a limited number of works have analyzed the impact of the momentum SGD. In \cite{Byz.momentum}, it has been shown that, in terms of Byzantine resilience, utilizing momentum locally at the clients is better than globally at the PS. In \cite{momentum_2022}, the authors propose RESAM (RESilient Averaging of Momentums), which specifically employs the local momentum of the benign clients instead of their gradients.  In \cite{CC}, the authors have shown that, besides reducing the variance among the updates, the momentum term from the previous iteration can also be used as a reference for the momentum of the next iteration in order to neutralize Byzantine attacks trough clipping. However, as we show in this work, malicious clients can also follow a similar strategy to improve their attack strength, and escape from clipping.

\textbf{Model poisoning attacks and defenses.}
We can identify three SotA model poisoning attacks that have often been studied in the literature to circumvent the existing robust aggregators \cite{LMPA,ALIE,IPM}. The common ground of these attacks is that Byzantine clients statistically stay close to benign clients to prevent easy detection and poison the global model by coupling their attacks across multiple iterations. However, these attacks do not consider momentum and directly target the gradient values of the clients. In a recent work \cite{ndss2022_Attack}, the authors show that existing Byzantine-robust FL algorithms are significantly more susceptible to model poisoning attacks than previous SOTA attacks of ALIE \cite{ALIE} and Fang \cite{LMPA} by introducing \emph{min-max} and \emph{min-sum} attacks to amplify the existing poisoning attacks and introduced a divide-and-conquer (DnC) framework to prevent such attacks. Some of the existing defenses for backdoor attacks, such as FLAME \cite{FLAME_Def}, offer some level of robustness against untargeted model poisoning attacks; however, they are not designed to prevent SotA time-coupled model poisonings attacks such as ALIE or IPM. Recently proposed FLtrust \cite{FLtrust} claims to offer more robustness than TM \cite{Trimmed_mean} and Krum\cite{Krum}; however, unlike other aggregators, FLtrust requires part of the dataset to be available at the PS, which may not be possible in most FL applications due to privacy concerns. Finally, proposed model poisoning defenses, namely DnC and FLtrust, do not employ either local or global momentum and only consider the gradient values of the clients.

\section{Preliminaries}

\subsection{Notation}

\revised{
We use $\textbf{bold}$ to denote vectors, i.e., $\boldsymbol{v}$ and capital calligraphic letters, e.g., $\mathcal{V}$, to denote the sets. When we have ordered set of  vectors $\mathcal{V}=\left\{ \boldsymbol{v}_{1},\ldots,\boldsymbol{v}_{i}\ldots,\boldsymbol{v}_{k}\right\}$, we use subscript index to identify $i^{th}$ vector in the set, $i\in[k]$, and  use double subscript $\mathbf{v}_{i,t}$, particularly when it is changing over time/iterations. For slicing operation, we use $[\cdot]$, such as $\mathbf{v}[j]$ for selecting $j^{th}$ index of a vector.  We use $|| \ \cdot \ ||_{p}$ to denote the $p$-norm of a vector; in this paper we use $l_{2}$ and $l_{1}$ norms, and usage of $|| \ \cdot \  ||$ without $p$ corresponds to $l_{2}$ norm. We use $< \cdot \ , \cdot  >$ for the inner product between two vectors. In Table \ref{tab:notations}, we list the variables that are widely used in this paper.
}

\begin{table}[t]
\caption{Notations}
\label{tab:notations}
\centering
\begin{tabular}{|l|l|}
\hline
\textbf{Notation} & \textbf{Description} \\ \hline
$\mathcal{K}$& Set of clients, $\mathcal{K}=\mathcal{K}_{b} \cup \mathcal{K}_{m}$ \\ \hline
$\mathcal{K}_{b}$& Subset of benign clients\\ \hline
$\mathcal{K}_{m}$& Subset of malicious clients\\ \hline
\revised{$\beta$} & \revised{Local momentum constant} \\ \hline
$k$ & Number of  clients, $k=\left |  \mathcal{K} \right |$ \\ \hline
$k_{b}$ & Number of benign clients, $k_{b} = \left |  \mathcal{K}_{b} \right |$ \\ \hline
$k_{m}$ & Number of Byzantines, $k_{m} = \left |  \mathcal{K}_{b} \right |$ \\ \hline
$T$ & Number of iterations \\ \hline
$\eta$ & Learning rate \\ \hline
$\boldsymbol{\theta}_{i,t}$ & Model parameters of client $i$ at iteration $t$ \\ \hline
$\mathbf{g}_{i,t}$ & Gradient vector of client $i$ at iteration $t$ \\ \hline
$\mathbf{m}_{i,t}$ & Momentum vector of client $i$ at iteration $t$ \\ \hline
$\tilde{\mathbf{m}}_{t}$ & Aggregate momentum at time $t$ \\ \hline
$\bar{\mathbf{m}}_{t}$ & Benign aggregate momentum at time $t$ \\ \hline
$\tau$ & Radius of the CC aggregator \\ \hline
$\revised{\lambda}$ & \revised{Reference point hyper-parameter} \\ \hline
$\revised{\rho}$ & \revised{Attack location point hyper-parameter} \\ \hline
$\revised{\pi}$ & \revised{Degree of the attack w.r.t reference point} \\ \hline
\end{tabular}
\end{table}

\subsection{Federated Learning (FL)}
The objective of  FL is to solve the following parameterized optimization problem over $k$ clients in a distributed manner
\begin{equation}
\min_{\boldsymbol{\theta}\in\mathbb{R}^{d}} f(\boldsymbol{\theta})= \frac{1}{k}\sum^{k}_{i=1}\underbrace{\mathds{E}_{\zeta_{i} \sim \mathcal{D}_{i}}f(\boldsymbol{\theta},\zeta_{i})}_{\mathrel{\mathop:}=f_{i}(\boldsymbol{\theta})},\label{DSO}
\end{equation}
where $\boldsymbol{\theta}\in\mathbb{R}^{d}$ denotes the model parameters, e.g., weights of a neural network,  $\zeta_{i}$ is the randomly sampled mini-batch from $\mathcal{D}_{i}$, which denotes the dataset of client $i$, and $f$ is the problem-specific empirical loss function. At each iteration of FL,  each client aims to minimize its local loss function $f_{i}(\boldsymbol{\theta})$ using {\em SGD}. Then, the clients seek a consensus on the model with the help of the PS.
\begin{algorithm}[t]
\small
\caption{Robust FL with Byzantines }\label{code:robustFL}
 \textbf{Input:} Learning rate $\eta $, Aggregator $AGG(\cdot)$, Attack $Attack(\cdot)$\\
  \textbf{Output:} Consensus model: $\boldsymbol{\theta}_{T}$
\begin{algorithmic}[1]
    \For{$t=1,\ldots,T$}
    \State \textbf{Client side:}
     \For{$i=1,\ldots,k$} in parallel
     \State Receive: $\boldsymbol{\theta}_{t-1}$ from PS
     \If {$i\in\mathcal{K}_{b}$}
     \State Update local model: $\boldsymbol{\theta}_{i,t} \xleftarrow{} \boldsymbol{\theta}_{t-1}$
     \State Compute SGD: $\mathbf{g}_{i,t}\gets \nabla_{\boldsymbol{\theta}}f_{i}(\boldsymbol{\theta}_{i,t},\zeta_{i,t})$
     \State Update Momentum: 
     \State $\mathbf{m}_{i,t} = (1-\beta) \mathbf{g}_{i,t} + \beta \mathbf{m}_{i,t-1}$
     \Else
     \State $\mathbf{m}_{i,t} \gets Attack(\mathcal{H}_{t})$
     \EndIf
    \EndFor
\textbf{Server side:}
\State{Aggregate local updates:}
\State $\tilde{\mathbf{m}}_{t}\gets AGG(\mathbf{{m}}_{1,t}, \ldots, \mathbf{m}_{k,t})$
\State Update server model: $\boldsymbol{\theta}_{t} \gets \boldsymbol{\theta}_{t-1} -  \eta_{t}\tilde{\mathbf{m}}_{t}$
\State Broadcast model $\boldsymbol{\theta}_{t}$
    \EndFor
\end{algorithmic}
\end{algorithm}

 In every communication round $t$, the PS sends its current model $\boldsymbol{\theta}_{t}$ to every client to synchronize their local models $\boldsymbol{\theta}_{i,t}$. After updating their local models, benign clients first compute  gradients $\mathbf{g}_{i,t}$ by randomly sampling a batch $\zeta_{i,t}$, then update their local momentum $\mathbf{m}_{i,t}$, using the local client update scheme: $\mathbf{m}_{i,t} = \ (1-\beta)\mathbf{g}_{i,t} +  \beta \mathbf{m}_{i,t-1}$ to further reduce the variance. The benign client-side of the FL framework corresponds to lines 3-9 of Algorithm \ref{code:robustFL}.

Malicious clients, so-called Byzantines, return a poisoned model to the PS according to a certain attack strategy $Attack(\cdot)$ by utilizing all the possible observations until time $t$, denoted by $\mathcal{H}_{t}$, corresponding to line 11 in in Algorithm $\ref{code:robustFL}$.

\textbf{Adversarial model:}
We assume that the Byzantine clients are omniscient, meaning that they have all the information on the dataset, and can use it to predict the gradients of benign clients, which is in line with other SotA attacks, such as ALIE\cite{ALIE} and IPM\cite{IPM}.
An omniscient Byzantine attacker can also calculate the gradient of the benign clients and store the benign and attacker momentum values to generate an arbitrary momentum and then use it to calculate the benign $\mathbf{m}_{i,t}$ momentum of the respective clients in $\mathcal{K}_{b}$. 
Byzantine attackers are also assumed to know the learning rate $\eta$ ; and thus can estimate the aggregated momentum value $\tilde{\mathbf{m}}_{t-1}$ generated by the PS. Ultimately, the Byzantine client can utilize this information to create a model poisoning attack in an agnostic manner, i.e., the attacker does not know the aggregator or any deployed defences used by the PS.   

\subsection{SotA model poisonings attacks} \label{attacks}
In this subsection, we provide a brief overview of the SotA model poisoning attacks that couple their attacks across iterations to increase their impact without being detected.
\subsubsection{ALIE} Traditional aggregators such as Krum \cite{Krum}, TM \cite{Trimmed_mean} and Bulyan \cite{Bulyan} assume that the selected set of parameters will lie within a ball centered at the real mean within a radius, which is a function of the number of benign clients. The attacker in \cite{ALIE} utilizes index-wise mean ($\bar{\mathbf{m}}$) and standard deviation ($\bar{\boldsymbol{\sigma}}$) vectors of the benign clients to induce small but consistent perturbations to the parameters. By keeping the momentum values close to $\bar{\mathbf{m}}$, ALIE can steadily achieve an accumulation of errors while concealing itself as a benign client during training. To avoid detection and stay close to the center of the ball, ALIE scales $\bar{\boldsymbol{\sigma}}$ with a $z^{max}$ parameter, which is calculated based on the numbers of benign and Byzantine clients. As such, let $s$ be the minimal number $s = \left[ \frac{k}{2} + 1 \right] - k_{m}$ of benign clients that are required as “supporters". The attacker will then use the properties of the normal distribution, specifically the cumulative standard normal function $\phi(z)$, and look for the maximal $z^{max}$ such that $s$  benign clients will have a greater distance to the mean compared to the Byzantine clients, such that, those $s$ clients are more likely to be classified as Byzantines. In a high level, $z^{max}$ can be calculated as: 

\revised{
\begin{equation}
z^{max} = \mathrm{max} \left \{ z: \phi(z) < \frac{k-k_{m}-s}{k-k_{m}}  \right \} 
\end{equation}
}

Ultimately, $z^{max}$ is employed as a scaling parameter for the standard deviation to perturb the mean of the benign clients in set $\mathcal{K}_{b}$:

\begin{equation}\label{eq:alie}
    \mathbf{m}_{i} = \bar{\mathbf{m}} - z^{max} \mathbf{\boldsymbol{\bar{\sigma}}} \: , i\in\mathcal{K}_{m},
\end{equation}

where $\mathcal{K}_{m}$ is the set of Byzantine clients. Each individual Byzantine client generates an attack with a momentum value near  $\bar{\mathbf{m}}$, following (\ref{eq:alie}).

\subsubsection{IPM}

In \cite{IPM}, the authors approach the problem from a stochastic optimization perspective and highlight the required condition for the convergence of the gradient descent framework, that is, the inner product between the benign gradient $\bar{\mathbf{g}}$ and the output of the robust estimator should be positively aligned, i.e.,
\begin{equation}\label{ipm_orig}
 \langle \bar{\mathbf{g}}, AGG(\mathbf{g}_{i}: i \ \epsilon \ \mathcal{K} ) \rangle \geq 0,
\end{equation}
which ensures that the loss is steadily minimized over iterations. IPM is designed to break this condition and obstruct convergence. From the attacker's perspective, the most effective strategy to make (\ref{ipm_orig}) invalid is to use benign gradient values with  inverted signs. However, since most robust aggregators are designed to ensure that the output of robust aggregation will not deviate from the benign gradient, often by using distance to the median as a trust metric, an adversary with $-\bar{\mathbf{g}}$ can be spotted easily. Therefore, the second step of  IPM is to choose the proper scaling parameter to make the adversary stealthy yet effective. Finally, we remark that though at first glance scaled version of the attack might seem insufficient, the convergence implies $\bar{\mathbf{g}}$ approaches to $0$ over iterations; hence, in such a regime accumulation of the adversarial gradients can invalidate condition (\ref{ipm_orig}).

\subsection{Robust Aggregators} 
Several robust aggregation algorithms have been proposed in the literature to limit the impact of attacks coupled across iterations.
The CC algorithm in \cite{CC} exploits the clipping function $f_{CC}$ to normalize a potential Byzantine client's momentum that resides far away from a selected reference point. CC considers $\tilde{\mathbf{m}}_{t-1}$ as the reference point to scale $\mathbf{m}_{i,t}$ from client $i\in\mathcal{K}$. Whether the client is Byzantine or a benign client with very heterogeneous data, $f_{CC}$ scales down and pulls back $\mathbf{m}_{i,t}$ closer to the reference point to ensure a stable update direction, and generates a new stable reference point for the forthcoming iteration:

\begin{align}
f_{CC}(\mathbf{m}\vert \tilde{\mathbf{m}},\tau ) =  \ &\tilde{\mathbf{m}} + \underbrace{\min\left\{1,\frac{\tau}{\vert\vert \tilde{\mathbf{m}} - \mathbf{m} \vert\vert}\right\}}_{\delta} (\mathbf{m}-\tilde{\mathbf{m}}).
\end{align}

\begin{algorithm}[t]
    \small
	\caption{Aggregation with CC}
	\textbf{Inputs:} $\boldsymbol{\tilde{m}_{t-1}}$, $\mathbf{\bar{m}_{t}}$, $\left\{\mathbf{m}_{i,t}\right\}_{i\in\mathcal{K}},f_{CC}(\cdot),\tau$
	\begin{algorithmic}[1]
	 \For{$i=1,\ldots,k$} in parallel
     \State $\tilde{\mathbf{m}}_{i,t} = f_{CC}(\mathbf{m}_{i,t}\vert \tilde{\mathbf{m}}_{t-1},\tau)$
     \EndFor
     \State $\tilde{\mathbf{m}}_{t}=\frac{1}{k}\sum_{i\in\mathcal{K}}\tilde{\mathbf{m}}_{i,t}$
	\end{algorithmic}
	\label{code:CC}
\end{algorithm}

RFA in \cite{RFA}  is a geometric median-based robust aggregation method. The significant difference between RFA and FedAVG is that the former replaces the weighted arithmetic mean with an approximate geometric median, thus limiting the effectiveness of the Byzantine parameters:

\revised{
\begin{equation}
    \text{RFA}(\mathbf{m}_{1}, \dotso, \mathbf{m}_{k}) =\underset{\tilde{\mathbf{m}}}{\mathrm{argmin}}\sum\limits_{i=1}^{k}|| \tilde{\mathbf{m}} - \mathbf{m}_{i}|| 
\end{equation}
}

In TM \cite{Trimmed_mean}, The average is computed after the $k_{m}$ largest and smallest values are discarded; hence the name trimming. Specifically, for a given dimension j, it sorts the values of $j$-th dimension of all updates, i.e.,
sorts $\mathbf{m}^{j}_{\left \{  i \in k \right \}}$. Then it removes $k_{m}$ largest and smallest values and computes the average of the rest of the values as its aggregate of dimension $j$. It is considered an improvement over the coordinate-wise median aggregator. Formally, let $\mathbf{m}_{\omega_{j}(i)}$ denote the sorted values for index $j$, and $\forall \ i \in \mathcal{K}$:

\revised{
\begin{equation}
    \text{TM}(\mathbf{m_{1}}[j], \dotso, \mathbf{m_{n}}[j]) = \frac{1}{k- 2k_{m}}\sum\limits_{i=k_{m+1}}^{k - k_{m-1}}\mathbf{m_{\omega_{j}(i)}}[j]
\end{equation}
}



\section{Vulnerabilities of CC and designing strong but imperceptible attack} \label{sec:core}
In this section, we identify the main limitations of the CC defence mechanism and underline certain aspects to design strong but imperceptible attacks.
\subsection{Relocation of the attack}
Existing Byzantine attacks set $\bar{\mathbf{m}}_{t}=\frac{1}{k_{b}} \sum\limits_{i\in\mathcal{K}_{b}} \boldsymbol{m}_{i}$ as a reference to design the attack. Hence, by clipping each individual update according to the previous update direction $\mathbf{\tilde{m}}_{t-1}$, it is possible to reduce the impact of a poisoning attack. Let $\boldsymbol{\Delta}_{i,t}$, $i\in\mathcal{K}_{m}$, denote the attack vector of Byzantine client $i$ at time $t$ to the mean benign update $\bar{\mathbf{m}}_{t}$.

 In the CC framework, the benign local momentum and global momentum evolve as follows:
 
\begin{equation}
\centering
\mathbf{m}_{i,t}=\beta \mathbf{m}_{i,t-1} + (1-\beta) \mathbf{g}_{i,t}, \quad i \in \mathcal{K}_{b},
\end{equation}

\revised{
\begin{equation}
\bar{\mathbf{m}}_{t}= \sum_{i\in \mathcal{K}_{b}}\left(\beta \mathbf{m}_{i,t-1} + (1-\beta) \mathbf{g}_{i,t}\right) = \beta \bar{\mathbf{m}}_{t-1} +(1-\beta)\sum_{i\in \mathcal{K}_{b}}\mathbf{g}_{i,t}.
\end{equation}
}
Further, let $\tilde{\boldsymbol{\Delta}}_{t}$ denote the distance between the reference point at $t-1$, $\tilde{\mathbf{m}}_{t-1}$, and the benign momentum at time $t$, $\bar{\mathbf{m}}_{t}$, i.e.,
\begin{equation}
\tilde{\boldsymbol{\Delta}}_{t} = \bar{\mathbf{m}}_{t}-\tilde{\mathbf{m}}_{t-1}.
\end{equation}
Existing attacks often target the benign momentum $\bar{\mathbf{m}}_{t}$; hence, the poisoned updates of the Byzantine client can be written in the following form, with respect to
$\bar{\mathbf{m}}_{t}$ and $\tilde{\mathbf{m}}_{t-1}$:
\begin{equation}
\begin{matrix}
\mathbf{m}_{i,t} = \bar{\mathbf{m}}_{t} + \boldsymbol{\Delta}_{i,t} \ , \quad i \in \mathcal{K}_{m}, 
\\ 
 \ \quad = \tilde{\mathbf{m}}_{t-1} + \underbrace{\tilde{\boldsymbol{\Delta}}_{t} +\boldsymbol{\Delta}_{i,t}}_{\tilde{\boldsymbol{\Delta}}_{i,t}},
\end{matrix}
\end{equation}
where $\tilde{\boldsymbol{\Delta}}_{i,t}$ is the aggregate form of the attack with respect to the reference point of the CC, $\tilde{\mathbf{m}}_{t-1}$. When $|| \tilde{\boldsymbol{\Delta}}_{i,t} ||$ is larger then the radius $\tau$ of CC, the aggregation mechanism of  CC scales $\tilde{\boldsymbol{\Delta}}_{i,t}$
with $\frac{\tau}{\vert\vert\tilde{\boldsymbol{\Delta}}_{i,t}\vert\vert}$.
The scaled version of the attack can be written as a sum of two components: one towards $\bar{\mathbf{m}}_{t}$ and the second in the direction of the attack, i.e., 
\begin{equation}\label{eqn:mismatch}
\tilde{\boldsymbol{\Delta}}_{t}\frac{\tau}{\vert\vert\tilde{\boldsymbol{\Delta}}_{i,t}\vert\vert} + \boldsymbol{\Delta}_{i,t} \frac{\tau}{\vert\vert\tilde{\boldsymbol{\Delta}}_{i,t}\vert\vert}.
\end{equation}
When the CC defense is employed, the clipped poisoned update often includes both a benign update and the attack with the corresponding scaling factor in (\ref{eqn:mismatch}). We emphasize that the strength of the attack is directly related to the impact of $\boldsymbol{\Delta}_{i,t}$ in $\tilde{\boldsymbol{\Delta}}_{i,t}$, which eventually determines the effective scaling of the pure attack $\boldsymbol{\Delta}_{i,t}$.
Significantly reducing $ \tilde{\boldsymbol{\Delta}}_{t}$ gives the attacker more room to further scale up the perturbation $\boldsymbol{\Delta}_{i,t}$  until $\frac{\tau}{\vert\vert\tilde{\boldsymbol{\Delta}}_{i,t}\vert\vert} =$ 1,  while still avoiding clipping.  
Setting $\boldsymbol{\Delta}_{i,t} >>  \tilde{\boldsymbol{\Delta}}_{t}$ maximizes the effectiveness of $\boldsymbol{\Delta}_{i,t}$ as it can perturb and poison the PS model by staying close to the center of clipping.  

Hence, CC around the previous update direction helps to suppress attacks targeting the current benign update direction. However, this strong aspect of the CC framework also becomes its vulnerability: if the attacker knows the reference point of CC,  it only requires the knowledge of the learning rate, which can be easily predicted, to modify the attack accordingly. In other words, the attack can be generated with respect to the reference point,  $\tilde{\mathbf{{m}}}_{t-1}$, and easily escape clipping.

We refer to this observation as the {\em target reference mismatch}, since the target of the attacker, which is often the benign update, is different from the reference of the defence mechanism. We further argue that CC relies on this mismatch and induces a false sense of security. Later, we numerically show how CC can be easily fooled if the attack is revised accordingly. To this end, we consider the \textit{relocation of the attack}, simply targeting the previous update instead of the current benign update. By doing so, we show that the accuracy under the CC defence significantly drops. We will further show that such a strategy is not only successful against CC but also against other SotA defense mechanisms such as TM \cite{Trimmed_mean} and RFA \cite{RFA}.

\subsection{Angular Invariance}

One of the major drawbacks of CC is the angular invariance against attacks that target the reference point $\tilde{\mathbf{m}}_{t}$. The CC performs a scaling operation by clipping the client's momentum $\mathbf{m}_{i,t}$ that lies beyond a certain radius.
This is achieved by scaling the gap $\boldsymbol{\Delta} =\mathbf{m}_{i,t}-\tilde{\mathbf{m}}_{t-1}$ with a factor $\delta$, which depends only on its norm $\vert\vert\boldsymbol{\Delta}\vert\vert$, and  it operates as an identity function if $\frac{\tau}{\vert\vert\boldsymbol{\Delta}\vert\vert}\geq1$.

Now, consider two vectors $\mathbf{m}_{1} = \tilde{\mathbf{m}}_{t-1} + \boldsymbol{\Delta}_{1}$ and $\mathbf{m}_{2} = \tilde{\mathbf{m}}_{t-1} + \boldsymbol{\Delta}_{2}$, where $\vert\vert\boldsymbol{\Delta}_{1}\vert\vert = \vert\vert\boldsymbol{\Delta}_{2}\vert\vert\leq \tau$, but $\boldsymbol{\Delta}_{1} \neq \boldsymbol{\Delta}_{2}$. $f_{CC}$ treats $\mathbf{m}_{1}$ and $\mathbf{m}_{2}$ in an identical manner; however, their angle with respect to the reference point, $\tilde{\mathbf{m}}_{t-1}$, can be significantly different.

A fundamental question for a fixed norm constraint, $\vert \vert\boldsymbol{\Delta}\vert\vert\leq \tau$, is how to decide on the attack $\boldsymbol{\Delta}$ that has the highest impact? A similar problem is discussed in \cite{IPM}, and from the theoretical analysis of the convergence behaviour, the authors argue that for a given benign momentum $\bar{\mathbf{m}}$, the aggregated momentum $\tilde{\mathbf{m}}$ is successfully poisoned if 
\begin{equation}\label{ipm}
 \langle \bar{\mathbf{m}}, \tilde{\mathbf{m}} \rangle < 0,
\end{equation}
in which case the aggregation framework does not meet the required convergence condition. Accordingly, in \cite{IPM}, the authors propose to use a scaled version of the benign gradient $-\epsilon\bar{\mathbf{m}}$, $0<\epsilon<1$, as a poisoned gradient. However, as highlighted in \cite{CC}, unless there is a sufficient number of Byzantine clients, it is often difficult to ensure (\ref{ipm}). To formally illustrate,  considering the naive averaging strategy as an aggregator, we have
\begin{equation}
\tilde{\mathbf{m}}_{t} = \frac{1}{k} \sum_{i\in\mathcal{K}} \mathbf{m}_{i,t} = \frac{1}{k} \left (\sum_{i\in\mathcal{K}_{b}} \mathbf{m}_{i}+ \underbrace{\sum_{i\in\mathcal{K}_{m}}\mathbf{m}_{i}}_{-k_{m}\delta\bar{\mathbf{m}}_{t}}\right ),
\end{equation}
and we have 
\begin{equation}\label{exp_ipm}
\begin{matrix}
\mathbb{E}\left[\tilde{\mathbf{m}}_{t}\right] = \frac{1}{k}\left(k-k_{m}(1+\delta)\right)\bar{\mathbf{m}},
\\ 
\qquad \ = \left  ( 1 - \frac{k_{m}}{k}(1+\delta) \right ) \bar{\mathbf{m}}_{t}. 
\end{matrix}
\end{equation}

Hence, if  $\frac{k}{k_{m}}<(1+\delta)$, then the expected $\tilde{\mathbf{m}}_{t}$ will be positively aligned with $\bar{\mathbf{m}}$. Nevertheless, as shown in \cite{IPM} (see Theorem 2), when there is certain variation among $\mathbf{m}_{i,t}, i\in\mathcal{K}_{b}$, then IPM can be successful in negatively aligning $\tilde{\mathbf{m}}$ with $\bar{\mathbf{m}}$.
Consequently, under certain conditions on the variation among the true gradients, the IPM attack can be successful. On the other hand, when $k>>k_{m}$, on average $\tilde{\mathbf{m}}_{t}$ is a scaled version of $\bar{\mathbf{m}}_{t}$ with a positive coefficient. To conclude, the impact of the IPM attack is highly dependent on the ratio of malicious clients and how the benign clients' gradients/updates are aligned with the benign update direction, i.e., benign clients with very heterogeneous data. Another drawback of the IPM attack is that, although a scaling parameter is used to hide malicious updates, a defence mechanism that utilizes the angular distance rather than a norm-based distance can easily detect malicious clients.
\subsection{Importance of temporal correlation}
At this point, we revisit another well-known attack strategy called {\em ALIE} to highlight the key notions behind our attack strategy. Contrary to IPM, ALIE does not specify a direction for the attack with respect to the benign update $\bar{\mathbf{m}}_{t}$, but introduces a radius $\tau_{i}$, for each index $i$, so that the poisoned update can not be arbitrarily far away from the benign one, index-wise, i.e.,
\begin{equation}
\vert\bar{\mathbf{m}}_{t}[i] - \mathbf{m}^{attack}_{t}[i]\vert  \leq \tau_{i},
\end{equation}
where $\tau_{i}$ is determined based on the statistics of the $i$th index of the updates of benign clients as well as the number of Byzantines, which is further discussed in Section \ref{attacks}. Contrary to (\ref{exp_ipm}), on average we have
\begin{equation}
\mathbb{E}\left[\tilde{\mathbf{m}}_{t}\right] = \bar{\mathbf{m}}_{t} + k_{m}\boldsymbol{\Delta}_{t},
\end{equation}
where $\boldsymbol{\Delta}_{t}$ is not aligned with $\bar{\mathbf{m}}_{t}$, indeed, it is often orthogonal to $\bar{\mathbf{m}}_{t}$. Overall, the objective is to perturb the update direction as much as possible without being detected as an outlier. To achieve consistent error accumulation, the attacker must employ similarly aligned perturbation vectors during training. Here we note that, to measure the alignment of two vectors a common metric is the {\em cosine similarity}, defined as:
\begin{equation}
\cos(\mathbf{m}_{1},\mathbf{m}_{2}) = \frac{\langle\mathbf{m}_{1},\mathbf{m}_{2}\rangle}{||\mathbf{m}_{1}||_{2}  ||\mathbf{m}_{2}||_{2}}.  
\end{equation}

ALIE is quite effective against TM\cite{Trimmed_mean}, Krum \cite{Krum}, and Bulyan \cite{Bulyan} defense mechanisms \cite{ALIE}. Apart from being statistically less visible, ALIE mostly gains from accumulating the perturbations over time.. We argue that the enabling factor behind the accumulation is the correlation of consecutive attacks; in other words, attack vectors are positively aligned over time, i.e., $\cos(\boldsymbol{\Delta}_{t},\boldsymbol{\Delta}_{t-1})\approx1$. We emphasize that, even though the attack is formed independently from the benign gradients without specifying a particular direction; since it directly utilizes the statistics of the benign client updates which often vary slowly during training, the adversarial perturbations end up being highly correlated over time.

\begin{table}[t]
\centering
\caption{Cosine similarity for the perturbation values of the ALIE attack on IID CIFAR-10 dataset. {The average cosine similarity is reported over 100 epochs (6250 communication rounds).}}
\label{tab:cosine_sim}
\begin{tabular}{@{}lccc@{}}
\toprule
   & $\beta=0$ & $\beta=0.9$ & $\beta=0.99$ \\ \midrule
$cos(\boldsymbol{\Delta}_{t-1},\boldsymbol{\Delta}_{t})$ & 0.94 & 0.995 & 0.999 \\\bottomrule
\end{tabular}
\end{table}

\begin{figure}[]
    \centering
    \includegraphics[width=.35\textwidth]{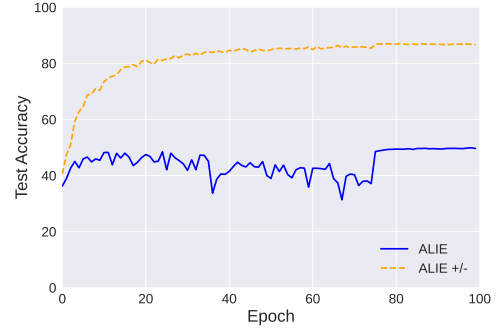}
    \caption{ALIE attack on TM aggregation, ALIE (blue) refers to the standard ALIE, while  ALIE $+/-$ (orange) is the version of  ALIE where the sign of the perturbation alternates at each iteration.}
    \label{fig:alie_alt}
\end{figure}

We argue that the variance $\bar{\boldsymbol{\sigma}}_{t}$ from benign clients $\mathcal{K}_{b}$, thus the direction of the attack vector, does not vary significantly during training. In Table \ref{tab:cosine_sim}, we demonstrate that  $\boldsymbol{\Delta}_{t}$ of ALIE is always aligned positively, which verifies our initial argument that the perturbation $z^{max} \bar{\boldsymbol{\sigma}}_{t}$ used by ALIE attack is consistent over iterations in terms of its direction. This temporal correlation leads to a stronger accumulation and enhances the strength of the attack. Now, to better visualise the impact of temporal accumulation, we consider a scenario where the sign of $\boldsymbol{\Delta}_{t}$ is alternated over consecutive communication rounds;
\begin{equation}\label{eq:altsign}
    \boldsymbol{\Delta}_{t} = \left\{ \begin{array}{cl}
z^{max} \mathbf{\boldsymbol{\bar{\sigma}}}_{t}, & \text{if } \ t \mod 2 =0  \\
-z^{max} \mathbf{\boldsymbol{\bar{\sigma}}}_{t}, & \text{if } \ t \mod 2 =1 
\end{array} \right\}.
\end{equation}
We observe that when the sign alternates, ALIE is unable to accumulate error throughout training, which leads to normal convergence. We illustrate the convergence behaviours of ALIE and ALIE with alternating sign of $\boldsymbol{\Delta}$ in Fig. \ref{fig:alie_alt} against the TM aggregator, which is an aggregator that is known not to be robust against ALIE\cite{ALIE}. This comparison verifies our argument that the ALIE's strength mainly comes from the accumulation of attacks over iterations due to the temporal alignment.

\revised{
\subsection{Structuring the attack}
 Based on the discussions above, for a given radius budget $\left \| \mathbf{\Delta}_{t} \right \| \leq r $, we identify two main design criteria for our attack:
\begin{itemize}
    \item Keeping attack positively aligned over time; $\max_{\left \| \mathbf{\Delta}_{t} \right \|, \left \| \mathbf{\Delta}_{t-1} \right \| \leq r} \cos(\boldsymbol{\Delta}_{t},\boldsymbol{\Delta}_{t-1})$ to maximize the  perturbation accumulated over time.
    \item Keeping attack orthogonal to the reference update direction, i.e., $cos(\boldsymbol{\Delta}_{t},\tilde{\mathbf{m}}_{t-1})\approx0$, which is sufficient to derail the global model, thanks temporal accumulation. On the other hand, unlike IPM, where the poisoned model update is a directly scaled version of the benign update in the opposite direction, i.e., $\alpha\bar{\mathbf{m}}_{t}$, the attack with orthogonal perturbations is less visible in terms of the angular variation. 
\end{itemize}
}

\indent We remark that, as the index-wise variation among the parameters of the benign model updates, $\mathbf{m}_{i,t}$, $i\in \mathcal{K}_{b}$, increases, it is harder to spot Byzantines as statistical outliers. Hence, by utilizing local momentum as a local model update, it is possible to minimize the variation and enhance the robustness against Byzantine clients \cite{CC,Byz.momentum}. On the other hand, such temporal consistency imposed by the momentum also helps the attacker to satisfy the first and second conditions simultaneously since the use of momentum imposes temporal correlation among the reference updates. To be more precise, having temporally correlated reference updates and forming attacks to be orthogonal to the reference updates, as described in the second criteria, induces aligned attacks over time, which implies that the first criteria, which emphasizes the importance of temporal accumulation, is simply satisfied.\\
\indent Although we specifically refer to orthogonal perturbation as an attack mechanism, one can utilize different angles to form the attack. We introduce the general form of the attack in Section \ref{S:ROP}. We promote the use of orthogonal perturbation since it is both effective and imperceptible at the same time. While it is possible to increase the strength of the attack by modifying the angle, this may hurt the imperceptibility. To illustrate this, we consider the following scenario, in which the attack is formed as orthogonal perturbations, and CC is used as a robust aggregator mechanism. We measure the cosine similarity between the both poisoned, and benign consensus updates and the
reference model, provided in Table \ref{tab:cosine_compare}. Interestingly, we observe that, on average, the consensus benign update has a lower cosine similarity with the reference update compared to the poisoned update with orthogonal perturbation, that is, from the aspect of angular variation, the benign update looks more outlier than the poisoned model.

\begin{table}[t]
\centering
\caption{Average cosine similarity between $\tilde{\mathbf{m}}_{t-1}$ and the benign and malicious momentums, respectively. Trained on the IID CIFAR-10 dataset. }
\label{tab:cosine_compare}
\resizebox{\columnwidth}{!}{%
\begin{tabular}{l|ccc|ccc}
 & \multicolumn{1}{l}{} & \multicolumn{1}{l}{\textbf{Benign}} & \multicolumn{1}{l|}{\textbf{}} & \multicolumn{1}{l}{\textbf{}} & \multicolumn{1}{l}{\textbf{Byzantine* (Orthogonal perturbation)}} & \multicolumn{1}{l}{\textbf{}} \\
$\beta$ & 0 & 0.9 & 0.99 & 0 & 0.9 & 0.99 \\ \hline
$\tau$=0.1 & 0 & 0.2 & 0.3 & 0.94 & 0.77 & 0.75 \\ \hline
$\tau$=1 & -0.03 & 0.07 & 0.04 & 0.48 & 0.44 & 0.27 \\ \hline
\end{tabular}%
}
\end{table}

\revised{
Another important design aspect is deciding on the target direction to form and accumulate the orthogonal perturbations. One possibility is to use the reference update, former consensus update, i.e., the $\mathbf{\tilde{m}}_{t-1}$, as the target to form the orthogonal perturbation $\boldsymbol{\Delta}_{t}$, which is quite effective against the CC mechanism. However, when the poisoned model is close to the reference update $\mathbf{\tilde{m}}_{t-1}$, its distance to benign consensus update $\bar{\mathbf{m}}$, particularly index-wise, can be visible to median-based defence mechanisms using $\bar{\mathbf{m}}_{t}$ as a reference to detect outliers. Byzantines can alter the reference point and location of the attack depending on their knowledge of the deployed defences and aggregators to maximize the effectiveness of the generated perturbation. 

In Section \ref{S:ROP} we present a generalized version of the attack where the target can be chosen as any point between $\mathbf{\tilde{m}}_{t-1}$ and $\bar{\mathbf{m}}_{t}$.}

\revised{
\begin{algorithm}[t]
    \small
	\caption{ROP Attack}
 	\textbf{Inputs:}  $\tilde{\mathbf{m}}_{t-1}$, $\bar{\mathbf{m}}_{t}$,
    $\pi$, $\lambda$, $\rho$
	\begin{algorithmic}[1]
	\State$\mathbf{p} \gets \boldsymbol{1}\in \mathbb{R}^{d}$ 
	\If{$t$ == 1}	
		\State $\tilde{\mathbf{m}}_{t-1} \gets \ \bar{\mathbf{m}}_{t}$
	\EndIf
	\State $\hat{\mathbf{m}}_{t} \gets \lambda \tilde{\mathbf{m}}_{t-1} \ + \ (1-\lambda) \bar{\mathbf{m}}_{t} $ \# target point
	\State $\tilde{\mathbf{p}} \gets$ $Proj(\mathbf{p},\mathbf{m}_{t})$
	\State $\hat{\mathbf{p}}\gets \mathbf{p} - \tilde{\mathbf{{p}}}$ \# orthogonal perturbation
        \State $\boldsymbol{\Delta}_{t} \gets \sin(\pi) \frac{\hat{\mathbf{p}}}{||\hat{\mathbf{p}}||} + \cos(\pi)\frac{\hat{\mathbf{m}}_{t}}{\left\| \hat{\mathbf{m}}_{t} \right\|}$ \# direction of the attack
        \State $\mathbf{m}^{attack}_{t} \gets z \boldsymbol{\Delta}_{t} + \rho
        \Tilde{\mathbf{m}}_{t-1} + (1-\rho) \Bar{\mathbf{m}}_{t}$ \# relocation
	\end{algorithmic}
	\label{code:ROP}
\end{algorithm}

\section{Relocated Orthogonal Perturbation (ROP) Attack}\label{S:ROP}
In order to exploit the aforementioned weaknesses of CC, we introduce a modular and scalable time-coupled model poisoning attack, called $relocated \ orthogonal \ perturbation$ (ROP). The proposed attack consists of two main steps; forming an orthogonal perturbation with respect to a target vector and relocating the perturbation, possibly closer to the reference point used by the robust aggregator, in order to avoid norm-based defence mechanisms. First, the attacker picks a point between $\tilde{\mathbf{m}}_{t-1}$ and $\bar{\mathbf{m}}_{t}$ as the target, denoted by $\hat{\mathbf{m}}_{t}$ (Algorithm \ref{code:ROP} line 4) using the $\lambda$ hyper-parameter. We emphasize that by using both $\tilde{\mathbf{m}}_{t-1}$ and $\bar{\mathbf{m}}_{t}$, we induce a certain temporal correlation between $\hat{\mathbf{m}}_{t}$, thanks to $\tilde{\mathbf{m}}_{t-1}$, but also take into account the current model update.\\
\indent Once the attack decides on the target $\hat{\mathbf{m}}_{t}$, an orthogonal vector $\hat{\boldsymbol{\Delta}}_{t}$ is formed by using the vector projection and rejection methods (Algorithm \ref{code:ROP}, lines 5-6). Next, the perturbation is generated as a linear combination of the generated orthogonal vector $\hat{\mathbf{p}}$ and the target $\hat{\mathbf{m}}_{t}$, so that one can play with the angle between the generated perturbation and the target point using the $\pi$ hyper-parameter between [0-360] (Algorithm \ref{code:ROP}, line 7). In the final step, the objective is to relocate the perturbation towards the reference point $\tilde{\mathbf{m}}_{t-1}$ in order to escape norm-based clipping strategies used to sanitize model updates as in CC.\\
\indent We remark that, though CC takes advantage of $\tilde{\mathbf{m}}_{t-1}$ to sanitize local updates before aggregation, once the described attack, successfully avoids sanitation, it acts as a catalyst for poisoning the CC aggregator, since  $\tilde{\mathbf{m}}_{t}$, used as a center for clipping in the next iteration, also becomes poisoned and unreliable. However, if the attacker knows the aggregator and defences deployed by the PS, such as index-wise aggregator like TM, similar to the ALIE, the attacker can relocate the perturbation back to $\bar{\mathbf{m}}_{t}$ using the $\rho$ hyper-parameter (Algorithm \ref{code:ROP} line 8). However, accumulating the perturbation to the $\tilde{\mathbf{m}}_{t-1}$ is almost equally effective according to our simulations. 
}

We carry out a thorough ablation study in our experiments to understand the effect of the attack location, reference point for perturbation, and angle of the perturbation with respect to the reference point in the Appendix \ref{app:parameters}.

\section{Robustness by randomizing the reference point}
In the previous section, we have shown that the predictability of the reference point can be exploited by the Byzantines to reconfigure their attack strategy based on the knowledge of the reference point. In this section, we argue that it is possible to enhance the robustness of the aggregators, particularly CC, by hiding the reference point from the Byzantines. Accordingly, we introduce the idea of inducing randomness in the selection of the reference point, in contrast to the static one used in the original CC framework.

S-CC is inspired by the recently introduced {\em bucketing} strategy \cite{Bucketing,bucket_krum}, which form 'buckets' of clients to reduce the variance before performing CC-based aggregation.

\revised{
To overcome the aforementioned vulnerabilities of the CC and to defend against the proposed model poisoning attack such as ROP, we propose an enhanced version of CC, named sequential CC (S-CC), given in Algorithm \ref{code:SCC}. The main idea of S-CC is to divide the clients into disjoint groups and then perform CC and aggregation sequentially over group, so that the reference point utilized for each group is different and depends on the previous groups, which makes it hard for Byzantines to collude and predict the reference point. The key difference from our proposed bucketing approach is that we employ cosine similarity to sort and cluster the clients based on their similarity to the reference point before applying it to bucketing. Whereas \cite{Bucketing} just randomly distributes clients to buckets and \cite{bucket_krum} employs Nearest Neighbor Mixing, which essentially employs Multi-Krum \cite{Krum} to form the buckets based on their geometric similarities. However, due to the nature of the Krum, it is a considerably slower bucketing approach than our proposed approach and random bucketing \cite{Bucketing}. 
}

\begin{algorithm}[]
	\small
	\caption{Sequential centered clipping (S-CC)}
	\textbf{Inputs:} $\boldsymbol{\tilde{m}_{t-1}}$, $\left\{\mathbf{m}_{i,t}\right\}_{i\in\mathcal{K}},f_{CC}(\cdot),\tau$
	\begin{algorithmic}[1]
            \State Determine number of buckets $R \gets \left \lceil k/n \right \rceil$
		\State Form buckets of size $n$, $\mathcal{C}_{1},\ldots,\mathcal{C}_{R}$ by selecting one client from each cluster w.o. repetition.
		\State Initialize auxiliary reference momentum: $\hat{\mathbf{m}}_{t} = \tilde{\mathbf{m}}_{t-1}$
		\For{$n=1,\ldots,R$} 
		\State $\bar{\mathbf{m}}_{t}=\frac{1}{\left \| \mathcal{C}_{n} \right \|}\sum_{i\in\mathcal{C}_{n}}\tilde{\mathbf{m}}_{i,t}$
		\State $\bar{\mathbf{m}}_{t} = f_{CC}(\bar{\mathbf{m}}_{i,t}\vert \hat{\mathbf{m}}_{t},\tau)$
		\State $\hat{\mathbf{m}}_{t} = \bar{\mathbf{m}}_{t} $
		\EndFor
		\State $\tilde{\mathbf{m}}_{t} = \hat{\mathbf{m}}_{t}$
	\end{algorithmic}
	\label{code:SCC}
\end{algorithm}

\subsection{Sequential Centered Clipping (S-CC)}
The key motivation behind S-CC is to introduce randomness into the CC framework. This is achieved by grouping the clients into buckets of size $n$, while performing CC in an iterative manner, instead of a single aggregation with a fixed reference point. S-CC performs the aggregation in $\left \lceil k/n \right \rceil$ consecutive phases while dynamically updating the reference point at the end of each phase to induce certain randomness to prevent the collusion of the Byzantine clients. Here $n$ is a static hyper-parameter chosen by the PS before the training starts (Alg. \ref{code:SCC}, line 1). At the beginning of each aggregation step, the clients are sorted by the PS based on the cosine similarity between their momentums and the reference point and grouped into $n$ clusters (line 3). PS randomly selects one client from each cluster without repetition to form a bucket, and performs CC to update the momentum using the average momentum of the bucket (Alg. \ref{code:SCC}, lines 4-6). After each average bucket is clipped, the reference point of S-CC is also updated (Alg. \ref{code:SCC}, line 7). Therefore, unlike in the standard CC, the momentum is updated $\left \lceil k/n \right \rceil$ times sequentially while also reducing the total number of clipping operations. \revised{One weakness of the S-CC aggregator is the decrease of robustness when there is a presence of Byzantine in multiple clusters, which is more apparent in attacks like ALIE, where Byzantines target the $\bar{\mathbf{m}}_{t}$. In such attacks, cosine similarity between the reference of the S-CC and the momentum of the Byzantine can vary depending on the variance among the benign clients, which can result in multiple clusters with Byzantine presence in them. Consequently, some buckets may contain more than one Byzantine, resulting in partial collusion. To prevent this, we recommend using S-CC with local momentum employed, more specifically $\beta=0.9$, to ensure lower variance among the clients, which results in higher cosine-similarity between $\bar{\mathbf{m}}_{t}$ and $\tilde{\mathbf{m}}_{t-1}$.}

\revised{Finally, we want to emphasize that although the proposed S-CC strategy has certain similarities with the iterative CC strategy introduced in \cite{CC} and the bucketing strategy introduced in \cite{Bucketing}, the way we utilize clustering and multi-phase aggregation methods significantly differs from those and aims to address a particular vulnerability of the CC mechanism.  Iterative CC is introduced in \cite{CC} as a refined version of CC, where clipping is performed in a successive manner to eventually converge to a true update over a certain number of iterations in order to achieve a better sensitivity for the clipping by updating reference point in each iteration and thus momentum's of the clients are clipped multiple iterations in a single aggregation step. The proposed S-CC strategy differs from iterative CC in two main design aspects: First, unlike the iterative CC, not all the clients are present at each iteration of S-CC. Performing CC iteratively over groups induces randomness in the reference points. This aims to prevent Byzantines from accurately predicting reference points and avoid clipping by relocating the perturbation accordingly. Second, by utilizing a systematic grouping strategy using cosine similarity based clustering and then bucketing, which is not considered in iterative CC,  S-CC aims to minimize the potential collusion among the Byzantines, since a different reference point is employed for each group. }

\section{Experiments}
\subsection{Datasets and model architectures}
\begin{figure}[!t]
\centering
\subfloat[]{\includegraphics[width=0.24\textwidth]{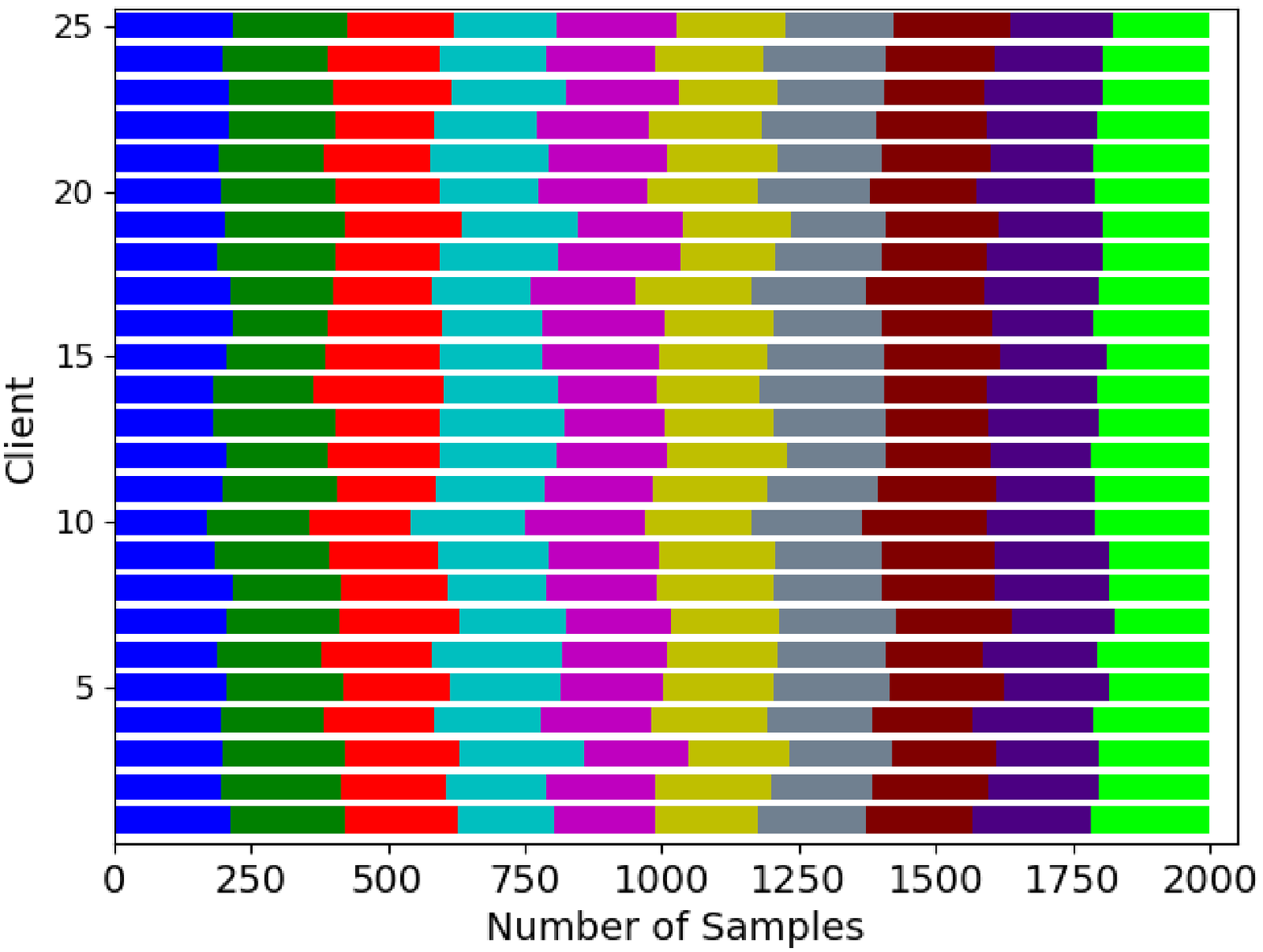}%
\label{IID}}
\hfil
\subfloat[]{\includegraphics[width=0.24\textwidth]{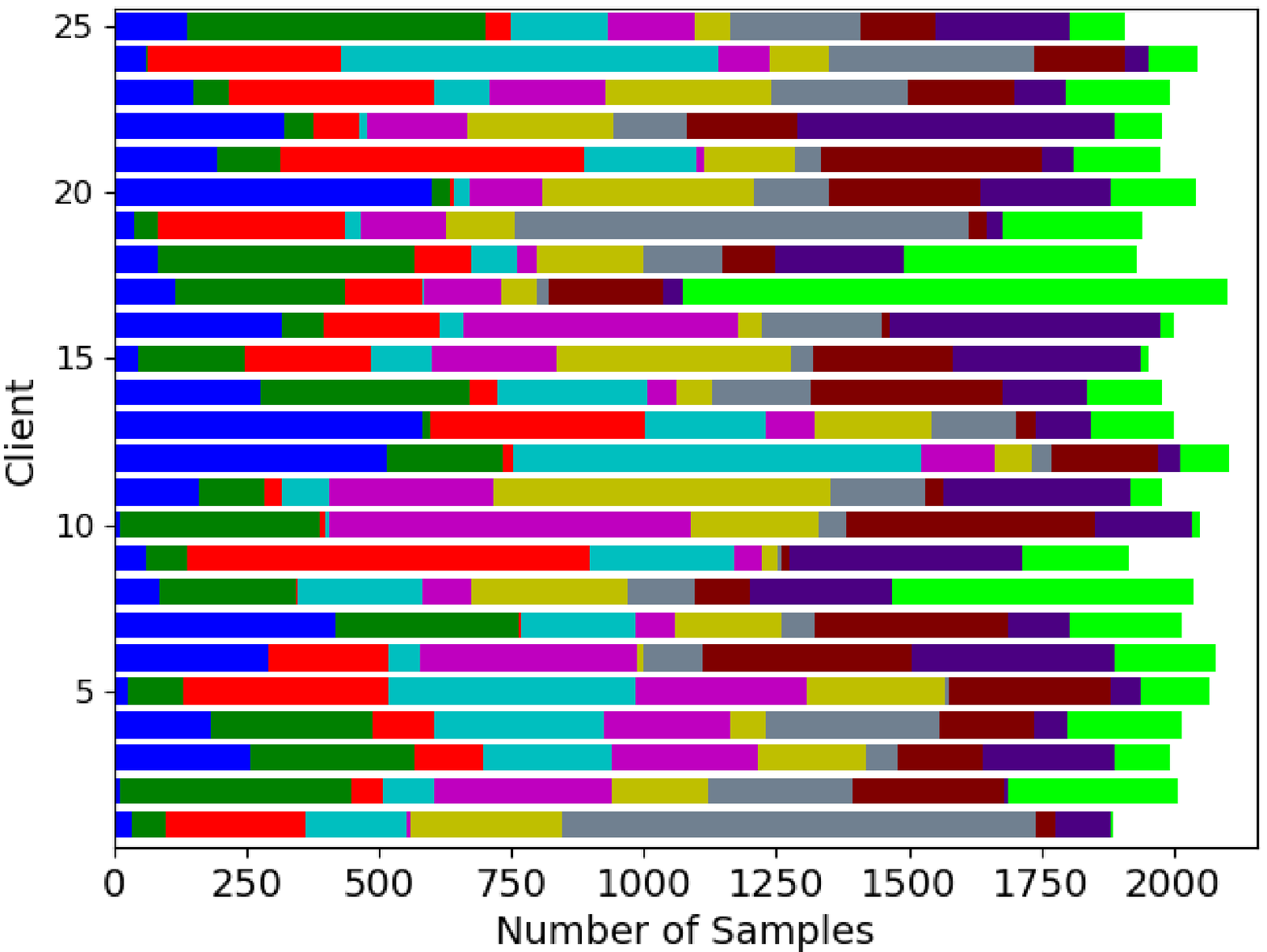}%
\label{fig:non-IID}}
\caption{Visualization of (a) IID and (b) non-IID distribution of 50.000 training samples, 5000 samples per each class, among 25 clients. Each color represents a different class.}
\label{fig:dirichlet_ex}
\end{figure}

We consider two scenarios, where we distribute the data among the clients in IID and non-IID manners, respectively. In the former scenario, we distribute the classes homogeneously among the clients and an equal number of training samples are allocated to each client. In the non-IID scenario, we partition the whole dataset according to the Dirichlet distribution \cite{dirichlet}, where the local dataset at each client $\mathcal{D}_{i}$ has heterogeneous class samples and the total number of samples in each local dataset may vary across the clients. Similar to \cite{B2DB}, we use Dir($\alpha=1$) for the Non-IID scenario, which is more in line with realistic  data distributions among distributed clients. Data distributions among 25 clients for the IID and non-IID scenarios are illustrated in Fig. \ref{fig:dirichlet_ex}.

For the grayscale image classification task, we consider MNIST\cite{mnist} and FMNIST\cite{fmnist} datasets and train with a 2-layer convolutional neural network (CNN). Due to the relative simplicity of the MNIST dataset, we only consider the non-IID scenario. For the RGB image classification task, we consider CIFAR-10 and CIFAR-100 datasets \cite{cifar10}, and train ResNet-20 and ResNet-9 architectures, respectively. However, since CIFAR-100 is a relatively challenging dataset for the image classification task, we only consider the IID scenario for this task.

\subsection{Adversary and FL model}
We consider synchronous FL with a total of $k$=25 clients. We assume \% 20 of the clients i.e. $k_{m}$=5 of these are malicious Byzantine clients which is inline with the \cite{CC}. More simulations on different number of client $k$ and Byzantines $k_{m}$ is available in Appendix \ref{sec:cli-and-byz} which we further demonstrate the effectiveness of our proposed attack. For training, we follow a similar setup as in \cite{CC}, where we train our neural networks for 100 epochs with local batch size of 32 and an initial learning rate of $\eta=0.1$, which is reduced at epoch 75 by a factor of $0.1$. For all the simulations, each local client considers the cross entropy loss to compute gradients.

For simulations with CC, we set its radius to $\tau=[0.1,1]$  and number of  clipping iterations to $l=1$. We observe that CC is more prone to divergence when $\tau \geq$ 10, and since the authors in \cite{CC}  argue that every $\tau$ is equally robust, we only consider these twp $\tau$ values. For the proposed S-CC aggregator, we consider a cluster size of $N=3$, and used the average of the clusters for all the simulations.

For the omniscient model poisoning attacks, we consider ALIE, IPM and ROP. In ROP, we experimentally set $z=$ 1,  $\lambda=$ 0.9, $\rho=$ 1. The impact of $z$, $\lambda$ and $\rho$ on the convergence are further discussed in the appendix \ref{app:parameters}. For IPM, we use  $\epsilon=0.2$. For non-omniscient attacks, we consider the bit-flip and label-flip \cite{Label_flip1,label_flip2} attacks. In the bit-flip attack, Byzantine clients flip the signs of their own gradient values, whereas in the label-flip attack, Byzantine clients flip the label of a sample by subtracting it from the total number of image classes in the dataset. 

\subsection{Numerical Results} \label{sec:num_results}
In this section, we empirically demonstrate the effectiveness of our proposed ROP attack against robust aggregators, particularly CC with local momentum. For simulations, we compare our proposed ROP with omniscient ALIE \cite{ALIE}, IPM \cite{IPM}, and non-omniscient bit-flip and label-flip attacks. In our results, we also report the baseline accuracy of the aggregators, where all the participating clients are benign, i.e. $k_{m}=0$. All numerical results are average of 5 runs with different seeds, we report the mean and standard deviation in Table \ref{tab:All_results}.

\begin{figure*}[h]
    \centering
    \includegraphics[width=.85\textwidth]{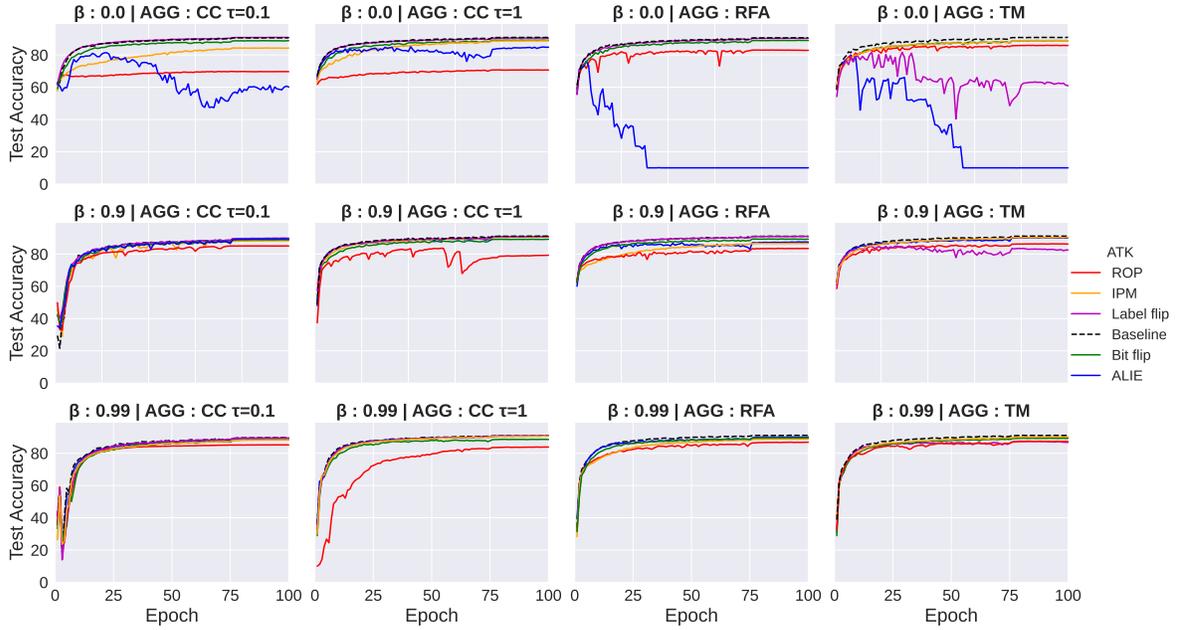}
    \caption{FMNIST IID test accuracy on various aggregation methods.}
    \label{fig:fmnist-iid}
\end{figure*}

In Fig \ref{fig:fmnist-iid}, we present the effect of  ROP and other attacks on the IID FMNIST dataset trained on a 2-layer CNN architecture. Due to the relative simplicity of the dataset and robust CNN architecture, most of the aggregators are capable of fending off the Byzantines in this scenario. Only on $\beta=0$, ALIE is able to utilize increased variance among the clients, therefore, resulting in the divergence of the RFA and TM aggregators. CC is more robust compared to the other aggregators, however, ROP can still hinder its convergence, especially when local momentum is employed with $\beta=0.99$, reducing the test accuracy by 5\% and 7\% for CC with $\tau=0.1$ and $\tau=1$, respectively, while also significantly reducing the convergence speed. g the convergence speed. 

\begin{figure*}[h]
    \centering
    \includegraphics[width=.85\textwidth]{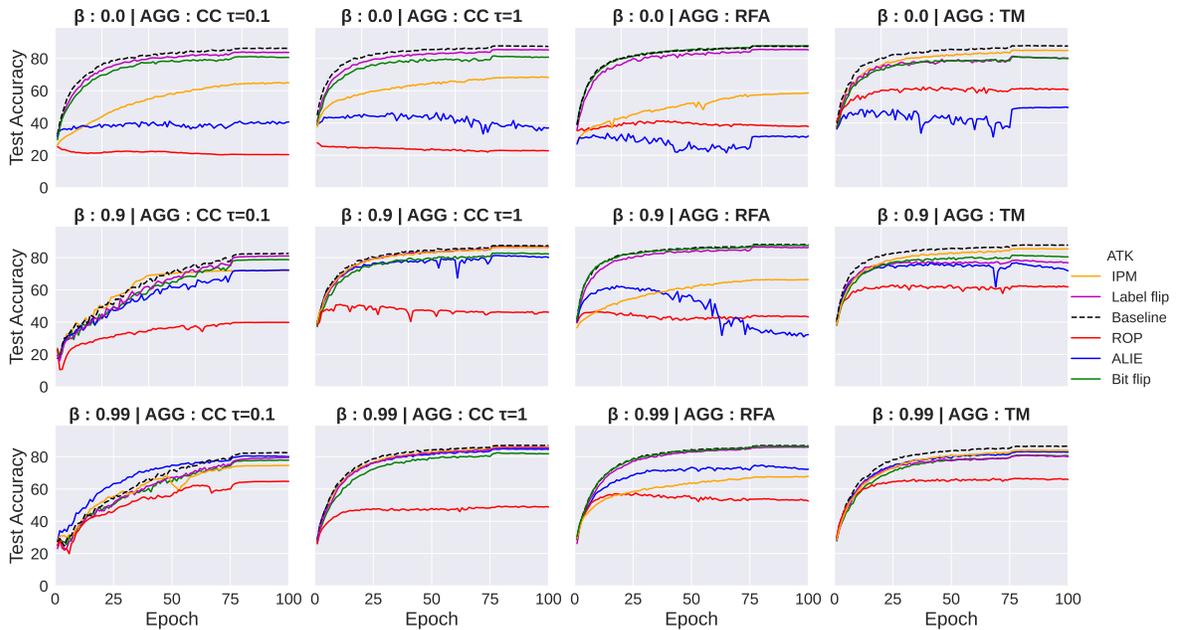}
    \caption{Cifar10 IID test accuracy on various aggregation methods.}
    \label{fig:cif10-iid}
\end{figure*}

In Fig \ref{fig:cif10-iid}, we show the convergence behaviour of the ResNet-20 architecture trained on the IID CIFAR-10 dataset. We can observe the effect of time-coupled omniscient attacks like ALIE and ROP. For $\beta=0$, ALIE can benefit from the increased variance, while ROP can still surpass all the attacks against the CC aggregator. Similar to the results reported in \cite{CC}, high local momentum benefits all the aggregators; however, ROP is able to achieve low test accuracy by nearly reducing  it by 35\% in the case of  CC with $\beta=0.99$, which is the best aggregator setup as advised by the  authors of CC \cite{CC}.

\begin{figure*}[h]
    \centering
    \includegraphics[width=.8\textwidth]{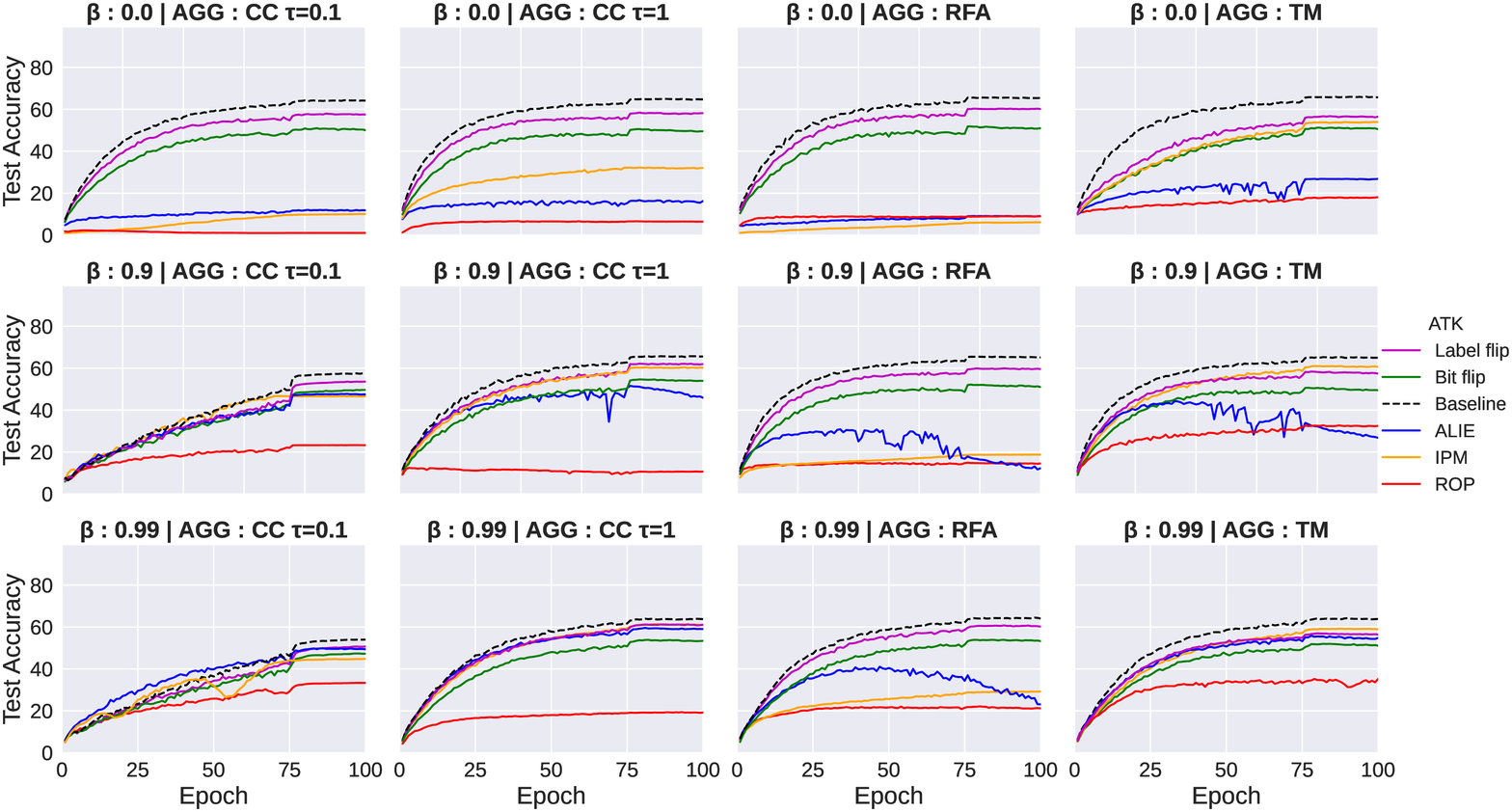}
    \caption{CIFAR100 IID test accuracy on various aggregation methods.}
    \label{fig:cif100}
\end{figure*}

For a more challenging setup of aggregators with IID data distribution, we consider the CIFAR-100 image classification task trained on a larger ResNet-9 architecture. The results are given in Fig \ref{fig:cif100}. For CIFAR-100, the RFA aggregator struggles to defend against ROP, ALIE, and IPM on every $\beta$ parameter, while TM can only converge on $\beta=0.99$ except of the ROP attack. Against the CC aggregator, without employing a local momentum, time-coupled attacks are capable of derailing the convergence at a certain point, while ROP is capable of hindering the learning process from the start of the training. Similar to the CIFAR-10 results, only ROP can prevent convergence consistently when local momentum is employed, where it can reduce the test accuracy by 40\% for $\beta=0.99$.

\begin{figure*}
    \centering
    \includegraphics[width=.8\textwidth]{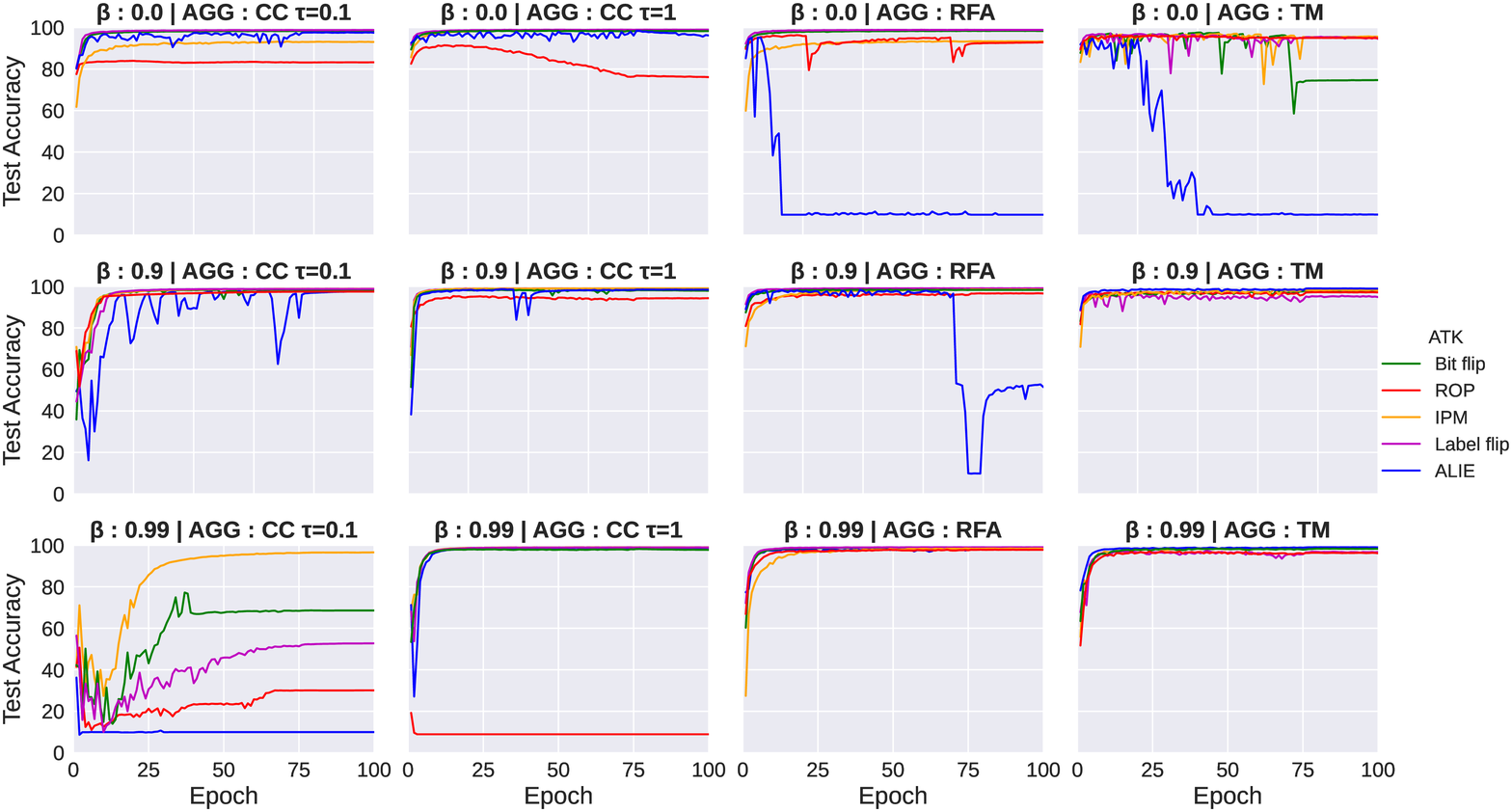}
    \caption{Mnist Non-IID test accuracy on various aggregation methods.}
    \label{fig:mnist-dir}
\end{figure*}

In Fig. \ref{fig:mnist-dir} we show the convergence of aggregators on the MNIST dataset distributed in a non-IID manner. Due to the simplicity of the MNIST dataset, all aggregators can provide normal convergence while employing local momentum. On $\beta=0$, ROP can reduce the test accuracy by 20-25 \%, while ALIE can diverge the RFA and TM aggregators. In the case of a CC aggregator with $\tau=0.1$ and $\beta=0.99$, the PS model cannot achieve the baseline level of the other aggregators even if there is no attacker. We obverse that with a large local momentum and non-IID data distribution, CC with low $\tau$ fails to converge, which contradicts the claim of the authors in \cite{CC} about CC being equally robust for all $\tau$ parameters.

\begin{figure*}
    \centering
    \includegraphics[width=.8\textwidth]{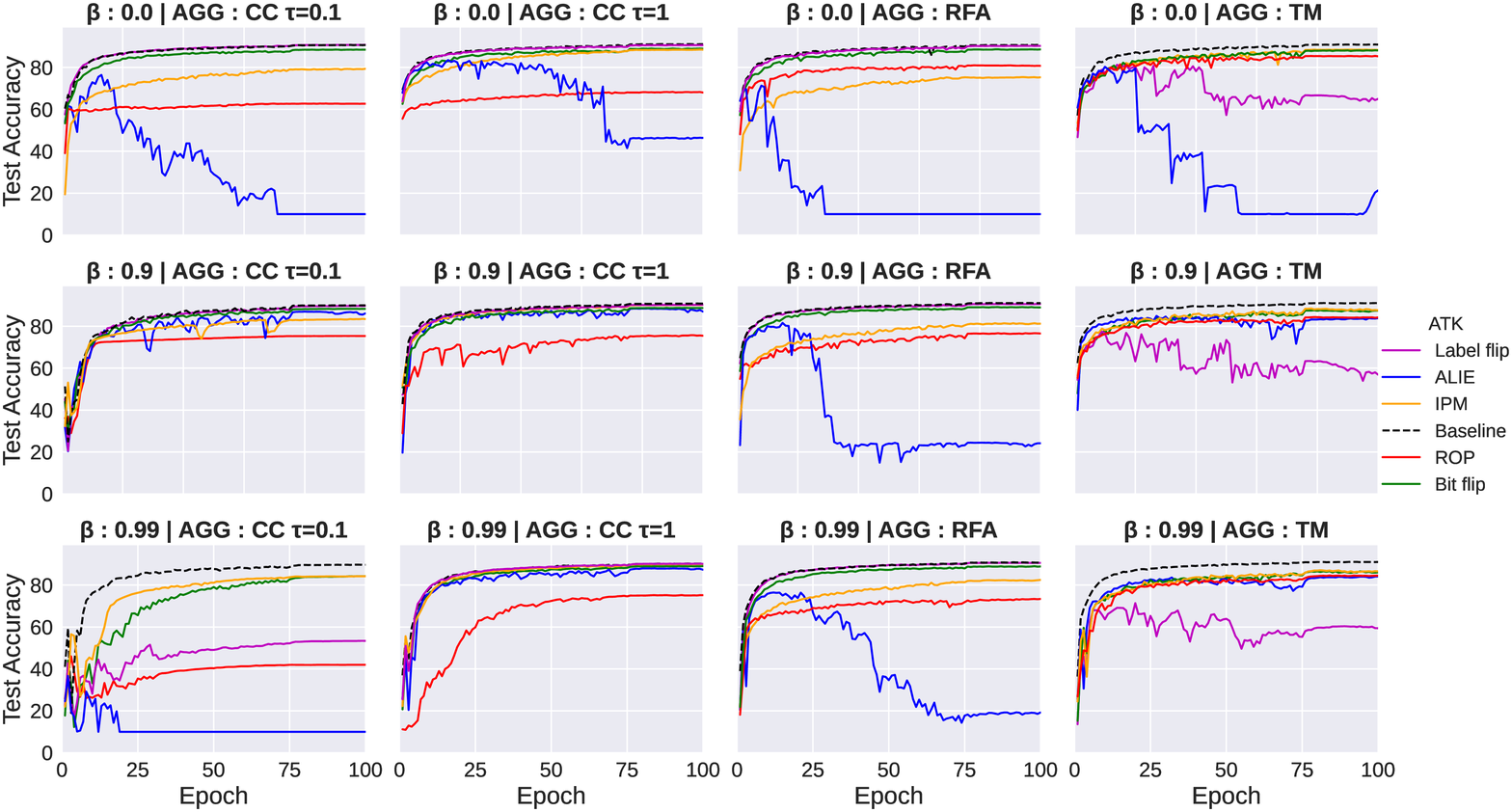}
    \caption{FMNIST Non-IID test accuracy on various aggregation methods.}
    \label{fig:fmnist-dir}
\end{figure*}

In Fig. \ref{fig:fmnist-dir}, we show the convergence behavior for the FMNIST dataset distributed in a non-IID manner trained on the same 2-layer CNN architecture. Although FMNIST is a single channel black and white dataset similar to the MNIST, it is considered a more complex dataset which is especially challenging when the dataset's distribution is very skewed among the clients. In this simulation, ALIE can able to diverge the RFA aggregator while ROP and IPM can able to yield sub-optimal convergence. Interestingly on the TM aggregator, the non-omniscient label-flip attack is the most successful when local momentum is employed. Overall, CC at $\tau=1$ with local momentum is the most successful aggregator however, ROP can still able to slow down convergence while also lowering the baseline accuracy by almost 20\%.

\begin{figure*}
    \centering
    \includegraphics[width=.8\textwidth]{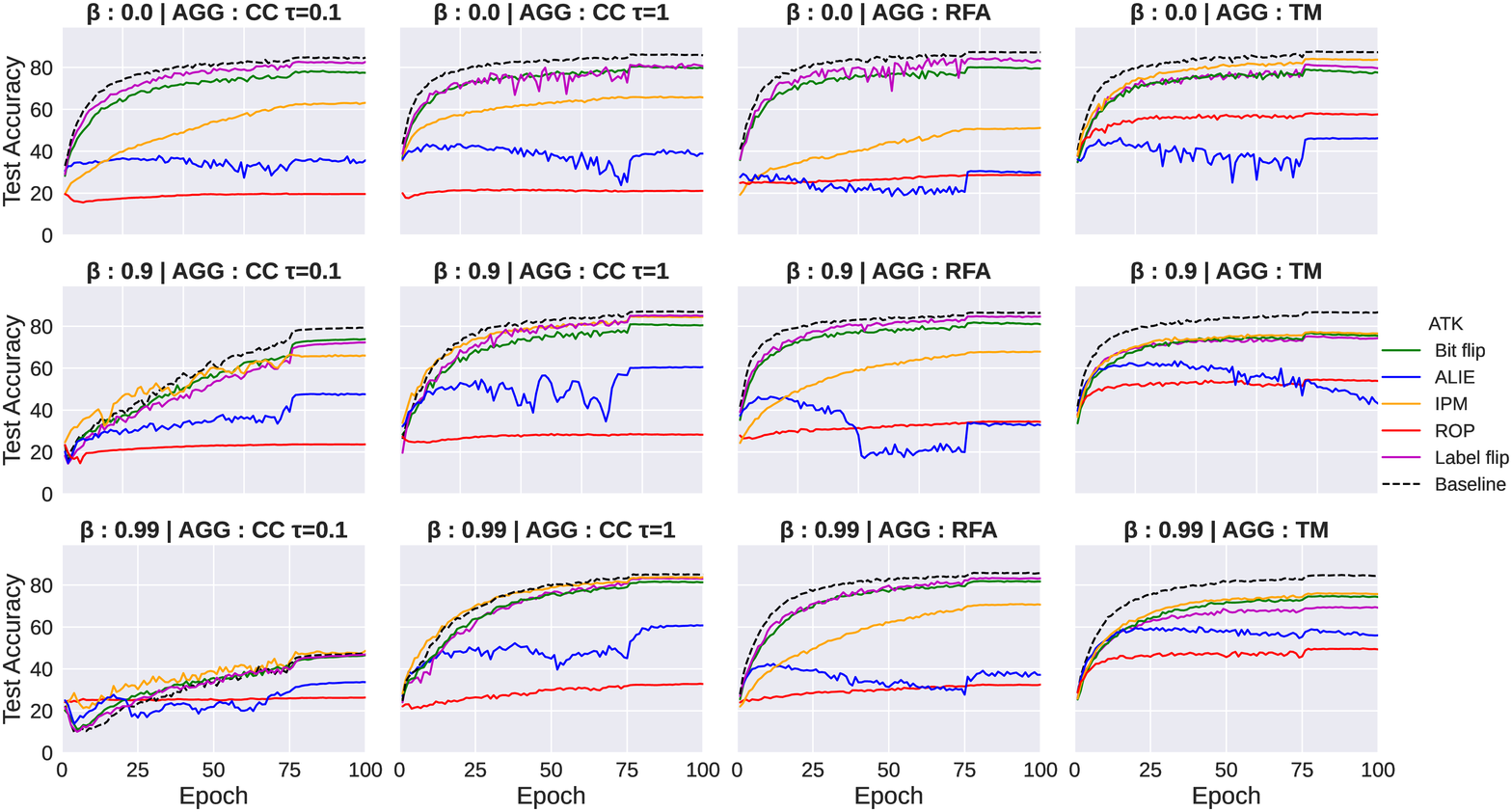}
    \caption{Cifar10 Non-IID test accuracy on various aggregation methods.}
    \label{fig:cif10-dir}
\end{figure*}

In Fig. \ref{fig:cif10-dir} we challenge the robust aggregators on the ResNet-20 architecture trained on the CIFAR-10 image classification task with non-IID data distribution. In terms of baseline accuracy i.e., without any attack, CC with $\tau=0.1$ with local momentum fails to converge, while its IID counterpart and other aggregators can provide normal convergence when there is no Byzantine client. Against all the aggregators and $\beta$ values, ROP is capable of preventing the convergence from the start of the training meanwhile, with the benefit of the increased variance due to the non-IID data distribution, ALIE is also a strong competitor to ROP especially when local momentum is not employed however alie assume to know variance thus increased variance greatly helps meanwhile ROP does not assume know the variance yet still surpass the ALIE. In this scenario, TM with $\beta=0.99$ surpasses the CC aggregator in terms of robustness, although ROP can still reduce the test accuracy by 35\%.

Overall, we show that ROP overcomes the robustness of CC regardless of the $\tau$ and $\beta$ parameters, and it is the most successful attack in the IID data distribution scenarios, especially when local momentum is employed. Other attacks fail to prevent the convergence of CC aggregator with $\tau=1$. In the Non-IID distribution scenario, ALIE is also a successful attack as ROP. This is mainly due to the increased variance among the clients which provides ALIE more room to scale up its perturbation while ROP does not assume to know variance and, uses the same amount of perturbation regardless of the variance among the benign clients. Although in its default configuration, ROP targets the CC, we use the same configuration on a median-based defence TM and a norm-based defence RFA, which both consider statistical calculations using only $\mathbf{{m}}_{i,t}$ from the clients. We observe that ROP can still compete with and even surpass ALIE. Further analyzing the robust  aggregators shows that CC with radius $\tau=0.1$ is not robust when the data distribution is non-IID and local momentum is employed, failing to converge even when there are no Byzantine clients. This can result from a combination of the low gradient scaling (1-$\beta$) and very small CC radius $\tau$, which lead to the PS converging very slowly or not learning properly from the clients.

\begin{figure*}
    \centering
    \includegraphics[width=.8\textwidth]{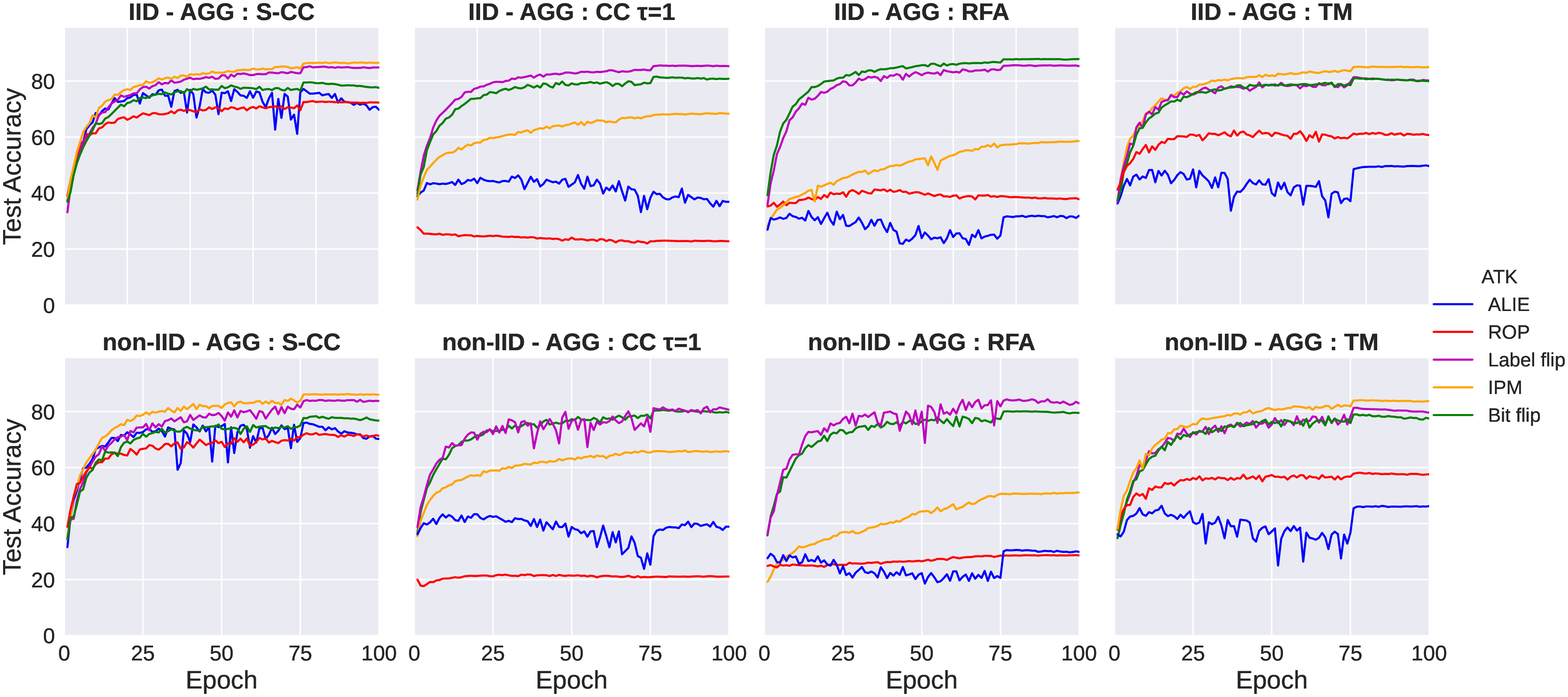}
    \caption{CIFAR-10 $\beta=$0 test accuracy on proposed Sequential CC and other aggregators.}
    \label{fig:cif10-b0-def}
\end{figure*}

\begin{figure*}
    \centering
    \includegraphics[width=.8\textwidth]{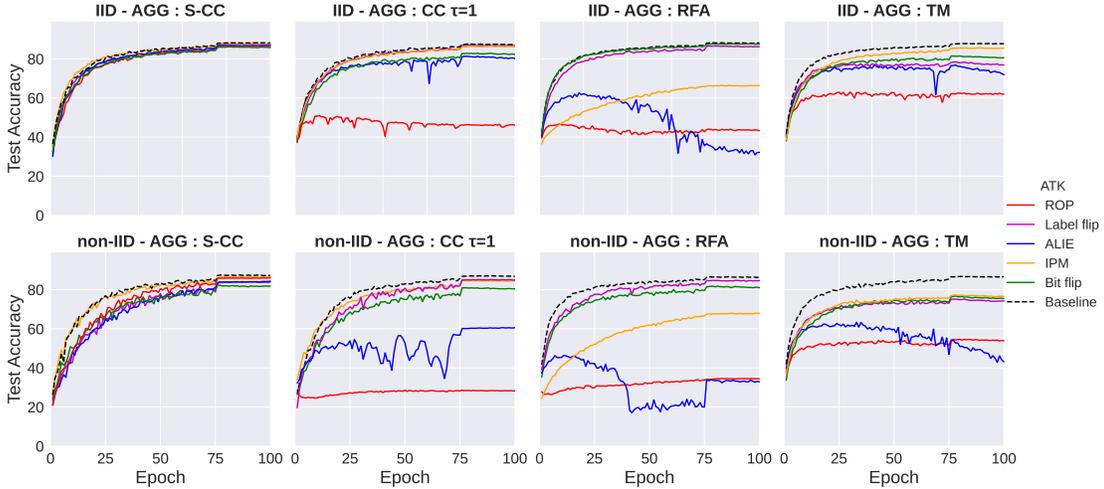}
    \caption{CIFAR-10 $\beta=$0.9 test accuracy on proposed Sequential-CC and other aggregators.}
    \label{fig:cif10-b9-def}
\end{figure*}

In Fig \ref{fig:cif10-b0-def} and \ref{fig:cif10-b9-def} we show that the S-CC aggregator can capable of increasing test accuracy for all attack types regardless of the data distribution and $\beta$ value. Furthermore, for $\beta=0$ in Fig \ref{fig:cif10-b0-def}, the S-CC aggregator is equally robust for both IID and non-IID data distribution, which is not the case for every other aggregator that we consider for our simulations. Furthermore, by enabling the double clipping S-CC aggregator can achieve baseline accuracy for ROP attack on any $\beta$ and data distributions; however, the rest of the attack schemes will result in test accuracy performances that are similar to the CC. However, we consider an aggregator that is robust to all model poisoning attacks while also keeping the same computational complexity of the CC scheme thus we recommend the S-CC aggregator with local momentum $\beta=0.9$ where it can achieve near baseline accuracy as seen in Fig. \ref{fig:cif10-b9-def}.

\section{Discussion and Conclusions}
The CC framework in \cite{CC} proposed to utilize the acceleration technique of momentum SGD to increase the robustness of the FL framework against Byzantine attacks. The advantage of local momentum is two-fold: First, it decreases the variance of the client updates, statistically reducing the available space for Byzantine attacks. Second, the consensus momentum from the previous iteration can be used to neutralize Byzantine attacks by taking it as a reference point and performing clipping accordingly. In this work, we showed that it is possible to circumvent the CC defence by redesigning existing attack mechanisms, such as ALIE and IPM, and the revised attacks can succeed against CC as well as other known defence mechanisms.

We highlighted two important aspects of the CC framework. First, it relies on the assumption that Byzantine attacks target benign updates. Hence, the CC mechanism considers the previous consensus update as the reference for clipping. We argue that CC benefits from the mismatch between the assumed target and the reference. Accordingly, it is possible to circumvent the CC defences by matching the target to the reference. Second, CC is an angle-invariant operation; that is the angle between the reference and the candidate vectors does not have an impact on the clipping operation. Based on these observations, we introduced a novel attack mechanism called ROP to circumvent the CC defences. We have shown through extensive numerical experiments that, ROP can successfully poison the model even when CC is deployed at PS as a defence mechanism. We have also shown that ROP is also effective against other well-known defence mechanisms, including TM and RFA as well. We further proposed a potential defence mechanism against ROP, called S-CC. By introducing randomness into the clipping process, bucketing the clients, and dynamically choosing a reference point for each bucket, the proposed S-CC mechanism offers complete robustness against ROP and drastically improves the test accuracy in the presence of many other known attacks.






\clearpage
\bibliographystyle{IEEEtranS}
\bibliography{IEEEabrv,main.bib}
%

\newpage

\appendix

\subsection{Training loss analysis}

\begin{figure*}[h]
    \centering
    \includegraphics[width=1\textwidth]{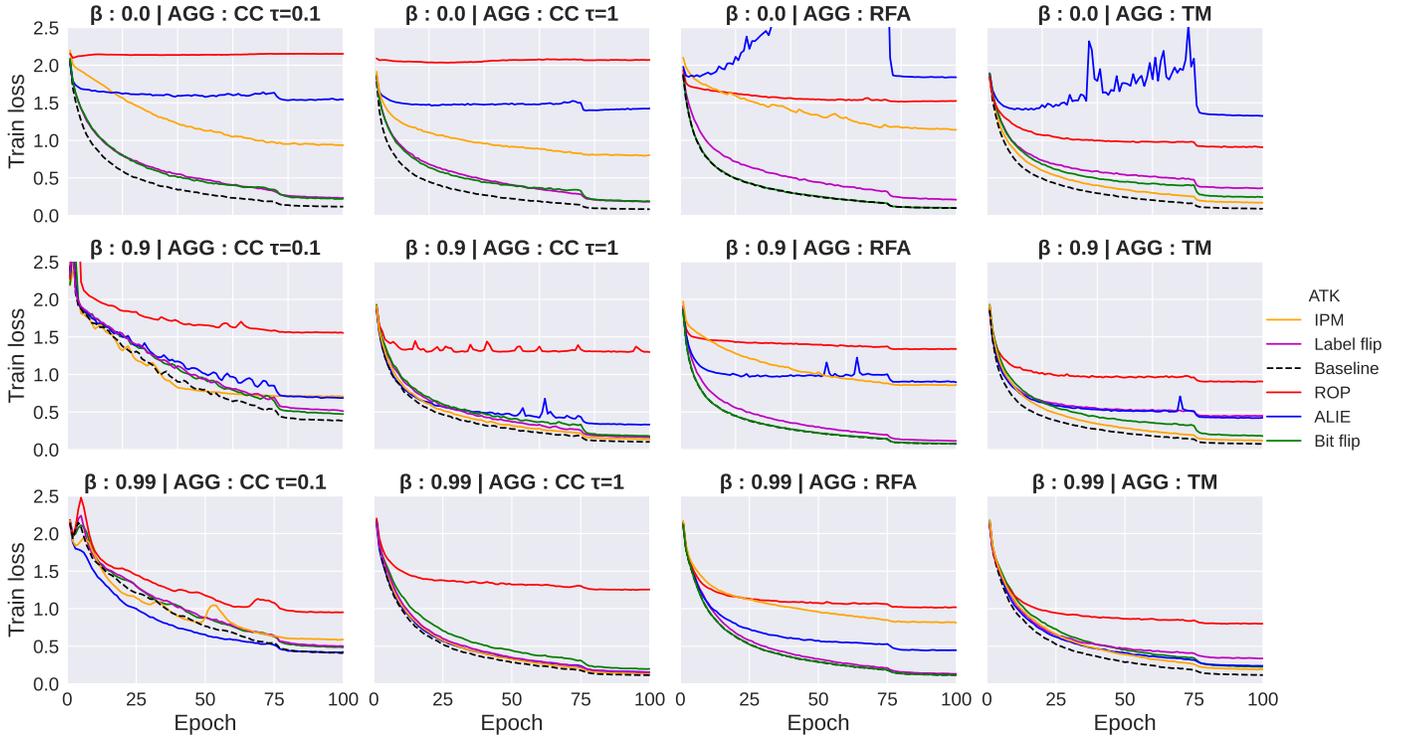}
    \caption{Cifar10 IID train loss on various aggregation methods.}
    \label{fig:cif10-iid-loss}
\end{figure*}

\begin{figure*}[h]
    \centering
    \includegraphics[width=1\textwidth]{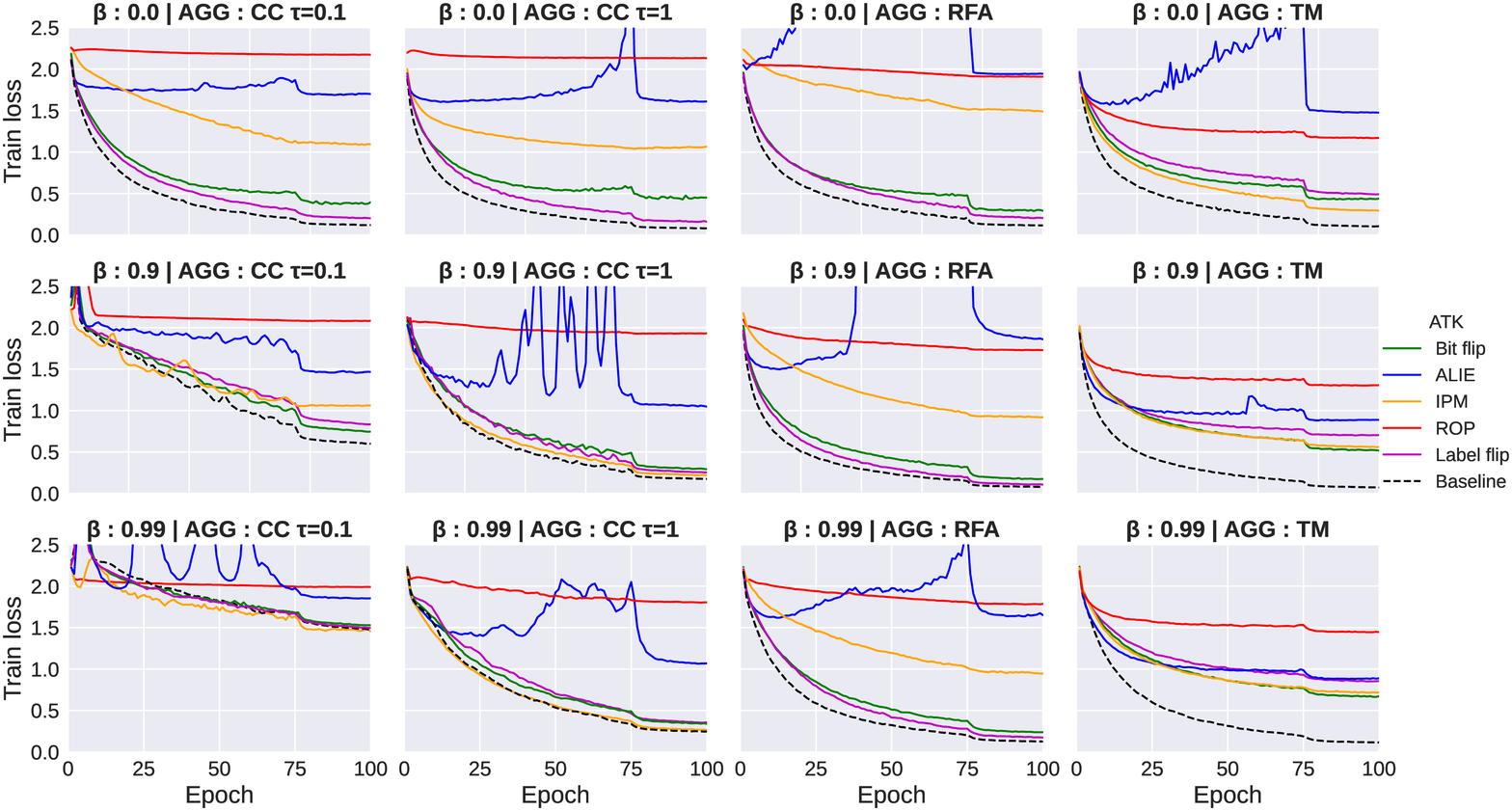}
    \caption{Cifar10 Non-IID Dir($\alpha=1$) train loss on various aggregation methods.}
    \label{fig:cif10-dir-loss}
\end{figure*}

This section shows that the proposed ROP attack prevents aggregators from reaching local minima by converging to saddle points instead. In Fig \ref{fig:cif10-iid-loss}, we show that on IID CIFAR-10, our ROP attack always converges to a saddle point for all aggregators and $\beta$ values. Unlike ROP, ALIE converges to saddle points at $\beta=0$ only, which can be explained by increased client variance when worker momentum is not employed. For the non-IID CIFAR10 image classification task, in Fig. \ref{fig:cif10-dir-loss}, we can further see the effects of the high variance for the ALIE attack. Although ALIE can increase its effectiveness when variance is high among the participating clients due to non-IID data distribution, it still has lower training loss on high $\beta$ values compared to the ROP thus converging to the local minima.

\subsection{Ablation study on ROP attack} \label{app:parameters}

In this section, we further illustrate the effects of the hyper-parameters of ROP, namely, $\lambda$ and $z$ parameters in Algorithm \ref{code:ROP}.

For the $\lambda$ hyper-parameter, we grid search the optimal value between [0, 0.5, 0.9, 1] on the IID CIFAR10 image classification problem. In Table \ref{tab:ablation}, we show that overall $\lambda = 0.9$ results in the strongest attack for multiple aggregators and $\beta$ values. We find that $\lambda=1$ is also quite effective to all aggregator types, meaning that Byzantine clients can generate strong attacks without being omniscient by only employing the broadcasted $\Tilde{\mathbf{m}}_{t-1}$ from the PS. 

Furthermore, we analyze the location of the attack and the angle of the attack with $\rho$ and $\pi$ hyper-parameters, respectively. In our extensive simulation results on Table \ref{tab:ablation}, we show that relocating the attack to $\Tilde{\mathbf{m}}_{t-1}$ has a greater effect on CC while targeting the $\Bar{\mathbf{m}}_{t}$ can significantly reduce the performance of the RFA. Regarding the angle of the perturbation, $\pi=90,120,135$ are equally capable of diminishing the test accuracy results.  By default, ROP employs $\rho=1$, $\lambda=0.9$ and $\pi=90$ 

\begin{figure*}[h]
    \centering
    \includegraphics[width=1\textwidth]{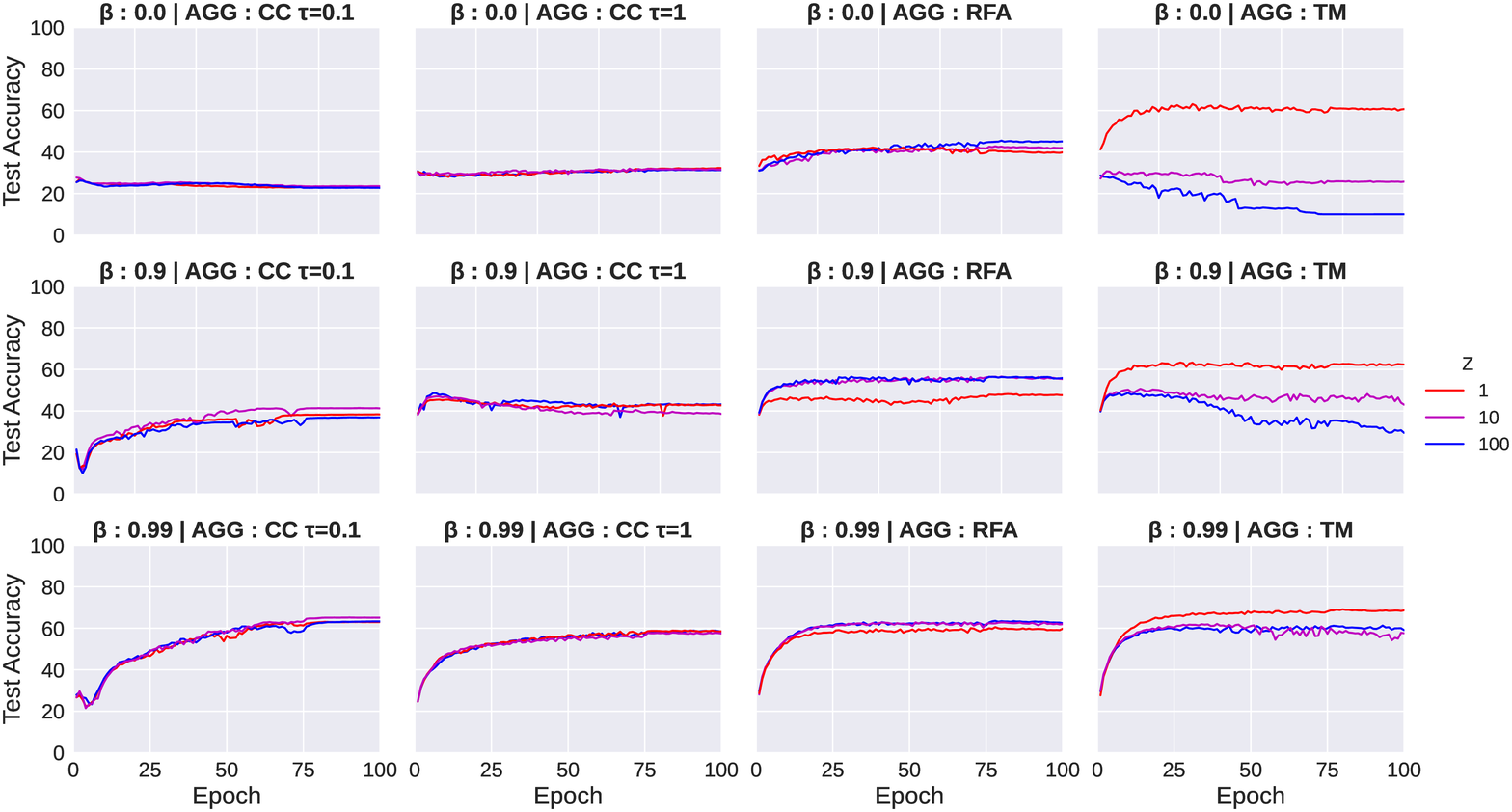}
    \caption{CIFAR10 test accuracy for ROP attack on various $z$ values for IID distribution.}
    \label{fig:cif10-z-iid}
\end{figure*}

\begin{figure*}[h]
    \centering
    \includegraphics[width=1\textwidth]{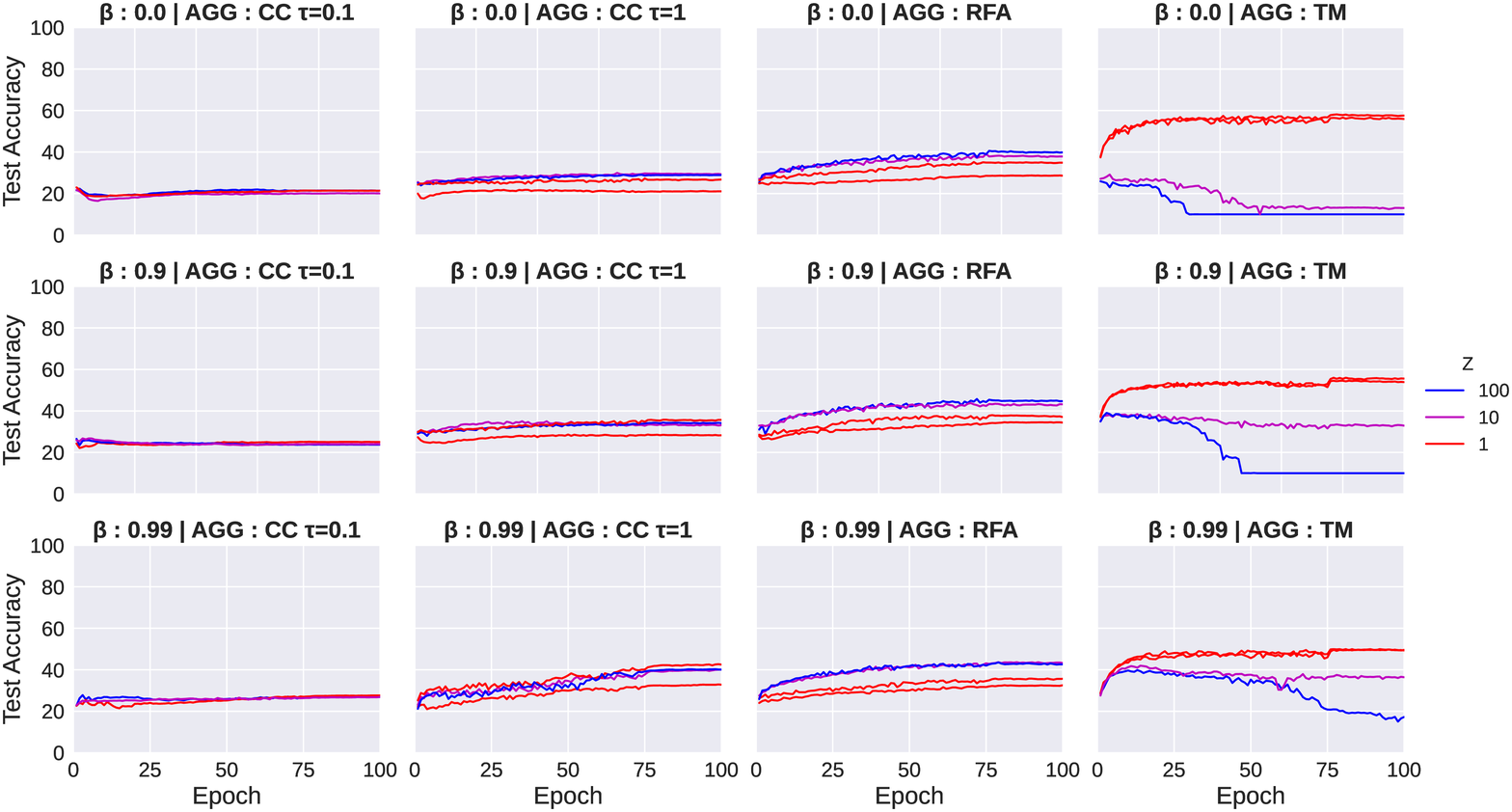}
    \caption{CIFAR10 test accuracy for ROP attack on various $z$ values for non-IID distribution.}
    \label{fig:cif10-z-dir}
\end{figure*}

For the $z$ hyper-parameter, we grid search the values [1, 10, 100] and find out that all $z$ values are equally robust to the CC aggregator due to the aforementioned relocation of the attack and angular invariance of the CC in Section \ref{sec:core}. On the TM aggregator, we find that an increased $z$ value also increases the robustness of the attack considerably compared to the other aggregators. We report our CIFAR-10 image classification task results for IID and non-IID data distributions in Fig. \ref{fig:cif10-z-iid} and Fig. \ref{fig:cif10-z-dir}, respectively. 

\begin{figure*}
    \centering
    \includegraphics[width=1\textwidth]{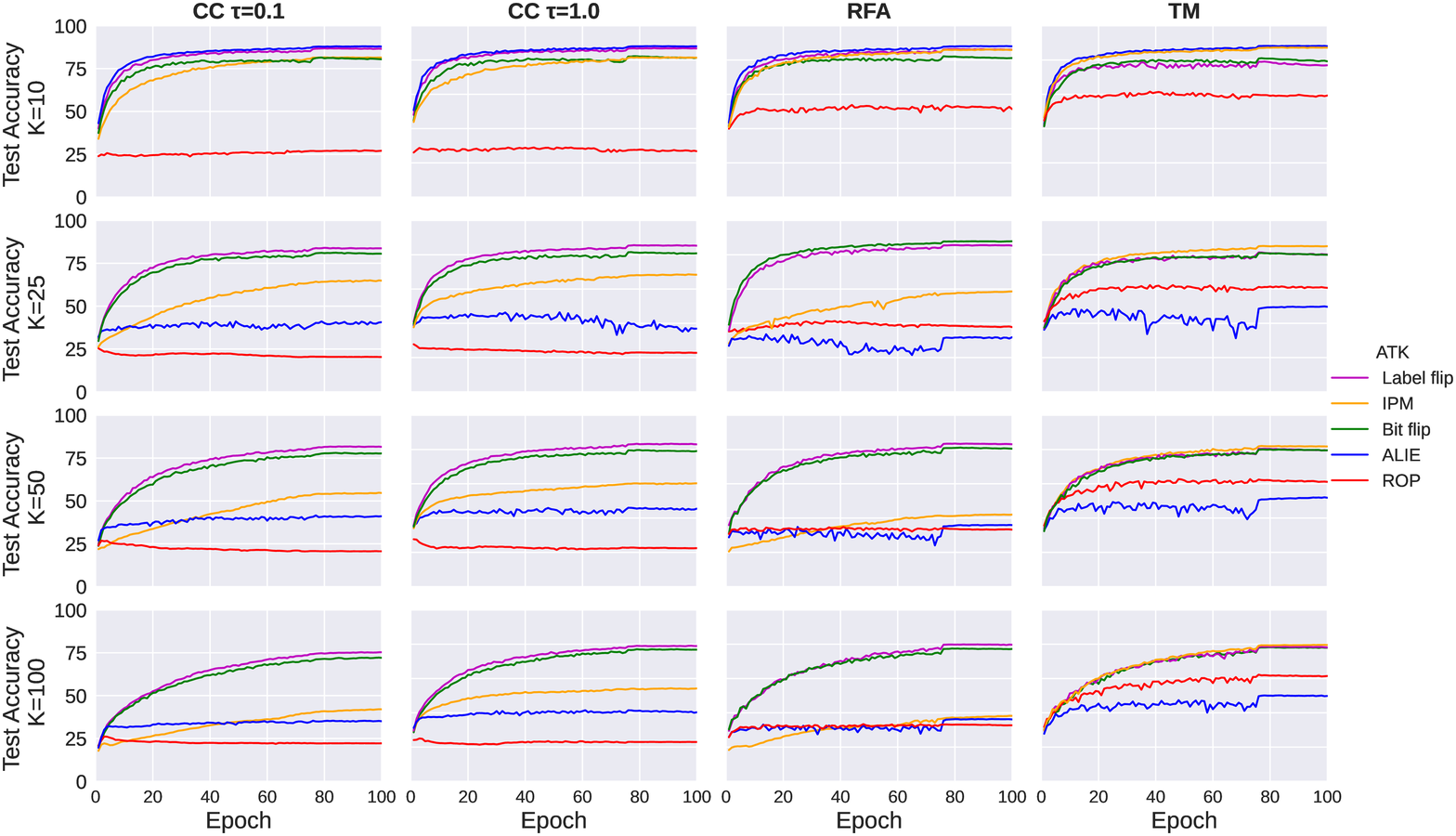}
    \caption{CIFAR-10 test accuracy results on IID data at $\beta$=0. Each row represents the total number of clients $k$ with \%20 Byzantines.}
    \label{fig:cli-iid0}
\end{figure*}

\begin{figure*}
    \centering
    \includegraphics[width=1\textwidth]{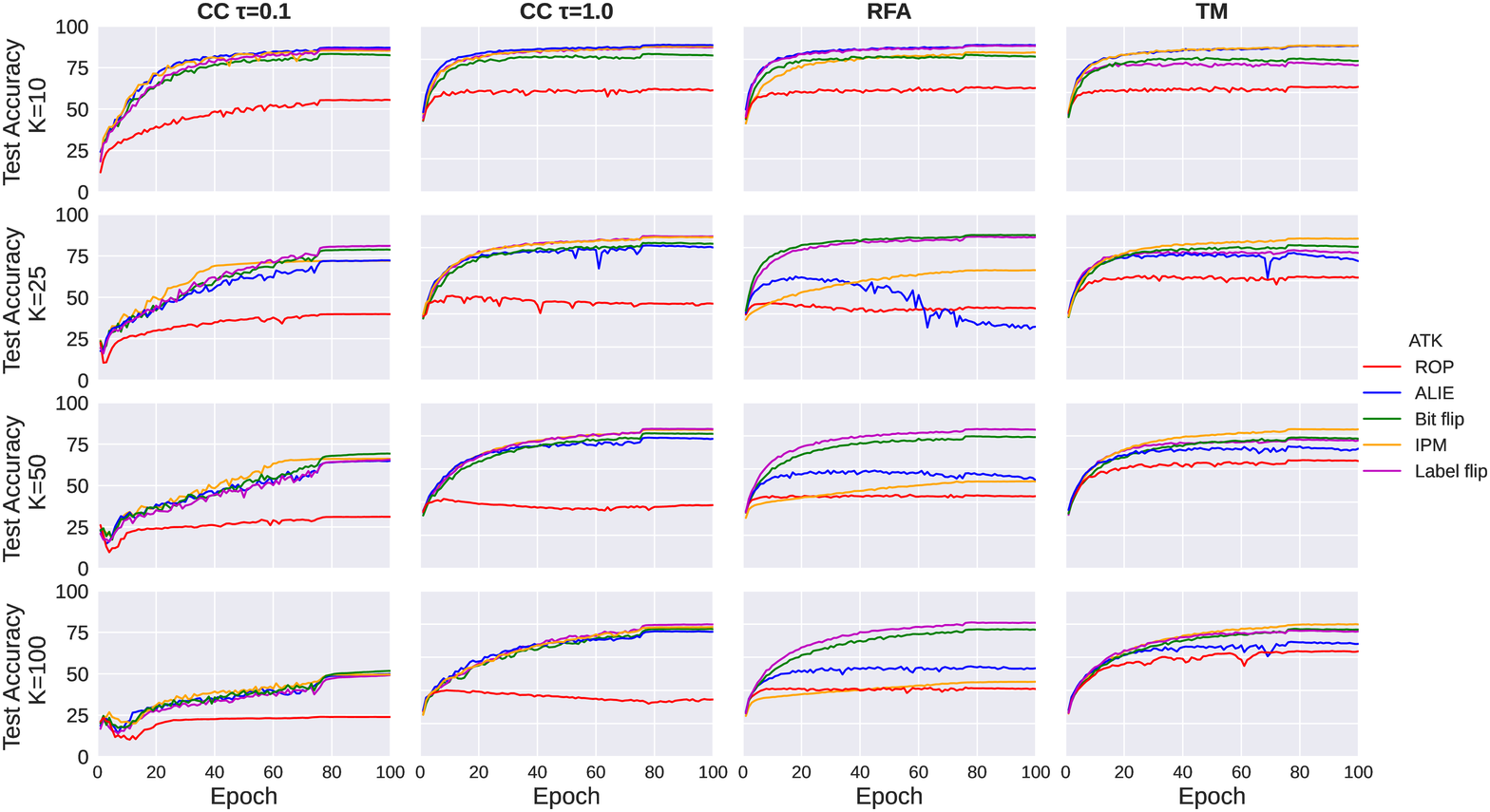}
    \caption{CIFAR-10 test accuracy results on IID data at $\beta$=0.9. Each row represents the total number of clients $k$ with \%20 Byzantines.}
    \label{fig:cli-iid9}
\end{figure*}

\begin{figure*}
    \centering
    \includegraphics[width=1\textwidth]{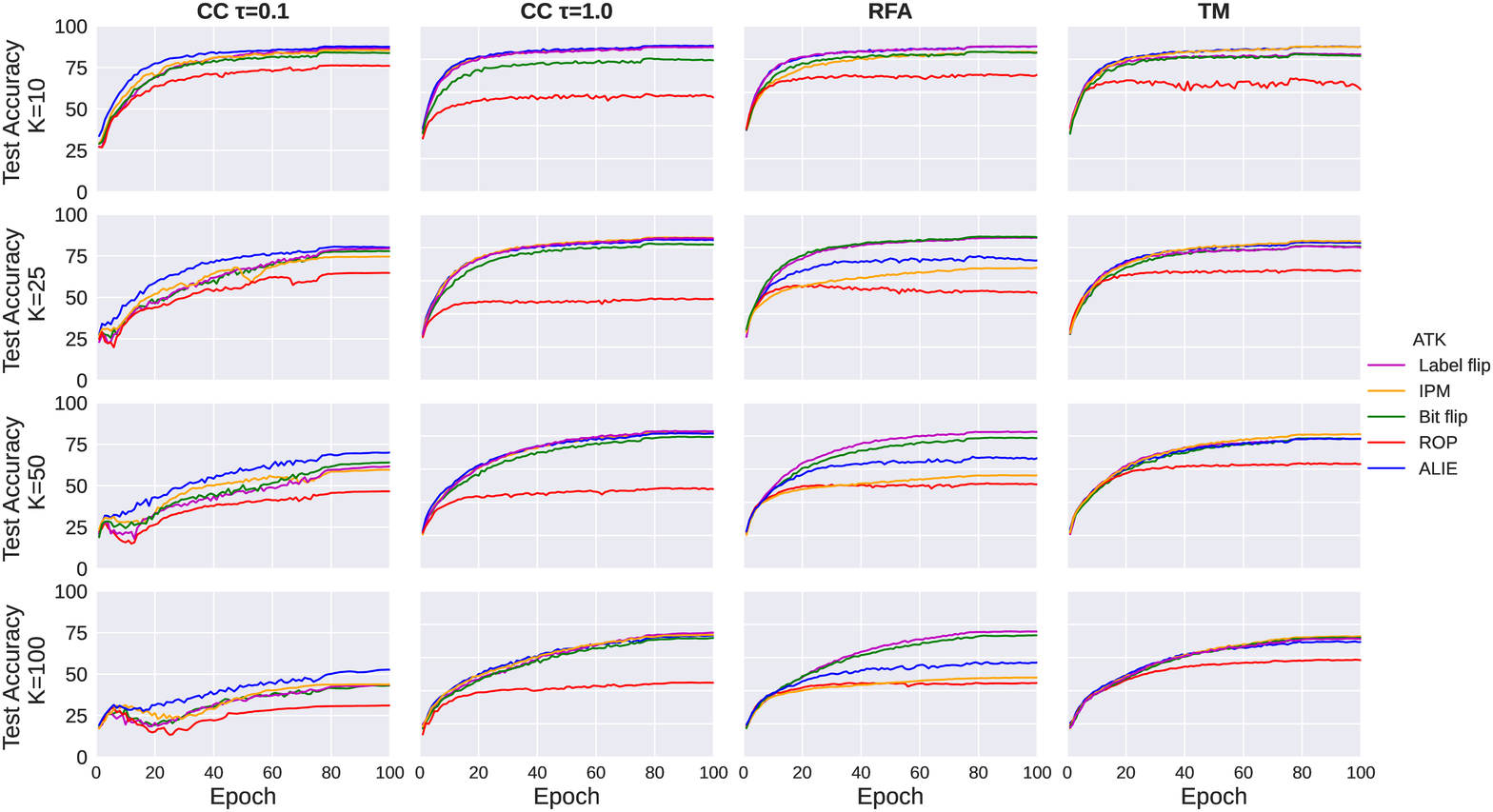}
    \caption{CIFAR-10 test accuracy results on IID data at $\beta$=0.99. Each row represents the total number of clients $k$ with \%20 Byzantines.}
    \label{fig:cli-iid99}
\end{figure*}

\begin{figure*}
    \centering
    \includegraphics[width=1\textwidth]{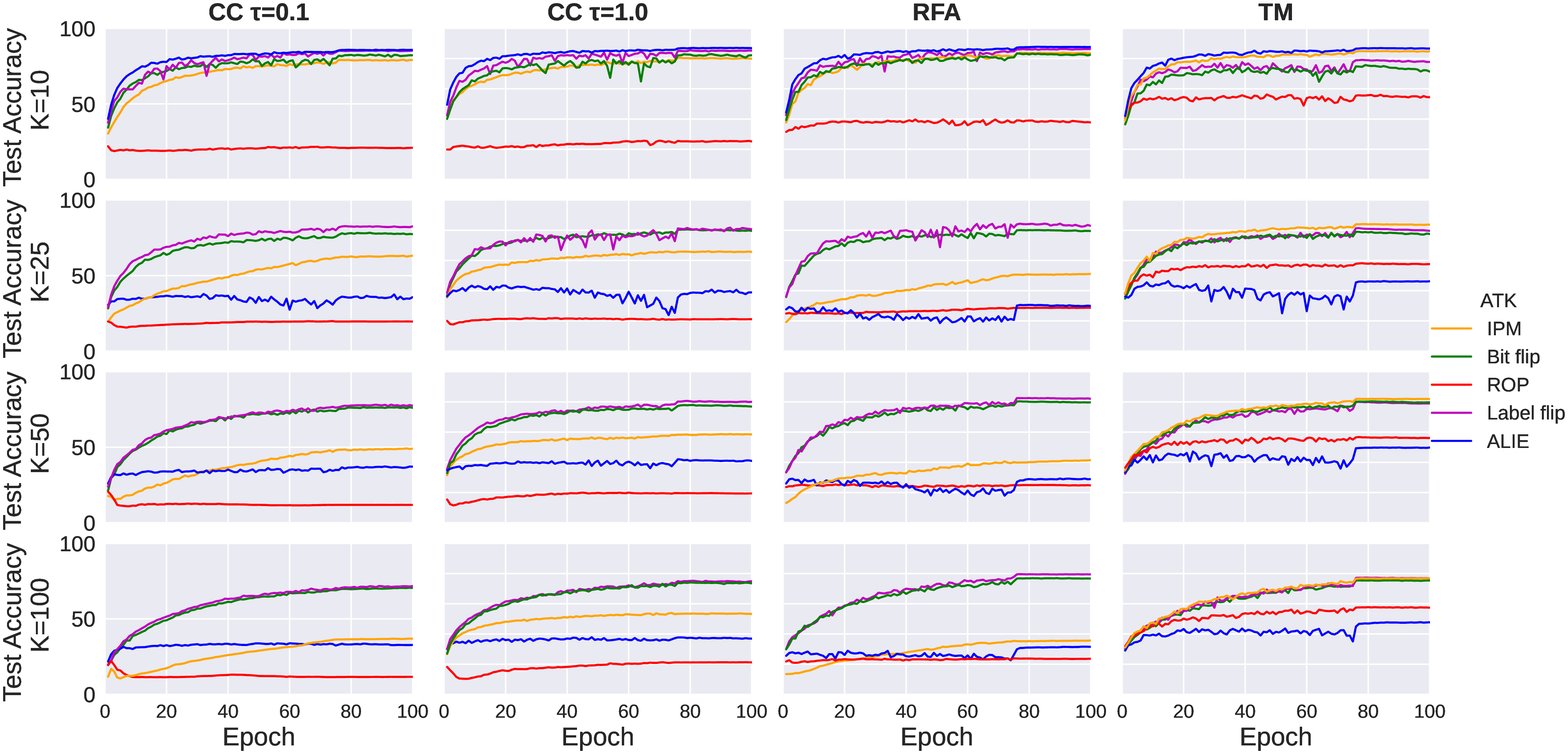}
    \caption{CIFAR-10 test accuracy results on non-IID data at $\beta$=0. Each row represents the total number of clients $k$ with \%20 Byzantines.}
    \label{fig:cli-dir0}
\end{figure*}

\begin{figure*}
    \centering
    \includegraphics[width=1\textwidth]{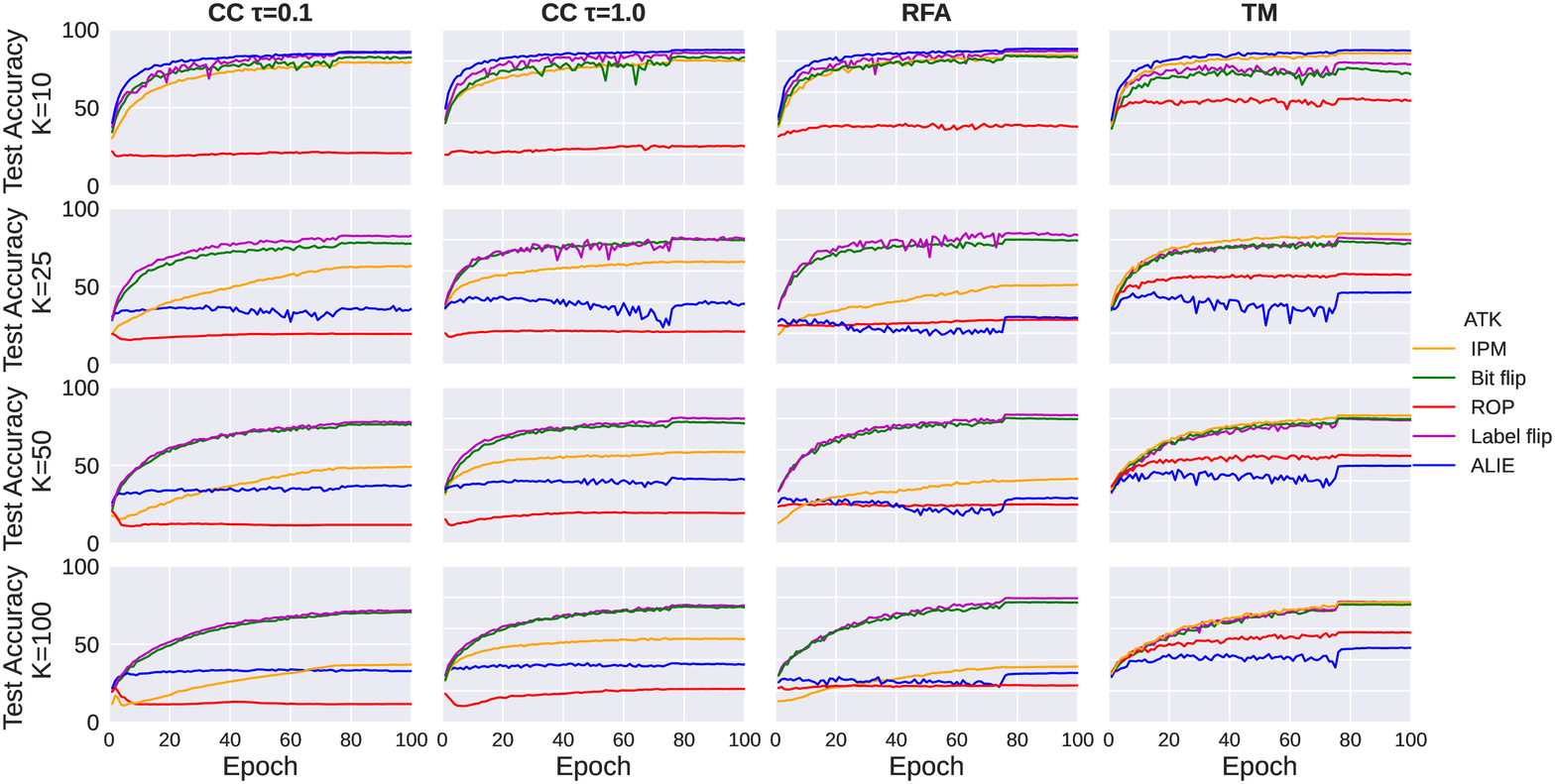}
    \caption{CIFAR-10 test accuracy results on non-IID data at $\beta$=0.9. Each row represents the total number of clients $k$ with \%20 Byzantines.}
    \label{fig:cli-dir9}
\end{figure*}

\begin{figure*}
    \centering
    \includegraphics[width=1\textwidth]{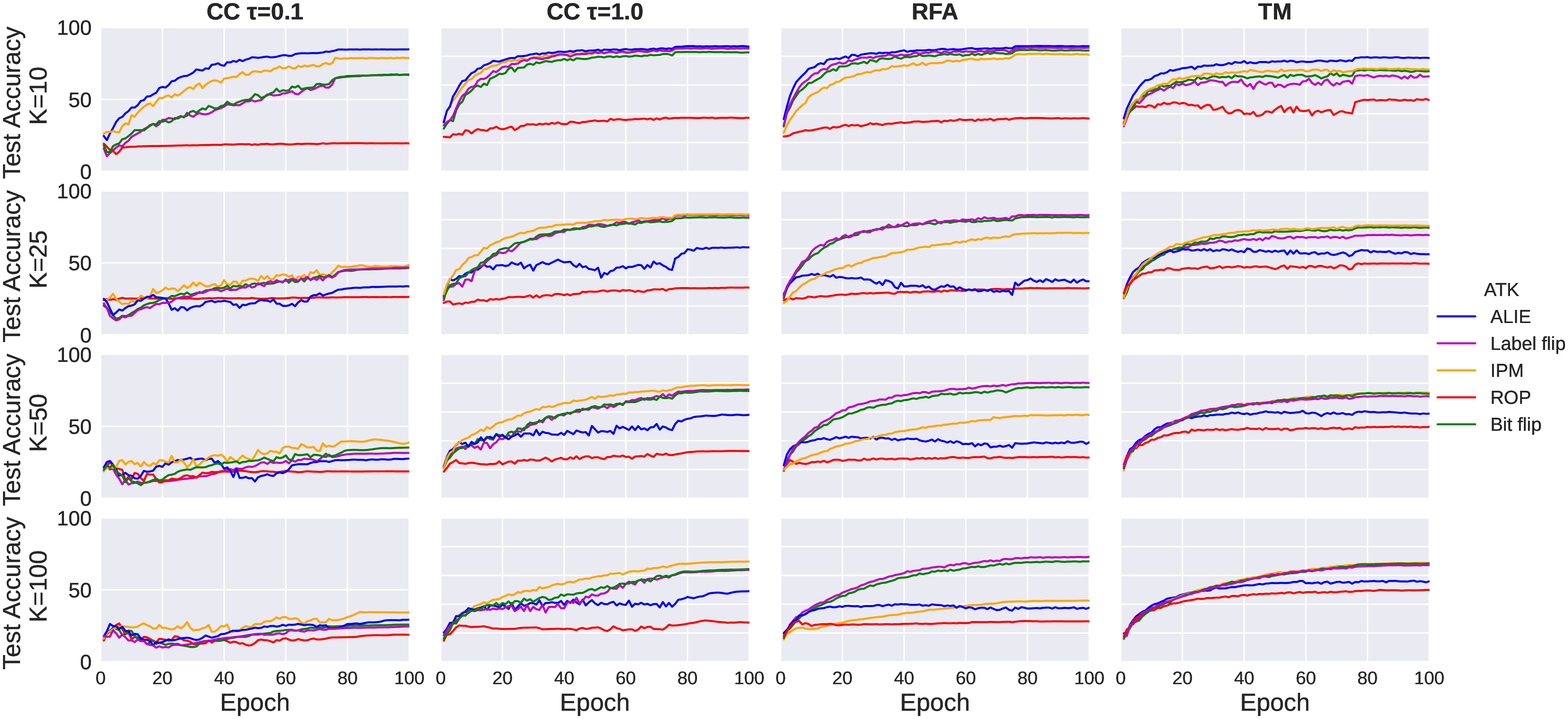}
    \caption{CIFAR-10 test accuracy results on non-IID data at $\beta$=0.99. Each row represents the total number of clients $k$ with \%20 Byzantines.}
    \label{fig:cli-dir99}
\end{figure*}

\revised{
\subsection{Study on total Byzantine and client numbers} \label{sec:cli-and-byz}
\subsubsection{Different Byzantine ratios}
For this study, we set the total number of clients $k$=25 with different numbers of Byzantines, specifically $k_{m}$=[1, 2, 3, 7, 8, 10, 12] where 12 is the upper bound for the maximum number of Byzantine clients that many aggregators can provide normal convergence.  We illustrated our results for IID data distribution in Table \ref{tab:Byzantines} and non-IID distribution in Table \ref{tab:Byzantines-dir}. We show that the proposed ROP attack can greatly reduce the test accuracy with only $k_{m}$=1 and $k_{m}$=2, while other attacks struggle to reduce the test accuracy. In Table \ref{tab:Byzantines}, with $\beta=0$, ROP is capable of reducing the test accuracy between 26-30 \%, while on Table \ref{tab:Byzantines-dir} at $\beta=0.9$, ROP can reduce the test accuracy between 5-36\%, clearly illustrating its effectiveness compared to other attacks. Furthermore, for $k_{m}$=[7, 8], ROP increased its effectiveness compared to other attacks. For $k_{m}$=[10, 12], we can finally see the effect of the IPM attack since it requires almost half of the clients as Byzantine to prevent convergence. However, IPM is still not as effective as ROP if local clients employ momentum, as seen in Table \ref{tab:Byzantines} with $\beta$=[0.9, 0.99].

\subsubsection{Effect of the total number of clients.}

In Figures \ref{fig:cli-iid0}, \ref{fig:cli-iid9}, \ref{fig:cli-iid99} \ref{fig:cli-dir0}, \ref{fig:cli-dir9}, \ref{fig:cli-dir99}, we illustrate the effectiveness of the proposed ROP and other attacks for different numbers of clients. For both IID and non-IID distributions, at $k$=10, ROP is by far the most effective attack that is capable of reducing the test accuracy up to 60\% for a CC aggregator while 25-30\% in RFA and TM in $\beta=$0, as seen in the Figs. \ref{fig:cli-iid0} and \ref{fig:cli-dir0}.  Even when the local momentum is employed, ROP is the only attack that has an impact on the test accuracy, as seen in Fig. \ref{fig:cli-dir9}. On high client numbers such as $k$=[50,100], Only ALIE surpasses the ROP on Figs \ref{fig:cli-iid0} and \ref{fig:cli-dir9} with a relatively small margin on TM aggregator. However, this is due to the ALIE assuming to know the standard deviation among the benign clients and thus capable of generating larger perturbation, meanwhile, ROP does not employ or need the standard deviation. To this end, ALIE can take advantage of large clients with no local momentum, as seen in Fig \ref{fig:cli-iid0} or large clients with very heterogeneous data in \ref{fig:cli-dir9} for the TM aggregator. However, on $\beta=$0.99 ROP is still is most successful attack as seen on Fig. \ref{fig:cli-iid99} and \ref{fig:cli-dir99} regardless of the data distribution. Overall out of 64 aggregator, client and local momentum combinations, in 56 scenarios, ROP is the most effective model poisoning attack.

}

\begin{table*}[]
\centering
\caption{ \revised{Cifar-10 IID test accuracy results on different numbers of Byzantines with a total of 25 clients.  Lower is better.}}
\label{tab:Byzantines}
\resizebox{\textwidth}{!}{%
\begin{tabular}{c|l|cccc|llll|cccc}
\multicolumn{1}{l|}{\multirow{2}{*}{\textbf{$k_{m}$}}} &  & \multicolumn{4}{c|}{\textbf{$\beta$=0}} & \multicolumn{4}{c|}{\textbf{$\beta$=0.9}} & \multicolumn{4}{c}{\textbf{$\beta$=0.99}} \\
\multicolumn{1}{l|}{} & \textbf{Attack / Aggr} & \multicolumn{1}{l}{\textbf{cc ($\tau$ =0.1)}} & \multicolumn{1}{l}{\textbf{cc}} & \multicolumn{1}{l}{\textbf{rfa}} & \multicolumn{1}{l|}{\textbf{tm}} & \textbf{cc ($\tau$ =0.1)} & \textbf{cc} & \textbf{rfa} & \textbf{tm} & \multicolumn{1}{l}{\textbf{cc ($\tau$ =0.1)}} & \multicolumn{1}{l}{\textbf{cc}} & \multicolumn{1}{l}{\textbf{rfa}} & \multicolumn{1}{l}{\textbf{tm}} \\ \hline
\multirow{5}{*}{1} & ROP & \textbf{56.6} & \textbf{60.4} & 81.2 & 83.2 & \textbf{79.98} & \textbf{85.40} & \textbf{83.27} & \textbf{81.98} & \textbf{80.95} & \textbf{85.10} & \textbf{85.65} & 85.86 \\
 & Alie & 86.9 & 87.0 & 87.4 & 87.3 & 82.88 & 87.73 & 87.49 & 87.74 & 82.20 & 86.74 & 86.26 & 86.07 \\
 & IPM & 84.92 & 85.82 & 87.00 & 86.26 & 82.15 & 87.79 & 87.20 & 87.01 & 81.54 & 86.39 & 86.06 & 85.37 \\
 & Bit-Flip & 86.26 & 86.82 & 87.09 & 86.39 & 80.93 & 87.30 & 86.44 & 86.97 & 82.03 & 85.85 & 86.04 & \textbf{85.16} \\
 & Label-flip & 86.30 & 87.00 & 86.87 & 85.60 & 82.20 & 87.77 & 87.10 & 86.58 & 82.02 & 86.63 & 86.48 & 85.77 \\ \hline
\multirow{5}{*}{2} & ROP & \textbf{40.77} & \textbf{45.05} & \textbf{72.71} & \textbf{71.05} & \textbf{71.29} & \textbf{79.96} & \textbf{79.66} & \textbf{63.93} & \textbf{79.63} & \textbf{79.55} & \textbf{81.86} & \textbf{82.39} \\
 & Alie & 80.73 & 79.74 & 67.32 & 74.98 & 82.94 & 87.47 & 84.71 & 85.84 & 82.29 & 86.75 & 85.71 & 85.70 \\
 & IPM & 82.26 & 84.21 & 85.85 & 85.29 & 82.28 & 86.67 & 85.65 & 86.09 & 80.72 & 86.76 & 84.02 & 84.61 \\
 & Bit-Flip & 84.94 & 85.48 & 85.99 & 85.12 & 82.31 & 86.57 & 86.36 & 85.59 & 80.44 & 85.20 & 85.46 & 84.25 \\
 & Label-flip & 86.09 & 86.76 & 86.81 & 84.94 & 82.66 & 86.99 & 86.77 & 85.19 & 81.04 & 86.15 & 86.37 & 85.10 \\ \hline
\multirow{5}{*}{3} & ROP & \textbf{31.91} & 34.44 & 60.84 & 60.64 & \textbf{62.67} & \textbf{71.57} & 66.49 & \textbf{65.76} & \textbf{75.63} & \textbf{65.88} & \textbf{70.91} & \textbf{76.13} \\
 & Alie & 34.16 & \textbf{29.42} & \textbf{50.73} & \textbf{53.83} & 80.94 & 85.81 & \textbf{60.04} & 73.38 & 83.30 & 85.86 & 84.09 & 83.65 \\
 & IPM & 78.39 & 81.42 & 83.86 & 84.41 & 80.94 & 86.46 & 82.39 & 85.32 & 79.18 & 85.28 & 80.22 & 83.59 \\
 & Bit-Flip & 84.63 & 84.55 & 84.63 & 84.34 & 80.32 & 85.21 & 85.08 & 84.79 & 79.56 & 83.64 & 84.01 & 83.69 \\
 & Label-flip & 84.73 & 86.52 & 86.32 & 83.42 & 82.17 & 86.65 & 86.94 & 83.40 & 80.86 & 86.40 & 86.05 & 84.47 \\ \hline
\multirow{5}{*}{7} & ROP & \textbf{10.52} & \textbf{22.61} & \textbf{24.74} & \textbf{22.09} & \textbf{28.36} & \textbf{33.95} & 37.60 & \textbf{38.51} & \textbf{44.47} & \textbf{45.82} & \textbf{46.59} & \textbf{49.23} \\
 & Alie & 33.21 & 32.52 & 25.99 & 22.16 & 59.97 & 54.80 & \textbf{33.50} & 45.07 & 71.38 & 79.97 & 56.53 & 52.93 \\
 & IPM & 45.28 & 58.24 & 36.35 & 50.87 & 61.79 & 84.73 & 54.58 & 68.70 & 60.85 & 84.91 & 62.01 & 70.71 \\
 & Bit-Flip & 74.04 & 73.17 & 74.75 & 75.07 & 76.13 & 77.40 & 75.60 & 76.02 & 74.94 & 78.15 & 78.14 & 77.97 \\
 & Label-flip & 82.92 & 83.33 & 83.55 & 77.09 & 79.12 & 86.01 & 85.03 & 74.37 & 78.00 & 84.87 & 84.97 & 78.62 \\ \hline
\multirow{5}{*}{8} & ROP & \textbf{10.62} & \textbf{19.83} & 22.69 & \textbf{11.37} & \textbf{23.39} & \textbf{33.35} & \textbf{35.43} & \textbf{33.86} & \textbf{32.62} & \textbf{40.28} & \textbf{43.92} & 52.83 \\
 & Alie & 29.19 & 31.39 & \textbf{21.09} & 18.36 & 51.77 & 27.37 & 48.70 & 39.70 & 63.08 & 75.23 & 56.09 & \textbf{48.79} \\
 & IPM & 22.23 & 34.87 & 28.59 & 31.75 & 55.61 & 83.71 & 50.33 & 62.68 & 64.21 & 84.05 & 59.81 & 66.92 \\
 & Bit-Flip & 69.47 & 69.71 & 70.52 & 69.65 & 73.82 & 74.13 & 71.81 & 72.19 & 74.62 & 74.53 & 76.12 & 75.38 \\
 & Label-flip & 80.85 & 81.78 & 81.87 & 72.74 & 78.65 & 85.29 & 84.39 & 71.64 & 76.32 & 84.99 & 84.46 & 76.51 \\ \hline
\multirow{5}{*}{10} & ROP & \textbf{10.66} & 19.53 & 13.99 & \textbf{12.39} & \textbf{19.88} & \textbf{33.52} & \textbf{32.74} & \textbf{21.14} & 37.25 & \textbf{32.05} & \textbf{40.00} & 40.83 \\
 & Alie & 19.98 & 19.25 & 16.41 & 14.57 & 37.50 & 38.10 & 33.98 & 26.29 & \textbf{34.80} & 56.67 & 45.88 & \textbf{39.11} \\
 & IPM & 14.56 & \textbf{12.98} & \textbf{10.27} & 12.40 & 21.95 & 81.67 & 41.62 & 48.86 & 59.04 & 83.09 & 54.02 & 57.30 \\
 & Bit-Flip & 56.77 & 62.73 & 59.86 & 51.43 & 63.02 & 62.79 & 63.42 & 58.90 & 65.01 & 67.12 & 65.93 & 66.23 \\
 & Label-flip & 75.23 & 74.92 & 76.56 & 72.48 & 75.39 & 82.47 & 82.06 & 67.90 & 73.52 & 83.79 & 84.29 & 72.50 \\ \hline
\multirow{5}{*}{12} & ROP & 10.00 & 10.01 & 10.01 & \textbf{10.63} & \textbf{10.36} & 31.40 & 27.13 & \textbf{17.96} & \textbf{22.51} & \textbf{25.04} & \textbf{33.84} & 34.32 \\
 & Alie & 16.92 & 16.79 & 13.87 & 16.49 & 28.64 & \textbf{24.78} & \textbf{24.00} & 20.80 & 31.17 & 40.00 & 36.64 & \textbf{28.92} \\
 & IPM & \textbf{7.68} & \textbf{8.84} & \textbf{10.00} & 10.97 & 10.80 & 75.21 & 24.38 & 31.94 & 35.79 & 82.38 & 34.10 & 47.08 \\
 & Bit-Flip & 33.91 & 33.90 & 36.71 & 36.71 & 33.02 & 38.27 & 40.62 & 38.94 & 46.24 & 48.24 & 47.20 & 49.53 \\
 & Label-flip & 53.85 & 52.07 & 51.04 & 59.58 & 60.89 & 57.37 & 58.96 & 61.49 & 62.67 & 69.97 & 69.77 & 62.23 \\ \cline{2-14} 
\end{tabular}%
}
\end{table*}

\begin{table*}[]
\centering
\caption{\revised{CIFAR-10 test accuracy results on different numbers of Byzantines with a total of 25 clients on the non-IID dataset.  Lower is better.}}
\label{tab:Byzantines-dir}
\resizebox{\textwidth}{!}{%
\begin{tabular}{c|l|llll|llll|llll}
\multicolumn{1}{l|}{\multirow{2}{*}{\textbf{$k_{m}$}}} &  & \multicolumn{4}{c|}{\textbf{$\beta$=0}} & \multicolumn{4}{c|}{\textbf{$\beta$=0.9}} & \multicolumn{4}{c}{\textbf{$\beta$=0.99}} \\
\multicolumn{1}{l|}{} & \textbf{Attack / Aggr} & \textbf{cc} & \textbf{cc} & \textbf{rfa} & \textbf{tm} & \textbf{cc} & \textbf{cc} & \textbf{rfa} & \textbf{tm} & \textbf{cc} & \textbf{cc} & \textbf{rfa} & \textbf{tm} \\ \hline
\multirow{5}{*}{1} & ROP & \textbf{55.60} & \textbf{64.59} & \textbf{44.61} & \textbf{57.53} & \textbf{66.06} & \textbf{81.50} & \textbf{77.56} & \textbf{72.30} & \textbf{47.42} & \textbf{75.24} & \textit{\textbf{68.88}} & \textbf{72.51} \\
 & Alie & 84.85 & 85.97 & 87.04 & 86.09 & 77.15 & 86.49 & 85.99 & 81.30 & 52.27 & 84.64 & 84.46 & 76.63 \\
 & IPM & 83.18 & 84.99 & 86.92 & 85.79 & 77.04 & 86.52 & 84.89 & 78.95 & 51.09 & 84.36 & 81.97 & 73.34 \\
 & Bit-Flip & 84.92 & 85.31 & 86.65 & 85.14 & 76.05 & 86.05 & 85.99 & 80.27 & 46.83 & 84.00 & 84.58 & 73.96 \\
 & Label-flip & 84.67 & 84.71 & 86.58 & 85.12 & 75.39 & 86.23 & 85.98 & 78.98 & 48.86 & 84.77 & 84.17 & \textbf{71.04} \\ \hline
\multirow{5}{*}{2} & ROP & \textbf{38.02} & \textbf{53.23} & \textbf{30.54} & \textbf{39.52} & \textbf{50.07} & \textbf{63.03} & \textbf{52.83} & \textbf{56.02} & \textbf{28.94} & \textbf{51.07} & \textbf{50.68} & \textbf{61.42} \\
 & Alie & 73.80 & 75.72 & 56.03 & 74.30 & 75.29 & 85.34 & 75.96 & 78.31 & 56.68 & 84.02 & 75.19 & 70.48 \\
 & IPM & 79.87 & 82.76 & 85.72 & 84.55 & 77.42 & 86.13 & 80.50 & 77.76 & 49.02 & 84.02 & 76.87 & 70.92 \\
 & Bit-Flip & 84.92 & 85.31 & 86.65 & 85.14 & 75.20 & 85.20 & 85.01 & 76.72 & 48.28 & 83.52 & 83.29 & 71.82 \\
 & Label-flip & 84.67 & 84.71 & 86.58 & 85.12 & 75.49 & 86.15 & 86.51 & 77.33 & 48.49 & 84.12 & 84.13 & 71.19 \\ \hline
\multirow{5}{*}{3} & ROP & \textbf{27.57} & 30.28 & \textbf{21.86} & \textbf{32.32} & \textbf{28.91} & \textbf{43.76} & 43.36 & 51.36 & \textbf{17.11} & \textbf{36.95} & 44.01 & \textbf{49.26} \\
 & Alie & 41.35 & \textbf{27.48} & 46.79 & 47.46 & 67.65 & 83.66 & \textbf{34.82} & \textbf{43.86} & 49.62 & 80.99 & \textbf{41.75} & 63.46 \\
 & IPM & 76.27 & 78.12 & 83.21 & 83.41 & 76.19 & 86.10 & 74.26 & 75.67 & 51.24 & 83.63 & 73.28 & 67.77 \\
 & Bit-Flip & 82.17 & 82.34 & 84.27 & 82.59 & 75.98 & 84.14 & 83.36 & 77.22 & 45.99 & 82.63 & 82.23 & 71.16 \\
 & Label-flip & 81.71 & 84.54 & 85.76 & 82.73 & 74.04 & 85.98 & 84.85 & 74.44 & 46.39 & 83.93 & 84.37 & 71.07 \\ \hline
\multirow{5}{*}{7} & ROP & \textbf{11.27} & \textbf{11.26} & \textbf{14.19} & \textbf{16.19} & \textbf{14.87} & \textbf{24.78} & 23.90 & \textbf{18.87} & \textbf{12.43} & \textbf{22.61} & \textbf{21.96} & 20.90 \\
 & Alie & 26.82 & 27.80 & 20.63 & 16.88 & 34.01 & 35.23 & \textbf{22.19} & 22.58 & 20.21 & 31.13 & 23.37 & \textbf{19.04} \\
 & IPM & 44.88 & 55.30 & 44.63 & 43.44 & 58.66 & 82.04 & 71.80 & 48.78 & 31.76 & 81.54 & 68.87 & 47.04 \\
 & Bit-Flip & 74.76 & 74.64 & 73.84 & 69.36 & 67.26 & 77.84 & 78.26 & 66.95 & 45.14 & 76.03 & 76.99 & 53.81 \\
 & Label-flip & 81.41 & 73.21 & 80.42 & 76.58 & 66.01 & 83.24 & 83.72 & 69.70 & 37.70 & 80.81 & 82.62 & 60.47 \\ \hline
\multirow{5}{*}{8} & ROP & \textbf{12.00} & \textbf{11.56} & 14.58 & \textbf{12.52} & \textbf{12.72} & \textbf{24.36} & \textbf{18.72} & \textbf{16.19} & \textbf{10.39} & \textbf{21.80} & \textbf{21.12} & 22.26 \\
 & Alie & 14.48 & 18.97 & \textbf{13.51} & 14.33 & 31.97 & 26.65 & 23.01 & 19.90 & 18.58 & 22.03 & 21.30 & \textbf{18.37} \\
 & IPM & 10.00 & 9.93 & 10.84 & 10.91 & 55.24 & 64.23 & 63.57 & 42.29 & 26.17 & 79.85 & 65.34 & 38.65 \\
 & Bit-Flip & 51.02 & 36.79 & 57.90 & 49.13 & 64.21 & 76.23 & 75.02 & 65.78 & 44.89 & 73.30 & 74.03 & 52.31 \\
 & Label-flip & 58.22 & 61.00 & 64.00 & 68.24 & 69.46 & 83.01 & 82.37 & 64.23 & 36.91 & 80.41 & 81.86 & 60.77 \\ \hline
\multirow{5}{*}{10} & ROP & \textbf{10.00} & \textbf{10.00} & \textbf{10.10} & \textbf{9.94} & \textbf{11.13} & \textbf{17.19} & \textbf{10.00} & \textbf{13.23} & \textbf{10.79} & 18.38 & \textbf{11.04} & 14.79 \\
 & Alie & 15.52 & 16.20 & 14.35 & 16.82 & 18.66 & 18.01 & 14.41 & 16.81 & 19.33 & \textbf{14.27} & 20.76 & \textbf{12.81} \\
 & IPM & \textbf{10.00} & \textbf{10.00} & \textbf{10.00} & 10.92 & 31.18 & 36.00 & 52.30 & 26.45 & 17.48 & 74.74 & 66.48 & 25.79 \\
 & Bit-Flip & 12.80 & 20.42 & 10.00 & 14.47 & 39.45 & 60.63 & 54.30 & 44.80 & 32.99 & 60.90 & 52.81 & 36.19 \\
 & Label-flip & 29.11 & 51.51 & 44.65 & 47.81 & 56.39 & 75.56 & 79.15 & 60.50 & 37.51 & 69.52 & 80.12 & 54.04 \\ \hline
\multirow{5}{*}{12} & ROP & 10.00 & 10.01 & 10.01 & \textbf{10.63} & 10.00 & 15.53 & \textbf{10.00} & \textbf{10.01} & \textbf{10.00} & 19.40 & \textbf{10.00} & \textbf{10.80} \\
 & Alie & 16.92 & 16.79 & 13.87 & 16.49 & 17.10 & \textbf{13.78} & 15.47 & 15.55 & 14.30 & 14.49 & 17.91 & 17.93 \\
 & IPM & \textbf{7.68} & \textbf{8.84} & \textbf{10.00} & 10.97 & \textbf{9.90} & 12.64 & \textbf{10.00} & 15.76 & 10.73 & \textbf{10.11} & 10.34 & 14.37 \\
 & Bit-Flip & 33.91 & 33.90 & 36.71 & 36.71 & 11.42 & 10.98 & 12.48 & 20.29 & 17.16 & 20.66 & 10.54 & 16.84 \\
 & Label-flip & 53.85 & 52.07 & 51.04 & 59.58 & 44.44 & 61.10 & 44.07 & 43.17 & 26.41 & 41.58 & 62.52 & 38.05 \\ \hline
\end{tabular}%
}
\end{table*}

\begin{table*}[h]
    \caption{Test accuracy comparisons on all datasets. Lower is better. $\textbf{*}$ denotes non-IID distribution.}
    \label{tab:All_results}
    \resizebox{\textwidth}{!}{%
    \begin{tabular}{l|l|lll|lll|lll|lll|}
\hline
\multicolumn{1}{c|}{\multirow{2}{*}{Dataset}} & \multicolumn{1}{c|}{\multirow{2}{*}{Attack}} &  & \multicolumn{1}{c}{CC $\tau=$ 0.1} & \textbf{} & \textbf{} & \multicolumn{1}{c}{CC $\tau=$ 1} & \textbf{} & \textbf{} & \multicolumn{1}{c}{RFA} & \textbf{} & \textbf{} & \multicolumn{1}{c}{TM} & \textbf{} \\
\multicolumn{1}{c|}{} & \multicolumn{1}{c|}{} & \textbf{$\beta$=0} & $\textbf{$\beta$=0.9}$ & \textbf{$\beta$=0.99} & \textbf{$\beta$=0} & \textbf{$\beta$=0.9} & \textbf{$\beta$=0.99} & \textbf{$\beta$=0} & \textbf{$\beta$=0.9} & \textbf{$\beta$=0.99} & \textbf{$\beta$=0} & \textbf{$\beta$=0.9} & \textbf{$\beta$=0.99} \\ \hline
 & ROP & 68.52 $\pm$ 0.73 & \textbf{85.02 $\pm$ 1.33} & \textbf{85.17 $\pm$ 1.33} & \textbf{70.63 $\pm$ 0.82} & \textbf{79.31 $\pm$ 7.76} & \textbf{83.8 $\pm$ 1.71} & 82.85 $\pm$ 0.45 & \textbf{83.45 $\pm$ 0.26} & \textbf{86.81 $\pm$ 0.27} & $84.43 \pm2.46$ & \textbf{86.2 $\pm$ 0.16} & \textbf{87.24 $\pm$ 0.16} \\
 & ALIE & \textbf{60.14 $\pm$ 17.0} & 89.42 $\pm$ 0.22 & 89.16 $\pm$ 0.52 & $84.8 \pm0.85$ & $90.55 \pm0.18$ & $90.71 \pm0.2$ & \textbf{10.0 $\pm$ 0.0} & $87.43 \pm1.19$ & $90.1 \pm0.2$ & \textbf{10.0 $\pm$ 0.0} & $90.08 \pm0.21$ & $90.72 \pm0.15$ \\
\textbf{FMNIST} & IPM & $84.28 \pm0.14$ & $88.27 \pm0.37$ & 88.31 $\pm$ 0.83 & $88.7 \pm0.17$ & $90.54 \pm0.22$ & $90.67 \pm0.15$ & $80.86 \pm1.11$ & $86.64 \pm0.14$ & $88.82 \pm0.18$ & $88.94 \pm0.3$ & $90.28 \pm0.14$ & $90.13 \pm0.19$ \\
 & Label flip & $90.82 \pm0.16$ & $89.83 \pm0.33$ & $89.46 \pm 0.29$ & $90.43 \pm0.16$ & $90.87 \pm0.22$ & $90.76 \pm0.22$ & $90.33 \pm0.34$ & $90.9 \pm0.14$ & $90.84 \pm0.19$ & $60.91 \pm25.56$ & $82.59 \pm0.56$ & $86.73 \pm0.93$ \\
 & Bit flip & $88.79 \pm0.14$ & $88.88 \pm0.19$ & $88.34 \pm 0.53$ & $89.11 \pm0.09$ & $89.19 \pm0.16$ & $88.5 \pm0.18$ & $88.97 \pm0.27$ & $89.28 \pm0.22$ & $89.56 \pm0.14$ & $88.82 \pm0.13$ & $89.3 \pm0.08$ & $89.22 \pm0.16$ \\ \hline
 & ROP & $\textbf{20.33 $\pm$ 1.73}$ & $\textbf{39.82 $\pm$ 3}$ & $\textbf{64.75 $\pm$ 0.18}$ & \textbf{22.79 $\pm$ 1.12} & \textbf{46.15 $\pm$ 1.92} & \textbf{48.91 $\pm$ 0.13} & $37.8 \pm 0.52$ & $43.25 \pm 2.23$ & \textbf{52.6 $\pm$ 1.8} & $60.7 \pm 0.79$ & \textbf{61.93 $\pm$ 0.75} & \textbf{65.9 $\pm$ 0.5} \\
 & ALIE & $40.68 \pm2.27$ & $72.24 \pm0.15$ & $80.1 \pm0.91$ & $36.89 \pm9.38$ & $80.09 \pm1.38$ & $84.59 \pm0.19$ & \textbf{31.9 $\pm$ 1.62} & \textbf{32.26 $\pm$ 16.08} & $72.26 \pm1.94$ & \textbf{49.64 $\pm$ 1.05} & $71.83 \pm2.6$ & $82.9 \pm0.4$ \\
\textbf{CIFAR10} & IPM & $64.9 \pm1.08$ & $72.02 \pm1.17$ & $74.62 \pm1.16$ & $68.33 \pm1.3$ & $86.28 \pm0.35$ & $85.81 \pm0.17$ & $58.58 \pm1.05$ & $66.36 \pm0.67$ & $67.85 \pm0.62$ & $85.0 \pm0.26$ & $85.42 \pm0.72$ & $83.88 \pm0.37$ \\
 & Label flip & $83.74 \pm0.1$ & $81.04 \pm0.28$ & $79.54 \pm0.28$ & $85.29 \pm0.61$ & $86.79 \pm0.19$ & $85.45 \pm0.28$ & $85.32 \pm0.32$ & $86.28 \pm0.08$ & $85.9 \pm0.38$ & $80.2 \pm0.04$ & $76.79 \pm0.24$ & $80.03 \pm1.28$ \\
 & Bit flip & $80.62 \pm0.38$ & $78.72 \pm0.4$ & $77.87 \pm0.08$ & $80.84 \pm0.24$ & $82.32 \pm1.03$ & $81.88 \pm0.08$ & $87.85 \pm0.3$ & $87.5 \pm0.65$ & $86.31 \pm0.29$ & $79.95 \pm0.16$ & $80.53 \pm0.42$ & $80.69 \pm0.45$ \\ \hline
 & ROP & \textbf{1.04 $\pm$0.07} & \textbf{23.24 $\pm$0.4} & \textbf{33.36 $\pm$ 0.6} & \textbf{6.43 $\pm$ 0.32} & \textbf{10.67 $\pm$ 0.46} & \textbf{19.22 $\pm$ 0.3} & 9 $\pm$ 0.12 & \textbf{14.56 $\pm$ 0.12} & \textbf{21.18 $\pm$ 0.2} & \textbf{18.1 $\pm$ 0.27} & $32.5 \pm1$ & \textbf{35.22 $\pm$ 0.76} \\
 & ALIE & $11.82 \pm0.04$ & $47.53 \pm0.38$ & $49.4 \pm0.86$ & $16.26 \pm0.1$ & $45.88 \pm3.68$ & $59.04 \pm0.48$ & 9.1 $\pm$ 0.35 & $17.75 \pm7.07$ & $23.11 \pm1.41$ & $26.83 \pm0.32$ & \textbf{26.48 $\pm$ 9.63} & $54.77 \pm0.4$ \\
\textbf{CIFAR100} & IPM & $10.06 \pm2.22$ & $46.56 \pm2.63$ & $44.78 \pm0.02$ & $32.0 \pm0.4$ & $60.22 \pm0.04$ & $60.95 \pm0.14$ & \textbf{6.1 $\pm$ 3.38} & $18.76 \pm0.39$ & $29.2 \pm0.21$ & $53.97 \pm0.17$ & $60.6 \pm0.16$ & $58.88 \pm0.11$ \\
 & Label flip & $57.52 \pm0.36$ & $53.58 \pm0.55$ & $50.72 \pm0.09$ & $58.16 \pm0.42$ & $61.95 \pm0.23$ & $61.08 \pm0.32$ & $60.12 \pm0.03$ & $59.55 \pm0.31$ & $60.2 \pm0.7$ & $56.48 \pm0.79$ & $57.4 \pm0.67$ & $56.38 \pm0.92$ \\
 & Bit flip & $50.04 \pm0.44$ & $49.62 \pm0.22$ & $47.26 \pm0.4$ & $49.52 \pm0.91$ & $53.94 \pm0.11$ & $53.37 \pm0.45$ & $51.02 \pm0.55$ & $50.99 \pm0.02$ & $53.26 \pm0.24$ & $50.5 \pm0.68$ & $49.45 \pm0.73$ & $51.07 \pm0.02$ \\ \hline
 & ROP & \textbf{82.16 $\pm$ 1.43} & \textbf{97.47 $\pm$ 0.13} & \textbf{9.88 $\pm$ 0.11} & \textbf{78.8 $\pm$ 5.16} & \textbf{95.4 $\pm$ 0.7} & \textbf{97.73 $\pm$ 0.3} & $91.79 \pm 8.34$ & $96.59 \pm0.16$ & \textbf{97.83 $\pm$ 0.29} & $95.42 \pm0.5$ & $97.36 \pm0.4$ & $97.7 \pm 2.6$ \\
 & ALIE & $97.5 \pm0.04$ & $97.61 \pm0.49$ & $9.96 \pm0.14$ & $96.19 \pm0.82$ & $98.4 \pm0.12$ & $98.74 \pm0.05$ & \textbf{9.82 $\pm$ 0.0} & \textbf{51.47 $\pm$ 41.37} & $98.5 \pm0.4$ & \textbf{9.89 $\pm$ 0.12} & $98.91 \pm0.08$ & $99.0 \pm0.07$ \\
\textbf{MNIST *} & IPM & $93.07 \pm2.78$ & $98.83 \pm0.04$ & $96.55 \pm0.59$ & $98.84 \pm0.09$ & $99.01 \pm0.03$ & $98.98 \pm0.01$ & $93.34 \pm2.74$ & $98.44 \pm0.11$ & $98.37 \pm0.28$ & $95.61 \pm0.9$ & $98.03 \pm1.23$ & $98.82 \pm0.07$ \\
 & Label flip & $98.8 \pm0.0$ & $98.78 \pm0.02$ & $52.74 \pm43.0$ & $98.98 \pm0.0$ & $99.02 \pm0.03$ & $98.95 \pm0.02$ & $93.34 \pm2.74$ & $99.09 \pm0.15$ & $99.01 \pm0.08$ & $94.54 \pm0.57$ & \textbf{94.85 $\pm$ 1.14} & \textbf{96.78 $\pm$ 0.46} \\
 & Bit flip & $98.36 \pm0.06$ & $98.38 \pm0.08$ & $68.58 \pm40.47$ & $98.29 \pm0.1$ & $97.88 \pm0.18$ & $97.78 \pm0.13$ & $98.37 \pm0.31$ & $98.28 \pm0.54$ & $98.74 \pm0.05$ & $74.68 \pm37.47$ & $97.86 \pm0.61$ & $98.27 \pm0.25$ \\ \hline
 & ROP & $62.7 \pm0.9$ & \textbf{75.35 $\pm$ 0.88} & $42 \pm 32$ & $68 \pm1.35$ & $\textbf{75.5 $\pm$ 2.61}$ & $\textbf{75.16 $\pm$ 0.4}$ & $80.74 \pm 1.6$ & $76.53 \pm 1.35$ & $73.39 \pm 1.8$ & $85.29 \pm0.37$ & $84.19 \pm 0.37$ & $84.52 \pm0.34$ \\
 & ALIE & \textbf{10.0 $\pm$ 0.0} & $86.13 \pm0.41$ & \textbf{10.0 $\pm$ 0.0} & \textbf{46.38 $\pm$ 36.38} & $86.98 \pm0.34$ & $87.33 \pm0.96$ & \textbf{10.0 $\pm$ 0.0} & \textbf{24.18 $\pm$ 28.35} & \textbf{19.24 $\pm$ 18.47} & \textbf{21.34 $\pm$ 22.69} & $83.85 \pm3.57$ & 84.13 $\pm$ 1.52 \\
\textbf{FMNIST *} & IPM & $79.34 \pm0.62$ & $83.43 \pm1.1$ & $84.18 \pm0.75$ & $88.44 \pm0.44$ & $90.17 \pm0.12$ & $89.97 \pm0.3$ & $75.28 \pm2.13$ & $81.23 \pm1.32$ & $82.46 \pm1.25$ & $88.33 \pm0.18$ & $87.61 \pm1.08$ & $86.55 \pm1.17$ \\
 & Label flip & $90.77 \pm0.06$ & $89.7 \pm0.14$ & $53.38 \pm35.44$ & $90.55 \pm0.14$ & $90.23 \pm0.26$ & $90.13 \pm0.19$ & $90.27 \pm0.27$ & $90.61 \pm0.2$ & $90.59 \pm0.21$ & $64.97 \pm28.01$ & \textbf{$56.9 \pm23.76$} & \textbf{59.43 $\pm$ 24.94} \\
 & Bit flip & $88.46 \pm0.27$ & $88.29 \pm0.48$ & $84.22 \pm1.33$ & $88.91 \pm0.28$ & $88.68 \pm0.21$ & $89.03 \pm0.41$ & $88.58 \pm0.21$ & $88.94 \pm0.16$ & $88.87 \pm0.2$ & $88.08 \pm0.49$ & $87.42 \pm0.26$ & $86.06 \pm0.87$ \\ \hline
 & ROP & $\textbf{21.8 $\pm$ 0.86}$ & $\textbf{23.18 $\pm$ 1.35}$ & $\textbf{18.18 $\pm$ 0.28}$ & $\textbf{19.6 $\pm$ 1}$ & $\textbf{23.64 $\pm$ 1.9}$ & $\textbf{26.36 $\pm$ 1.72}$ & $\textbf{28.64 $\pm$ 1}$ & $34.4 \pm 2$ & $\textbf{32.5 $\pm$ 1.4}$ & $57.6 \pm 0.6$ & $53.9 \pm 1.88$ & $\textbf{49.32 $\pm$ 1.4}$ \\
 & ALIE & $35.71 \pm3.89$ & $47.54 \pm3.36$ & $33.69 \pm0.29$ & $38.87 \pm2.96$ & $60.54 \pm3.59$ & $60.78 \pm1.42$ & $29.91 \pm1.7$ & $\textbf{32.85 $\pm$ 3.35}$ & $37.17 \pm5.38$ & $\textbf{46.22 $\pm$ 1.56}$ & \textbf{43.1 $\pm$ 12.38} & $56.02 \pm4.89$ \\
\textbf{CIFAR10 *} & IPM & $63.11 \pm1.1$ & $65.98 \pm3.11$ & $48.63 \pm2.79$ & $65.66 \pm1.26$ & $84.48 \pm0.73$ & $83.76 \pm0.44$ & $51.12 \pm1.72$ & $67.88 \pm1.91$ & $70.67 \pm1.23$ & $83.7 \pm0.77$ & $76.54 \pm1.7$ & $75.7 \pm0.37$ \\
 & Label flip & $82.58 \pm0.58$ & $72.28 \pm1.52$ & $46.84 \pm1.16$ & $80.63 \pm0.02$ & $84.99 \pm0.45$ & $82.96 \pm0.81$ & $82.91 \pm2.86$ & $84.63 \pm0.03$ & $83.2 \pm0.57$ & $79.6 \pm0.73$ & $74.31 \pm1.12$ & $69.19 \pm3.06$ \\
 & Bit flip & $77.42 \pm1.41$ & $73.84 \pm1.19$ & $46.56 \pm2.66$ & $79.6 \pm0.5$ & $80.49 \pm0.51$ & $81.33 \pm0.33$ & $79.48 \pm1.12$ & $80.94 \pm0.64$ & $81.71 \pm0.43$ & $77.52 \pm3.27$ & $75.39 \pm0.64$ & $74.31 \pm1.28$ \\ \hline
\end{tabular}%
}
\end{table*}

\begin{table*}[h]
\centering
\caption{Hyper-parameter search on attack location $\rho$, reference point $\lambda$, and angle of the perturbation $\pi$ w.r.t reference for the CIFAR-10 image classification task. The lowest score for each aggregator for the respective simulation setup is denoted in $\textbf{bold}$}
\label{tab:ablation}
\resizebox{\textwidth}{!}{%
\begin{tabular}{l|lll|llllll}
\hline
 & \multicolumn{3}{c|}{IID $\beta$ = 0.9} & \multicolumn{3}{c|}{IID $\beta$ = 0.99} & \multicolumn{3}{c}{non-IID $\beta$ = 0.9} \\ \hline
Atk(.) setup & CC & TM & RFA & CC & TM & RFA & CC & TM & RFA \\
$\rho=0$ ,$\lambda=0$, $\pi=45$ & 80.74 $\pm$ 0.0 & 80.66 $\pm$ 0.18 & 77.78 $\pm$ 0.78 & 82.74 $\pm$ 0.02 & 83.41 $\pm$ 0.11 & \multicolumn{1}{l|}{83.3 $\pm$ 0.11} & 82.74 $\pm$ 0.02 & 75.76 $\pm$ 0.9 & 72.77 $\pm$ 1.38 \\
$\rho=0$ ,$\lambda=0$, $\pi=60$ & 74.62 $\pm$ 1.76 & 74.55 $\pm$ 2.34 & 74.3 $\pm$ 0.55 & 79.97 $\pm$ 0.39 & 79.48 $\pm$ 1.18 & \multicolumn{1}{l|}{81.71 $\pm$ 0.38} & 65.54 $\pm$ 2.6 & 72.31 $\pm$ 0.46 & 66.75 $\pm$ 1.51 \\
$\rho=0$ ,$\lambda=0$, $\pi=90$ & 69.4 $\pm$ 1.76 & 70.58 $\pm$ 0.34 & 65.51 $\pm$ 2.02 & 71.34 $\pm$ 0.37 & 73.04 $\pm$ 0.36 & \multicolumn{1}{l|}{74.51 $\pm$ 0.27} & 57.44 $\pm$ 1.4 & 62.53 $\pm$ 0.28 & 54.14 $\pm$ 2.06 \\
$\rho=0$ ,$\lambda=0$, $\pi=120$ & 65.42 $\pm$ 1.84 & 69.47 $\pm$ 0.92 & 54.4 $\pm$ 0.88 & 64.56 $\pm$ 0.05 & 62.59 $\pm$ 4.93 & \multicolumn{1}{l|}{64.54 $\pm$ 2.28} & 48.9 $\pm$ 0.54 & 57.59 $\pm$ 2.26 & 39.06 $\pm$ 1.27 \\
$\rho=0$ ,$\lambda=0$, $\pi=135$ & 66.52 $\pm$ 0.06 & 65.62 $\pm$ 0.5 & 50.73 $\pm$ 0.12 & 61.72 $\pm$ 2.34 & 60.93 $\pm$ 1.77 & \multicolumn{1}{l|}{55.19 $\pm$ 1.42} & 50.02 $\pm$ 1.18 & 60.0 $\pm$ 1.85 & 32.98 $\pm$ 1.61 \\
$\rho=0$ ,$\lambda=0$, $\pi=180$ & 84.5 $\pm$ 0.28 & 84.86 $\pm$ 0.18 & 65.84 $\pm$ 0.14 & 77.79 $\pm$ 0.06 & 76.18 $\pm$ 0.74 & \multicolumn{1}{l|}{74.53 $\pm$ 0.16} & 82.94 $\pm$ 0.06 & 79.33 $\pm$ 0.7 & 60.93 $\pm$ 0.75 \\
$\rho=0$ ,$\lambda=0.5$, $\pi=45$ & 77.24 $\pm$ 0.17 & 76.5 $\pm$ 0.64 & 72.42 $\pm$ 0.92 & 72.68 $\pm$ 0.66 & 77.36 $\pm$ 1.42 & \multicolumn{1}{l|}{79.49 $\pm$ 0.42} & 61.24 $\pm$ 1.18 & 71.7 $\pm$ 1.23 & 53.06 $\pm$ 3.43 \\
$\rho=0$ ,$\lambda=0.5$, $\pi=60$ & 73.1 $\pm$ 1.46 & 69.76 $\pm$ 1.63 & 66.6 $\pm$ 1.19 & 68.64 $\pm$ 0.64 & 74.32 $\pm$ 0.48 & \multicolumn{1}{l|}{74.21 $\pm$ 0.13} & 56.45 $\pm$ 0.8 & 64.42 $\pm$ 0.75 & 41.88 $\pm$ 1.89 \\
$\rho=0$ ,$\lambda=0.5$, $\pi=90$ & 66.74 $\pm$ 1.14 & 69.88 $\pm$ 1.32 & 50.83 $\pm$ 1.74 & 60.94 $\pm$ 2.08 & 67.07 $\pm$ 1.66 & \multicolumn{1}{l|}{56.4 $\pm$ 0.97} & 48.08 $\pm$ 0.18 & 57.26 $\pm$ 0.43 & 34.18 $\pm$ 0.7 \\
$\rho=0$ ,$\lambda=0.5$, $\pi=120$ & 66.86 $\pm$ 0.64 & 69.27 $\pm$ 0.57 & 37.09 $\pm$ 0.73 & 59.77 $\pm$ 1.48 & 67.43 $\pm$ 1.41 & \multicolumn{1}{l|}{\textbf{44.65 $\pm$ 1.38}} & 48.96 $\pm$ 0.62 & 59.13 $\pm$ 1.06 & 25.15 $\pm$ 2.38 \\
$\rho=0$ ,$\lambda=0.5$, $\pi=135$ & 66.64 $\pm$ 0.72 & 70.8 $\pm$ 0.01 & \textbf{30.22 $\pm$ 3.31} & 60.33 $\pm$ 1.38 & 63.04 $\pm$ 4.03 & \multicolumn{1}{l|}{50.1 $\pm$ 1.69} & 47.58 $\pm$ 1.01 & 59.89 $\pm$ 1.24 & \textbf{22.63 $\pm$ 1.98} \\
$\rho=0$ ,$\lambda=0.5$, $\pi=180$ & 85.11 $\pm$ 0.25 & 83.96 $\pm$ 1.0 & 62.4 $\pm$ 0.86 & 75.61 $\pm$ 0.81 & 75.6 $\pm$ 0.4 & \multicolumn{1}{l|}{68.08 $\pm$ 0.94} & 81.81 $\pm$ 0.71 & 78.95 $\pm$ 0.94 & 48.79 $\pm$ 2.98 \\
$\rho=0$ ,$\lambda=0.9$, $\pi=45$ & 75.0 $\pm$ 1.35 & 73.44 $\pm$ 0.22 & 66.34 $\pm$ 1.12 & 64.78 $\pm$ 0.2 & 74.76 $\pm$ 0.18 & \multicolumn{1}{l|}{70.66 $\pm$ 1.33} & 55.48 $\pm$ 2.13 & 67.68 $\pm$ 1.03 & 42.87 $\pm$ 1.94 \\
$\rho=0$ ,$\lambda=0.9$, $\pi=60$ & 69.65 $\pm$ 0.01 & 70.06 $\pm$ 0.62 & 53.77 $\pm$ 3.36 & 56.09 $\pm$ 0.88 & 70.62 $\pm$ 1.39 & \multicolumn{1}{l|}{60.68 $\pm$ 1.15} & 45.06 $\pm$ 0.93 & 63.28 $\pm$ 2.25 & 42.02 $\pm$ 3.0 \\
$\rho=0$ ,$\lambda=0.9$, $\pi=90$ & 63.6 $\pm$ 2.12 & 67.02 $\pm$ 2.43 & 55.34 $\pm$ 1.72 & 59.34 $\pm$ 0.48 & 68.2 $\pm$ 2.56 & \multicolumn{1}{l|}{61.78 $\pm$ 1.86} & 43.14 $\pm$ 2.15 & 59.69 $\pm$ 1.42 & 45.15 $\pm$ 2.04 \\
$\rho=0$ ,$\lambda=0.9$, $\pi=120$ & 67.51 $\pm$ 0.16 & 71.4 $\pm$ 1.04 & 60.86 $\pm$ 0.4 & 64.02 $\pm$ 1.08 & 71.14 $\pm$ 1.01 & \multicolumn{1}{l|}{66.65 $\pm$ 0.26} & 47.54 $\pm$ 1.8 & 61.98 $\pm$ 2.85 & 48.15 $\pm$ 0.55 \\
$\rho=0$ ,$\lambda=0.9$, $\pi=135$ & 69.77 $\pm$ 0.26 & 74.19 $\pm$ 1.74 & 64.76 $\pm$ 1.3 & 70.81 $\pm$ 0.38 & 71.92 $\pm$ 1.5 & \multicolumn{1}{l|}{71.2 $\pm$ 1.13} & 54.41 $\pm$ 2.03 & 65.58 $\pm$ 2.66 & 55.11 $\pm$ 4.07 \\
$\rho=0$ ,$\lambda=0.9$, $\pi=180$ & 83.4 $\pm$ 0.15 & 75.95 $\pm$ 0.22 & 85.4 $\pm$ 0.2 & 82.71 $\pm$ 0.28 & 83.6 $\pm$ 0.72 & \multicolumn{1}{l|}{84.2 $\pm$ 0.28} & 73.81 $\pm$ 0.76 & 79.25 $\pm$ 0.25 & 83.84 $\pm$ 0.46 \\
$\rho=0$ ,$\lambda=1$, $\pi=45$ & 73.63 $\pm$ 1.3 & 72.25 $\pm$ 0.24 & 57.04 $\pm$ 0.46 & 63.2 $\pm$ 0.96 & 73.67 $\pm$ 0.52 & \multicolumn{1}{l|}{63.82 $\pm$ 0.95} & 49.98 $\pm$ 3.24 & 65.3 $\pm$ 1.51 & 43.35 $\pm$ 0.44 \\
$\rho=0$ ,$\lambda=1$, $\pi=60$ & 70.76 $\pm$ 0.64 & 69.98 $\pm$ 0.62 & 57.49 $\pm$ 0.5 & 60.09 $\pm$ 0.32 & 68.32 $\pm$ 1.47 & \multicolumn{1}{l|}{58.74 $\pm$ 2.35} & 42.62 $\pm$ 2.6 & 64.78 $\pm$ 0.43 & 40.7 $\pm$ 1.8 \\
$\rho=0$ ,$\lambda=1$, $\pi=90$ & 63.25 $\pm$ 1.15 & 67.22 $\pm$ 1.9 & 58.24 $\pm$ 0.02 & 63.34 $\pm$ 1.15 & 68.53 $\pm$ 2.8 & \multicolumn{1}{l|}{63.96 $\pm$ 2.37} & 46.2 $\pm$ 1.1 & 59.63 $\pm$ 1.78 & 43.94 $\pm$ 1.02 \\
$\rho=0$ ,$\lambda=1$, $\pi=120$ & 67.46 $\pm$ 1.54 & 72.51 $\pm$ 0.31 & 60.29 $\pm$ 0.72 & 67.58 $\pm$ 0.63 & 70.99 $\pm$ 0.33 & \multicolumn{1}{l|}{70.51 $\pm$ 0.93} & 55.86 $\pm$ 0.3 & 65.7 $\pm$ 2.44 & 53.0 $\pm$ 1.89 \\
$\rho=0$ ,$\lambda=1$, $\pi=135$ & 68.72 $\pm$ 0.8 & 75.48 $\pm$ 0.53 & 66.25 $\pm$ 0.62 & 73.2 $\pm$ 0.78 & 73.39 $\pm$ 0.21 & \multicolumn{1}{l|}{73.1 $\pm$ 0.43} & 62.06 $\pm$ 0.72 & 68.86 $\pm$ 1.31 & 59.02 $\pm$ 2.18 \\
$\rho=0$ ,$\lambda=1$, $\pi=180$ & 83.36 $\pm$ 0.62 & 74.18 $\pm$ 0.57 & 85.24 $\pm$ 0.04 & 84.17 $\pm$ 0.36 & 83.92 $\pm$ 0.09 & \multicolumn{1}{l|}{84.38 $\pm$ 0.16} & 74.47 $\pm$ 4.82 & 80.69 $\pm$ 0.77 & 83.65 $\pm$ 0.35 \\
$\rho=0.5$ ,$\lambda=0$, $\pi=45$ & 79.2 $\pm$ 1.74 & 78.75 $\pm$ 0.63 & 77.94 $\pm$ 0.08 & 82.5 $\pm$ 0.48 & 83.84 $\pm$ 0.05 & \multicolumn{1}{l|}{83.66 $\pm$ 0.47} & 64.54 $\pm$ 0.3 & 76.36 $\pm$ 1.33 & 71.34 $\pm$ 1.68 \\
$\rho=0.5$ ,$\lambda=0$, $\pi=60$ & 73.91 $\pm$ 0.96 & 73.27 $\pm$ 0.25 & 73.48 $\pm$ 0.96 & 78.79 $\pm$ 1.12 & 79.94 $\pm$ 0.56 & \multicolumn{1}{l|}{81.19 $\pm$ 0.24} & 62.55 $\pm$ 0.02 & 67.66 $\pm$ 1.39 & 65.64 $\pm$ 1.08 \\
$\rho=0.5$ ,$\lambda=0$, $\pi=90$ & 67.13 $\pm$ 0.24 & 66.51 $\pm$ 0.7 & 64.76 $\pm$ 1.29 & 67.55 $\pm$ 2.8 & 71.62 $\pm$ 0.8 & \multicolumn{1}{l|}{75.3 $\pm$ 0.35} & 50.4 $\pm$ 1.56 & 57.47 $\pm$ 2.15 & 50.4 $\pm$ 1.71 \\
$\rho=0.5$ ,$\lambda=0$, $\pi=120$ & 62.43 $\pm$ 0.78 & 63.54 $\pm$ 1.74 & 56.21 $\pm$ 0.96 & 57.65 $\pm$ 0.59 & 64.9 $\pm$ 1.31 & \multicolumn{1}{l|}{64.27 $\pm$ 1.86} & 45.53 $\pm$ 2.59 & 52.5 $\pm$ 3.33 & 38.24 $\pm$ 2.75 \\
$\rho=0.5$ ,$\lambda=0$, $\pi=135$ & 64.72 $\pm$ 0.31 & 63.4 $\pm$ 1.17 & 55.61 $\pm$ 2.06 & 56.16 $\pm$ 1.57 & \textbf{58.85 $\pm$ 0.22} & \multicolumn{1}{l|}{59.64 $\pm$ 0.28} & 42.62 $\pm$ 1.86 & 55.41 $\pm$ 0.38 & 34.75 $\pm$ 0.48 \\
$\rho=0.5$ ,$\lambda=0$, $\pi=180$ & 84.38 $\pm$ 0.14 & 83.76 $\pm$ 0.3 & 75.84 $\pm$ 0.12 & 78.22 $\pm$ 0.42 & 76.28 $\pm$ 0.47 & \multicolumn{1}{l|}{77.12 $\pm$ 1.06} & 82.35 $\pm$ 0.36 & 75.47 $\pm$ 1.1 & 67.69 $\pm$ 0.08 \\
$\rho=0.5$ ,$\lambda=0.5$, $\pi=45$ & 75.68 $\pm$ 0.9 & 73.66 $\pm$ 0.18 & 67.72 $\pm$ 0.49 & 68.74 $\pm$ 0.26 & 77.98 $\pm$ 0.7 & \multicolumn{1}{l|}{79.87 $\pm$ 0.58} & 57.4 $\pm$ 0.8 & 66.56 $\pm$ 1.15 & 55.33 $\pm$ 1.28 \\
$\rho=0.5$ ,$\lambda=0.5$, $\pi=60$ & 71.44 $\pm$ 0.2 & 68.34 $\pm$ 1.02 & 69.12 $\pm$ 0.56 & 64.46 $\pm$ 0.11 & 70.82 $\pm$ 2.62 & \multicolumn{1}{l|}{73.99 $\pm$ 1.44} & 45.32 $\pm$ 0.42 & 62.47 $\pm$ 1.08 & 40.93 $\pm$ 0.52 \\
$\rho=0.5$ ,$\lambda=0.5$, $\pi=90$ & 63.36 $\pm$ 0.9 & 64.08 $\pm$ 0.35 & 54.08 $\pm$ 1.4 & 56.25 $\pm$ 0.32 & 66.78 $\pm$ 0.45 & \multicolumn{1}{l|}{60.15 $\pm$ 2.1} & 41.36 $\pm$ 2.2 & 58.04 $\pm$ 0.76 & 32.42 $\pm$ 2.11 \\
$\rho=0.5$ ,$\lambda=0.5$, $\pi=120$ & 59.81 $\pm$ 0.87 & 67.31 $\pm$ 0.34 & 40.05 $\pm$ 0.52 & 52.92 $\pm$ 0.9 & 65.65 $\pm$ 0.42 & \multicolumn{1}{l|}{48.64 $\pm$ 1.07} & 39.33 $\pm$ 1.42 & 58.17 $\pm$ 0.82 & 27.22 $\pm$ 2.78 \\
$\rho=0.5$ ,$\lambda=0.5$, $\pi=135$ & 63.23 $\pm$ 0.24 & 68.01 $\pm$ 0.35 & 37.77 $\pm$ 3.74 & 55.47 $\pm$ 1.96 & 63.97 $\pm$ 0.22 & \multicolumn{1}{l|}{50.86 $\pm$ 3.3} & 42.0 $\pm$ 0.18 & 58.03 $\pm$ 1.19 & 28.53 $\pm$ 1.82 \\
$\rho=0.5$ ,$\lambda=0.5$, $\pi=180$ & 83.14 $\pm$ 0.0 & 84.04 $\pm$ 0.9 & 66.88 $\pm$ 0.36 & 74.13 $\pm$ 0.22 & 74.72 $\pm$ 0.66 & \multicolumn{1}{l|}{72.44 $\pm$ 0.42} & 80.82 $\pm$ 0.4 & 77.24 $\pm$ 0.61 & 58.43 $\pm$ 0.58 \\
$\rho=0.5$ ,$\lambda=0.9$, $\pi=45$ & 70.84 $\pm$ 1.88 & 71.6 $\pm$ 0.28 & 68.07 $\pm$ 0.2 & 62.46 $\pm$ 1.0 & 73.7 $\pm$ 1.22 & \multicolumn{1}{l|}{72.79 $\pm$ 1.86} & 41.02 $\pm$ 1.62 & 63.47 $\pm$ 1.73 & 38.94 $\pm$ 1.49 \\
$\rho=0.5$ ,$\lambda=0.9$, $\pi=60$ & 65.84 $\pm$ 0.22 & 67.81 $\pm$ 0.48 & 49.15 $\pm$ 5.92 & 53.92 $\pm$ 0.82 & 70.62 $\pm$ 0.9 & \multicolumn{1}{l|}{58.2 $\pm$ 3.36} & 33.94 $\pm$ 0.86 & 61.6 $\pm$ 0.8 & 37.58 $\pm$ 2.1 \\
$\rho=0.5$ ,$\lambda=0.9$, $\pi=90$ & 57.82 $\pm$ 1.56 & 65.45 $\pm$ 0.66 & 49.6 $\pm$ 0.68 & 54.36 $\pm$ 1.68 & 68.96 $\pm$ 0.04 & \multicolumn{1}{l|}{57.62 $\pm$ 1.72} & 37.25 $\pm$ 0.76 & 59.63 $\pm$ 1.3 & 38.34 $\pm$ 3.65 \\
$\rho=0.5$ ,$\lambda=0.9$, $\pi=120$ & 57.59 $\pm$ 1.91 & 69.04 $\pm$ 0.46 & 54.74 $\pm$ 1.32 & 63.17 $\pm$ 1.27 & 68.11 $\pm$ 0.25 & \multicolumn{1}{l|}{62.53 $\pm$ 2.68} & 38.23 $\pm$ 0.59 & 59.3 $\pm$ 1.36 & 40.91 $\pm$ 1.4 \\
$\rho=0.5$ ,$\lambda=0.9$, $\pi=135$ & 59.22 $\pm$ 0.97 & 68.9 $\pm$ 0.31 & 58.1 $\pm$ 1.38 & 67.46 $\pm$ 0.44 & 71.56 $\pm$ 1.25 & \multicolumn{1}{l|}{65.13 $\pm$ 2.35} & 38.83 $\pm$ 2.17 & 63.68 $\pm$ 1.31 & 40.49 $\pm$ 1.03 \\
$\rho=0.5$ ,$\lambda=0.9$, $\pi=180$ & 79.65 $\pm$ 1.06 & 72.14 $\pm$ 0.77 & 82.68 $\pm$ 0.14 & 80.94 $\pm$ 0.8 & 83.18 $\pm$ 0.54 & \multicolumn{1}{l|}{82.68 $\pm$ 0.0} & 70.48 $\pm$ 1.19 & 79.63 $\pm$ 0.94 & 80.34 $\pm$ 0.45 \\
$\rho=0.5$ ,$\lambda=1$, $\pi=45$ & 71.42 $\pm$ 0.71 & 70.6 $\pm$ 0.38 & 66.76 $\pm$ 1.11 & 71.42 $\pm$ 0.71 & 72.82 $\pm$ 0.78 & \multicolumn{1}{l|}{68.27 $\pm$ 2.2} & 31.82 $\pm$ 1.16 & 62.89 $\pm$ 1.4 & 40.73 $\pm$ 1.95 \\
$\rho=0.5$ ,$\lambda=1$, $\pi=60$ & 64.32 $\pm$ 0.1 & 66.16 $\pm$ 2.16 & 51.94 $\pm$ 0.86 & 56.32 $\pm$ 1.52 & 67.78 $\pm$ 1.26 & \multicolumn{1}{l|}{59.39 $\pm$ 1.46} & 36.55 $\pm$ 3.87 & 57.72 $\pm$ 2.99 & 39.61 $\pm$ 0.39 \\
$\rho=0.5$ ,$\lambda=1$, $\pi=90$ & 59.14 $\pm$ 1.2 & 65.63 $\pm$ 1.44 & 54.02 $\pm$ 0.74 & 59.32 $\pm$ 2.04 & 69.46 $\pm$ 1.3 & \multicolumn{1}{l|}{61.13 $\pm$ 1.53} & 38.74 $\pm$ 0.5 & 57.86 $\pm$ 1.37 & 40.1 $\pm$ 1.72 \\
$\rho=0.5$ ,$\lambda=1$, $\pi=120$ & 60.86 $\pm$ 0.76 & 71.09 $\pm$ 0.96 & 57.49 $\pm$ 0.21 & 65.48 $\pm$ 2.62 & 70.2 $\pm$ 0.34 & \multicolumn{1}{l|}{67.14 $\pm$ 1.0} & 49.18 $\pm$ 0.4 & 59.7 $\pm$ 1.72 & 43.98 $\pm$ 2.4 \\
$\rho=0.5$ ,$\lambda=1$, $\pi=135$ & 65.24 $\pm$ 1.14 & 73.39 $\pm$ 1.24 & 59.14 $\pm$ 1.06 & 70.03 $\pm$ 0.68 & 70.26 $\pm$ 0.22 & \multicolumn{1}{l|}{68.51 $\pm$ 2.32} & 56.76 $\pm$ 0.2 & 62.95 $\pm$ 3.02 & 44.72 $\pm$ 0.72 \\
$\rho=0.5$ ,$\lambda=1$, $\pi=180$ & 81.13 $\pm$ 0.08 & 73.08 $\pm$ 0.08 & 84.24 $\pm$ 0.14 & 82.57 $\pm$ 0.5 & 83.18 $\pm$ 0.72 & \multicolumn{1}{l|}{84.35 $\pm$ 0.08} & 75.31 $\pm$ 0.48 & 79.23 $\pm$ 0.97 & 82.08 $\pm$ 1.04 \\
$\rho=1$ ,$\lambda=0$, $\pi=45$ & 77.97 $\pm$ 0.59 & 70.6 $\pm$ 0.38 & 83.57 $\pm$ 0.39 & 56.62 $\pm$ 1.14 & 83.74 $\pm$ 0.34 & \multicolumn{1}{l|}{81.76 $\pm$ 0.3} & 62.4 $\pm$ 1.27 & 74.17 $\pm$ 0.65 & 62.73 $\pm$ 0.4 \\
$\rho=1$ ,$\lambda=0$, $\pi=60$ & 72.95 $\pm$ 0.81 & 71.41 $\pm$ 0.16 & 77.84 $\pm$ 1.2 & 56.32 $\pm$ 1.52 & 79.7 $\pm$ 0.1 & \multicolumn{1}{l|}{81.78 $\pm$ 0.35} & 53.87 $\pm$ 1.48 & 64.83 $\pm$ 2.02 & 61.99 $\pm$ 2.75 \\
$\rho=1$ ,$\lambda=0$, $\pi=90$ & 66.06 $\pm$ 1.54 & 63.78 $\pm$ 0.45 & 66.27 $\pm$ 1.93 & 66.31 $\pm$ 0.06 & 68.94 $\pm$ 0.22 & \multicolumn{1}{l|}{73.9 $\pm$ 1.06} & 42.37 $\pm$ 3.8 & 55.06 $\pm$ 0.52 & 48.34 $\pm$ 2.59 \\
$\rho=1$ ,$\lambda=0$, $\pi=120$ & 61.7 $\pm$ 1.29 & 59.42 $\pm$ 1.96 & 59.28 $\pm$ 0.42 & 51.92 $\pm$ 0.0 & 66.7 $\pm$ 0.14 & \multicolumn{1}{l|}{65.43 $\pm$ 0.37} & 40.41 $\pm$ 1.84 & 49.21 $\pm$ 1.06 & 41.15 $\pm$ 1.27 \\
$\rho=1$ ,$\lambda=0$, $\pi=135$ & 59.11 $\pm$ 1.46 & \textbf{60.09 $\pm$ 1.01} & 57.63 $\pm$ 1.33 & 48.44 $\pm$ 1.36 & 65.54 $\pm$ 0.3 & \multicolumn{1}{l|}{64.59 $\pm$ 1.68} & 36.44 $\pm$ 0.9 & \textbf{47.74 $\pm$ 2.82} & 38.73 $\pm$ 0.94 \\
$\rho=1$ ,$\lambda=0$, $\pi=180$ & 79.66 $\pm$ 0.4 & 83.56 $\pm$ 0.18 & 78.46 $\pm$ 0.04 & 77.6 $\pm$ 0.18 & 73.66 $\pm$ 0.58 & \multicolumn{1}{l|}{78.3 $\pm$ 0.3} & 82.09 $\pm$ 0.19 & 74.09 $\pm$ 1.24 & 82.31 $\pm$ 1.53 \\
$\rho=1$ ,$\lambda=0.5$, $\pi=45$ & 72.04 $\pm$ 0.04 & 70.78 $\pm$ 1.27 & 74.02 $\pm$ 0.22 & 62.9 $\pm$ 1.85 & 78.52 $\pm$ 0.08 & \multicolumn{1}{l|}{79.57 $\pm$ 0.66} & 49.55 $\pm$ 1.95 & 63.84 $\pm$ 0.54 & 55.97 $\pm$ 2.3 \\
$\rho=1$ ,$\lambda=0.5$, $\pi=60$ & 67.86 $\pm$ 0.12 & 65.67 $\pm$ 1.1 & 68.53 $\pm$ 0.56 & 58.76 $\pm$ 0.3 & 71.89 $\pm$ 1.6 & \multicolumn{1}{l|}{74.79 $\pm$ 0.69} & 29.7 $\pm$ 0.18 & 56.87 $\pm$ 1.22 & 42.81 $\pm$ 0.67 \\
$\rho=1$ ,$\lambda=0.5$, $\pi=90$ & 59.88 $\pm$ 1.1 & 61.26 $\pm$ 1.25 & 55.5 $\pm$ 0.14 & 50.5 $\pm$ 0.52 & 68.51 $\pm$ 0.39 & \multicolumn{1}{l|}{60.48 $\pm$ 4.17} & 37.16 $\pm$ 0.38 & 54.25 $\pm$ 1.2 & 34.75 $\pm$ 0.88 \\
$\rho=1$ ,$\lambda=0.5$, $\pi=120$ & 56.36 $\pm$ 1.03 & 62.98 $\pm$ 0.46 & 45.28 $\pm$ 3.65 & 50.24 $\pm$ 2.12 & 60.91 $\pm$ 1.44 & \multicolumn{1}{l|}{56.32 $\pm$ 0.98} & 32.55 $\pm$ 1.17 & 49.54 $\pm$ 3.02 & 33.49 $\pm$ 1.06 \\
$\rho=1$ ,$\lambda=0.5$, $\pi=135$ & 55.2 $\pm$ 1.68 & 65.5 $\pm$ 1.75 & 47.92 $\pm$ 1.01 & 52.74 $\pm$ 0.86 & 65.12 $\pm$ 0.1 & \multicolumn{1}{l|}{54.17 $\pm$ 3.76} & 33.95 $\pm$ 3.45 & 52.47 $\pm$ 2.49 & 34.41 $\pm$ 2.39 \\
$\rho=1$ ,$\lambda=0.5$, $\pi=180$ & 77.22 $\pm$ 0.04 & 83.62 $\pm$ 0.44 & 74.9 $\pm$ 0.57 & 74.42 $\pm$ 1.1 & 73.74 $\pm$ 0.66 & \multicolumn{1}{l|}{73.22 $\pm$ 0.1} & 76.93 $\pm$ 0.64 & 75.72 $\pm$ 0.57 & 79.61 $\pm$ 1.69 \\
$\rho=1$ ,$\lambda=0.9$, $\pi=45$ & 68.08 $\pm$ 0.83 & 67.66 $\pm$ 1.16 & 67.55 $\pm$ 0.74 & 55.0 $\pm$ 1.29 & 73.68 $\pm$ 0.46 & \multicolumn{1}{l|}{73.93 $\pm$ 1.33} & 31.69 $\pm$ 0.89 & 61.48 $\pm$ 1.15 & 38.46 $\pm$ 1.76 \\
$\rho=1$ ,$\lambda=0.9$, $\pi=60$ & 60.9 $\pm$ 0.58 & 64.58 $\pm$ 0.98 & 59.2 $\pm$ 2.06 & 50.93 $\pm$ 0.62 & 68.66 $\pm$ 0.8 & \multicolumn{1}{l|}{56.73 $\pm$ 1.95} & 28.98 $\pm$ 2.57 & 55.85 $\pm$ 1.37 & 47.86 $\pm$ 1.0 \\
$\rho=1$ ,$\lambda=0.9$, $\pi=90$ & 46.23 $\pm$ 1.03 & 64.12 $\pm$ 0.54 & 34.26 $\pm$ 0.42 & \textbf{49.32 $\pm$ 0.4} & 63.82 $\pm$ 2.46 & \multicolumn{1}{l|}{56.49 $\pm$ 1.3} & \textbf{28.86 $\pm$ 0.26} & 55.55 $\pm$ 1.81 & 34.26 $\pm$ 0.42 \\
$\rho=1$ ,$\lambda=0.9$, $\pi=120$ & 40.42 $\pm$ 1.96 & 66.54 $\pm$ 0.48 & 47.15 $\pm$ 2.6 & 57.18 $\pm$ 0.14 & 69.9 $\pm$ 0.2 & \multicolumn{1}{l|}{60.5 $\pm$ 0.84} & 27.26 $\pm$ 2.84 & 56.03 $\pm$ 2.02 & 32.46 $\pm$ 1.07 \\
$\rho=1$ ,$\lambda=0.9$, $\pi=135$ & 44.78 $\pm$ 1.2 & 66.49 $\pm$ 1.57 & 47.51 $\pm$ 0.33 & 61.53 $\pm$ 1.36 & 70.78 $\pm$ 1.22 & \multicolumn{1}{l|}{62.04 $\pm$ 0.76} & 31.08 $\pm$ 1.56 & 58.76 $\pm$ 1.24 & 33.16 $\pm$ 1.73 \\
$\rho=1$ ,$\lambda=0.9$, $\pi=180$ & 68.32 $\pm$ 0.04 & 83.57 $\pm$ 0.67 & 71.32 $\pm$ 0.2 & 78.47 $\pm$ 0.01 & 82.04 $\pm$ 1.06 & \multicolumn{1}{l|}{79.68 $\pm$ 0.24} & 63.46 $\pm$ 1.36 & 75.76 $\pm$ 1.79 & 63.91 $\pm$ 1.91 \\
$\rho=1$ ,$\lambda=1$, $\pi=45$ & 67.66 $\pm$ 1.9 & 67.03 $\pm$ 0.1 & 66.05 $\pm$ 0.99 & 53.1 $\pm$ 1.26 & 71.58 $\pm$ 0.22 & \multicolumn{1}{l|}{69.68 $\pm$ 2.6} & 28.4 $\pm$ 2.78 & 58.36 $\pm$ 0.63 & 35.13 $\pm$ 2.72 \\
$\rho=1$ ,$\lambda=1$, $\pi=60$ & 58.69 $\pm$ 1.18 & 64.78 $\pm$ 0.62 & 44.98 $\pm$ 6.9 & 53.72 $\pm$ 0.6 & 67.88 $\pm$ 0.06 & \multicolumn{1}{l|}{58.6 $\pm$ 2.19} & 29.7 $\pm$ 1.34 & 56.57 $\pm$ 0.95 & 35.56 $\pm$ 0.64 \\
$\rho=1$ ,$\lambda=1$, $\pi=90$ & \textbf{40.08 $\pm$ 4.93} & 62.5 $\pm$ 0.06 & 48.2 $\pm$ 4.14 & 57.02 $\pm$ 1.76 & 67.1 $\pm$ 0.6 & \multicolumn{1}{l|}{60.3 $\pm$ 0.45} & 35.77 $\pm$ 1.67 & 55.22 $\pm$ 0.86 & 37.85 $\pm$ 0.73 \\
$\rho=1$ ,$\lambda=1$, $\pi=120$ & 51.84 $\pm$ 1.03 & 67.36 $\pm$ 0.42 & 52.54 $\pm$ 0.32 & 63.75 $\pm$ 0.14 & 69.73 $\pm$ 0.11 & \multicolumn{1}{l|}{61.04 $\pm$ 1.14} & 39.39 $\pm$ 2.1 & 57.47 $\pm$ 1.68 & 36.05 $\pm$ 1.6 \\
$\rho=1$ ,$\lambda=1$, $\pi=135$ & 57.11 $\pm$ 2.46 & 69.75 $\pm$ 0.2 & 53.56 $\pm$ 0.42 & 68.96 $\pm$ 1.66 & 69.62 $\pm$ 2.43 & \multicolumn{1}{l|}{64.91 $\pm$ 0.71} & 47.06 $\pm$ 0.32 & 61.63 $\pm$ 2.86 & 36.35 $\pm$ 0.91 \\
$\rho=1$ ,$\lambda=1$, $\pi=180$ & 76.18 $\pm$ 1.58 & 84.67 $\pm$ 0.68 & 79.09 $\pm$ 0.22 & 81.12 $\pm$ 0.9 & 82.35 $\pm$ 0.37 & \multicolumn{1}{l|}{81.98 $\pm$ 0.46} & 71.3 $\pm$ 1.21 & 77.19 $\pm$ 1.41 & 68.29 $\pm$ 1.38 \\ \hline
\end{tabular}%
}
\end{table*}

\end{document}